\documentclass[journal, 12pt, draftclsnofoot, onecolumn]{resources/IEEEtran}

\usepackage[margin=1in]{geometry}

\usepackage[utf8]{inputenc}

\usepackage{amsthm, amsmath, amsfonts, amssymb}
\usepackage{enumerate}
\usepackage{graphicx}
\usepackage{mathtools}
\usepackage{thmtools}
\usepackage{thm-restate}
\usepackage{float}
\usepackage{enumerate}
\usepackage{marginnote}
\usepackage{hyperref}
\usepackage{cleveref}
\usepackage{subfigure}
\usepackage{algpseudocode}
\usepackage{algorithm}
\usepackage{mathabx}
\usepackage[affil-it]{authblk}

\usepackage{resources/custom_commands}
\usepackage{resources/coloredboxes}

\newcommand{\cost}{{\, \text{\rm{\textcentoldstyle}}}}

\newcommand{\cspace}{\cQ}
\newcommand{\wspace}{\cX}
\newcommand{\proj}[1]{[#1]_\wspace}
\newcommand{\projinv}[1]{[#1]^\cspace}
\newcommand{\defeq}{:=}
\newcommand{\tsp}{{\mathrm{TSP}}}

\newcommand{\cbo}{{\mathrm{CBO}}}

\newcommand{\scp}{{\mathrm{SCP}}}
\newcommand{\hcp}{{\mathrm{HCP}}}
\newcommand{\hcs}{{\mathrm{HCS}}}
\newcommand{\vol}{{\mathrm{Vol}}}
\newcommand{\bern}{{\mathrm{Bern}}}

\newcommand{\wvhp}{wvhp}
\newcommand{\wspacef}{\wspace_f}

\title{Agility and Target Distribution in the Dynamic Stochastic Traveling Salesman Problem}

\author[1]{Aviv Adler}
\author[2]{Oren Gal}
\author[3]{Sertac Karaman}

\affil[1,3]{Laboratory for Information \& Decision Systems, MIT}
\affil[2]{Technion, Israel Institute of Technology}
\affil[ ]{$~^{1,3}$\textit{\{adlera,sertac\}@mit.edu, $~^2$orengal@alumni.technion.ac.il}}

\date{\today}

\usepackage{subfiles}

\begin{document}




\maketitle

\begin{abstract}
    An important variant of the classic Traveling Salesman Problem (TSP) is the Dynamic TSP, in which a system with dynamic constraints is tasked with visiting a set of $n$ target locations (in any order) in the shortest amount of time. Such tasks arise naturally in many robotic motion planning problems, particularly in exploration, surveillance and reconnaissance, and classical TSP algorithms on graphs are typically inapplicable in this setting. An important question about such problems is: if the target points are random, what is the length of the tour (either in expectation or as a concentration bound) as $n$ grows? This problem is the Dynamic Stochastic TSP (DSTSP), and has been studied both for specific important vehicle models and for general dynamic systems; however, in general only the order of growth is known. In this work, we explore the connection between the distribution from which the targets are drawn and the dynamics of the system, yielding a more precise lower bound on tour length as well as a matching upper bound for the case of symmetric (or \emph{driftless}) systems. We then extend the symmetric dynamics results to the case when the points are selected by a (non-random) adversary whose goal is to maximize the length, thus showing worst-case bounds on the tour length.
\end{abstract}

\section{Introduction and Motivation} \label{sec:intro}

\subsection{Introduction, motivation, and previous work}

The Traveling Salesman Problem (TSP) is a classic problem in computer science and optimization, in which an agent is tasked with visiting a set of target locations, in any order, in the minimum amount of time. In its most common form, the target locations are modeled as a graph with weighted edges indicating the distance between their endpoints. In addition to having an astounding range of practical applications, it attracts a great deal of theoretical interest in many settings.

The TSP is well-known to be computationally challenging -- specifically, finding the shortest tour through a given set of points is NP-complete~\cite{karp-72}, even for Euclidean paths through points in $\bbR^2$~\cite{papadimitriou-77}. Nevertheless, there are a number of successful algorithmic methods for this problem, including approximation algorithms and heuristics.

Although the TSP originated in operations research, it has found numerous applications in the context of robotics as well. Most notably, a number of robot motion planning and routing algorithms employ TSP algorithms at their core~\cite{lavalle-06}. The applications of TSP in the robotics domain are far reaching, including persistent monitoring, surveillance, reconnaissance, exploration, among other important problems.

However, in most cases in robotics, the vehicles are subject to non-trivial differential constraints which have a substantial impact on the optimal tour. Furthermore, the addition of dynamic constraints means there is no fixed notion of `distance' between target points, as the point-to-point travel time depends on the specific configurations (e.g. heading for a Dubins car) one wishes to have at the two target points in question. This means that the classic approximation algorithms do not apply to this setting, and new algorithms must be used to deal with the TSP in these instances~\cite{leny-06}. Furthermore, understanding the impact of the dynamics on the length of the tour would allow a system designer to pick the best robot for the task at hand. We refer to this as the \emph{Dynamic TSP} (DTSP).

An important variant of the TSP is the case in which the targets are independently and identically distributed (iid), in which case the goal is to understand the behavior of the shortest tour length as a random variable dependent on the targets, and in particular to understand how this tour length grows as the number $n$ of targets is increased. We refer to this as the \emph{Stochastic TSP}; when combined with dynamic constraints, it is the \emph{Dynamic Stochastic TSP} (DSTSP).

The importance of the Stochastic TSP has not gone unnoticed~\cite{frieze-07}. The general asymptotic behavior of the Euclidean Stochastic TSP was characterized by Beardwood et al. in 1959~\cite{beardwood-59}; later, it was studied extensively from an algorithmic perspective, in particular on discovering algorithms which on average achieve high-quality approximations in polynomial time~\cite{karp-77}.
In addition, it was found to have connections to a variety of practical and theoretical problems, such as vehicle routing~\cite{bertsimas-95} and matching on the Euclidean plane~\cite{anthes-01}. In the second half of the 2000's, tight (up to a constant factor) probabilistic asymptotic bounds on the stochastic TSP were found for a number of vehicles, notably the Dubins car, Reeds-Shepp car, differential-drive vehicles, and double integrators in $2$ and $3$ dimensions (which are commonly used to model quadcopter dynamics)~\cite{enright-06, savla-05, savla-06, savla-06-2, savla-08}. The stochastic TSP for a symmetric dynamical system in 2 dimensions was considered by Itani et al.~\cite{itani-08}, using a technique based on subadditive functionals, which was then expanded in Itani's Ph.D thesis~\cite{itani-thesis-12} to include a wide class of translation-invariant dynamics in any Euclidean space $\bbR^d$. Finally, in 2016, we used an alternative elementary approach based on a discretization technique developed by Arias-Castro et al.~\cite{arias-ctd-05} for the related Stochastic Orienteering problem to find the asymptotic order-of-growth as $n \to \infty$ for a wide class of dynamics on manifolds of any dimension~\cite{adler-icra-16}. In all of these works, algorithms were given which achieved (with high probability as the number of targets grows) a tour length within a constant factor of the optimal with respect to $n$.

\subsection{Contribution}

While previous work on this problem strongly characterizes the asymptotic behavior of the length of the DSTSP tour as the number $n$ of targets increases - in particular, the dynamics induce a parameter, called the \emph{small-time constraint factor}~\cite{itani-thesis-12,adler-icra-16} and denoted in this work as $\gamma$, such that the length of the tour is with high probability $\Theta(n^{1-\frac{1}{\gamma}})$ -- the constant factor associated with this growth was generally not studied closely, and its relationship to the target point distribution was not generally known.

Furthermore, previous work generally assumed that the distribution of target points is uniform over a unit square or rectilinear region. While this does not affect the growth rate, as any full-dimensional continuous probability distribution with bounded support can be divided into finitely many approximately square regions, which can then be handled in sequence, changing the target point distribution may greatly affect the constant factor attached to the $\Theta(n^{1-\frac{1}{\gamma}})$ DSTSP tour length growth rate. 

In this work, we give the first high-probability DSTSP bounds with explicit constants that hold over a wide class of dynamic constraints and target point probability distributions. Specifically, we give an explicit high-probability lower bound for a broad class of dynamics, and a matching upper bound (to a factor constant in both the number $n$ and density function $f$ of the target points) for the special case of \emph{symmetric} (or \emph{driftless}) dynamics, which admit a simplified algorithm. To get these bounds, for any dynamics we define the \emph{agility function} it induces over the workspace, and show how the agility function and the probability density function of the target point distribution affect the constant factor of the DSTSP tour length.

\subsection{Paper organization}

The rest of this work is organized as follows. In \Cref{sec:prob-formulation} we formally define the problem and the terms used in the results. Because our results require several assumptions we state the results first, in \Cref{sec:results}, before we give the assumptions in \Cref{sec:assumptions}; many of the assumptions are also shown to hold in general when the dynamic constraints are \emph{control-affine}. We then show our probabilistic lower bound for the TSP tour length in \Cref{sec:dstsp-lb}, where we first set up a version of the Stochastic Orienteering problem and show an important preliminary result which we then use to show the TSP lower bound, and our probabilistic upper bound in \Cref{sec:dstsp-ub}. Finally, we extend these results to the case where the targets are placed to maximize the tour length in \Cref{sec:worst-case}. 

\section{Preliminaries and Problem Definition} \label{sec:prob-formulation}

In this work we study the Dynamic Stochastic TSP, in which a dynamic system on a manifold $\wspace$, which we call the \emph{workspace}, must visit a set of independently and identically-distributed (iid) target points $X_1, \dots, X_n \in \wspace$, in any order, in as short a time as possible. Common examples of such dynamic systems include the Dubins Car, the Differential-Drive Vehicle, and the Double Integrator (either in $\bbR^2$ or $\bbR^3$, commonly used as a model of quadcopter motion).

In what follows, the time derivative of a process $a(t)$ (say, a vehicle's position) on a manifold $\cA$ is given by $\dot{a}(t)$, which represents a vector tangent to $\cA$ at $a$; we do not use $a'$ to indicate derivatives. Instead, $a'$ will often be used to denote another point in the same space, e.g. `for any $a$ and $a'$ in $\cA$, the following condition holds'.

\subsection{The dynamic system} 

Although we are primarily concerned about the system's trajectory through the workspace $\wspace$ (which determines if and when each target point is visited), the dynamic constraints mean the state of the system at any given time is more naturally represented as a point $\bq$ in the \emph{configuration space} (or \emph{phase space}) $\cspace$. Its position in $\wspace$ is then a function of its position in $\cspace$, denoted as $\proj{\bq} = x \in \wspace$;
\footnote[1]{Typically $\cspace = \wspace \times \cC$ for some $\cC$, and $\proj{\cdot}$ is the projection function. It is possible that $\cspace = \wspace$, but this is a special case.} 
we denote the preimage of $x \in \wspace$ as $\projinv{x} = \{\bq \in \cspace : \proj{\bq} = x\}$. We assume this function works locally as a projection from a $\dim(\cspace)$-dimensional Euclidean space onto a $\dim(\wspace)$-dimensional subspace, i.e. for any $\bq, x$ such that $\proj{\bq} = x$, there are some neighborhoods around $\bq$ and $x$ and coordinate maps on these neighborhoods so that $\proj{\cdot}$ behaves linearly with respect to these maps.

For convenience we assume this mapping is well-behaved:
\begin{assumption}
    $\proj{\cdot}$ is smooth.
\end{assumption}
This doesn't need to hold completely everywhere for our results to still hold (see \Cref{sec:appx-everywhere} and Appendix~\ref{appx:appx-everywhere}) but we assume it for what follows.

Given a position $\bq(t) \in \cspace$ at time $t$ and a control input $\bu$ from some control set $\cU$, the system evolves according to a \emph{control law} $\dot{\bq}(t) = \hcont(\bq(t), \bu)$, where $\hcont$ takes a position in $\cspace$ and a control and returns a tangent vector indicating the direction in which the system moves \cite{lavalle-06}. We call a trajectory $\pi : [0,T] \to \cspace$ \emph{valid} if there is some control function $\bu : [0, T] \to \cU$ such that $\dot{\pi}(t) = \hcont(\pi(t), \bu(t))$ for all $t \in [0,T]$. 


We now introduce some notation which we will use for this work.
We denote the set of all valid trajectories as $\Pi$ (we will also refer to the dynamics by the set $\Pi$ of valid trajectories it induces). Interpreting $t$ as time, $T$ is then the amount of time it takes to execute trajectory $\pi \in \Pi$ via the control function $\bu$, which we refer to as the \emph{length} of the trajectory $\pi$ and denote as $\ell(\pi)$, so that $\pi : [0,\ell(\pi)] \to \cspace$.
Since the targets are points in $\wspace$ but the dynamics are specified over $\cspace$, we will be interested in the projection of trajectory $\pi \in \Pi$ onto $\wspace$. We will denote:
\begin{align}
    \bar{\pi}(t) = \proj{\pi(t)}
\end{align}
In general $\bar{\cdot}$ will denote the workspace projection of a set or function. We will also slightly abuse notation by writing ``$x \in \bar{\pi}$'' to mean ``$\exists \, t \text{ s.t. } x = \bar{\pi}(t)$'' i.e. that $\pi$ visits $x$ at some point, and similarly ``$\bq \in \pi$'' if there exists $t$ such that $\bq = \pi(t)$.

\subsection{Symmetric dynamics}

We say that dynamics $\Pi$ defined by control law $\hcont$ and control set $\cU$ are \emph{symmetric} if for every $\bq \in \cspace, \bu \in \cU$ there is a $\bu' \in \cU$ such that
\begin{align}
    \hcont(\bq,\bu') = -\hcont(\bq,\bu)
\end{align}
i.e. any possible motion can also be reversed by changing the control. This then means that for any valid trajectory $\pi : [0,\ell(\pi)] \to \cspace$, the reverse trajectory $\ola{\pi} : [0,\ell(\pi)] \to \cspace$ where
\begin{align}
    \ola{\pi}(t) \defeq \pi(\ell(\pi)-t)
\end{align}
is also valid. This allows the system to maneuver indefinitely in arbitrarily small regions without having to leave and come back, and the DSTSP algorithm we discuss in this work depends on this ability; for an efficient nonsymmetric DSTSP algorithm, see \cite{adler-icra-16}.

\subsection{Control-affine dynamics} \label{sec:control-affine}

A special type of control system is the \emph{control-affine system}~\cite{lavalle-06} (also called \emph{affine-in-control}) in which the control set is $\cU \subseteq \bbR^m$ (so controls are $m$-dimensional real vectors) and the control function satisfies
\begin{align}
    \hcont(\bq,\bu) = \hcont_0(\bq) + \sum_{i=1}^m \hcont_i(\bq) u_i
\end{align}
where $\hcont_0, \hcont_1, \dots, \hcont_m$ are smooth vector fields on $\cspace$. The uncontrolled vector field $\hcont_0$ is called the \emph{drift}; if $\hcont_0(\bq) = \bzero$ (the trivial tangent vector at $\bq$) for all $\bq \in \cspace$, the system is symmetric, which for control-affine systems is also called \emph{driftless}. 

In some control-affine systems, $\hcont_0, \dots, \hcont_m$ span the tangent space at $\bq$, allowing linear combinations to move in any direction (subject to drift); however, the systems we are interested in typically do not have this property, making them \emph{nonholonomic} \cite{jean-14}. In this case, the direction of movement is restricted to the subspace of the tangent plane spanned by the vector fields $\hcont_i$.

Nevertheless, in symmetric control-affine systems, it is often possible to produce motion in a direction not directly allowed by making small forward and backward motions using commutators. Formally, the Lie algebra of the vector fields at $\bq$ yields a set of vectors in which small motions can be made. A control-affine system is called \emph{controllable} at $\bq$ if this Lie algebra spans the tangent space. In this case, for any $\varepsilon > 0$, the $\varepsilon$-ball in $d_\Pi$ around $\bq$ will contain $\bq$ in its interior.

\subsection{Metrics and distances}

Let $d_{\wspace}(\cdot,\cdot)$ and $d_{\cspace}(\cdot,\cdot)$ denote metrics on the manifolds $\wspace$ and $\cspace$. These induce norms $\norm{\cdot}_\wspace$ and $\norm{\cdot}_\cspace$ on the tangent spaces at any $x \in \wspace$ and $\bq \in \cspace$. 
When $\Pi$ is symmetric, this is a metric, but when $\Pi$ is not symmetric it is a \emph{quasimetric} since $d_\Pi(\bq,\bq') \neq d_\Pi(\bq',\bq)$ in general.

Note that the lengths of the tours are \emph{not} necessarily measured in the metrics $d_\wspace$ or $d_\cspace$ but rather by the time it takes to execute the tour -- these are not necessarily equivalent because different control inputs to the system might cause motions of different speeds relative to these metrics. We introduce these to aid in the analysis and in particular so that the assumption of a `speed limit' of the system (see \Cref{assm:speed-limits}) is well-defined.


\subsection{Reachability sets, small-time constraint factor, and agility function}

We first define the distance function induced by dynamics $\Pi$:
\begin{align}
    d_\Pi(\bq,\bq') = \inf_{\pi \in \Pi} (\ell(\pi) : \pi(0) = \bq \text{ and } \pi(\ell(\pi)) = \bq')
\end{align}
Note that the minimum time required to go from $\bq$ through $\bq'$ and end at $\bq''$ for any $\bq, \bq', \bq'' \in \cspace$ is $d_\Pi(\bq,\bq') + d_\Pi(\bq',\bq'')$. Furthermore, making a useful equivalent definition of the distance between $x,x' \in \wspace$ poses difficulties as the time required to go $x \to x' \to x''$ generally does not add in the same way it does for $\bq \to \bq' \to \bq''$ since the shortest valid path $x \to x'$ might require a much different configuration at $x'$ from the shortest valid path $x' \to x''$.

\begin{definition} \label{def:reachable-sets}
    Given a system $\Pi$, $\bq \in \cspace$ and $\varepsilon > 0$, the \emph{$\varepsilon$-reachable set} and the \emph{workspace $\varepsilon$-reachable set} from $\bq$ are defined as
    \begin{align}
        R_{\varepsilon}(\bq) \defeq \{\bq' \in \cspace : d_\Pi(\bq, \bq') \leq \varepsilon \} ~~\text{ and }~~ \bar{R}_{\varepsilon}(\bq) \defeq [R_{\varepsilon}(\bq)]_\wspace \, .
    \end{align}
    We refer to $\bq$ as the \emph{anchor} of $R_{\varepsilon}(\bq)$ and $\bar{R}_{\varepsilon}(\bq)$.
\end{definition}
A key aspect of our results is the volume of these reachable sets, particularly the workspace reachable sets, which determines in a sense how maneuverable the dynamic system is. In general, this volume will scale polynomially with $\varepsilon$, i.e. $\vol_\wspace(\bar{R}_\varepsilon(\bq)) \, \propto \, \varepsilon^\gamma$ for some appropriate $\gamma$, which we call the \emph{small-time constraint factor} (following \cite{itani-thesis-12}), and which will ultimately determine how the TSP tour length scales as the number $n$ of targets grows to $\infty$. Note that we are assuming that $\gamma$ is constant over the space. 

However, in order to obtain more precise bounds (particularly for the case of symmetric systems) we are also interested in the constant factor attached to $\varepsilon^\gamma$, which can vary depending on the configuration $\bq$ in question. We call this the \emph{configuration agility function} $g : \cspace \to \bbR_{> 0}$ and broadly speaking for small $\varepsilon > 0$ we have
\begin{align}
    \vol_\wspace(\bar{R}_\varepsilon(\bq)) \approx g(\bq) \varepsilon^\gamma
\end{align}
Formally we can define these as follows:
\begin{definition} \label{def:stcf-agility}
    The \emph{small-time constraint factor} of $\Pi$ is $\gamma$ such that at all $\bq$
    \begin{align}
        \gamma \defeq \lim_{\varepsilon \to 0} \frac{\log(\vol_\wspace(\bar{R}_\varepsilon(\bq))) }{\log(\varepsilon)}
    \end{align}
    and the \emph{agility function} $g : \cspace \to \bbR_{> 0}$ is
    \begin{align}
        g(\bq) \defeq \lim_{\varepsilon \to 0} \frac{\vol_\wspace(\bar{R}_\varepsilon(\bq))}{\varepsilon^\gamma} \,.
    \end{align}
\end{definition}
Note that the configuration agility function takes inputs from $\cspace$. Thus, we need to define the \emph{workspace agility function} which roughly measures the maximum possible agility the system can have when at a configuration projecting to $x \in \wspace$. This is important because if the system can be in many different configurations corresponding to $x \in \wspace$, it will generally be best to use the configuration maximizing the agility. For simplicity we will also denote this by $g$; whether $g$ refers to the configuration agility function or the workspace agility function can be determined by whether its input is in $\cspace$ or $\wspace$. We will use the term \emph{agility function} to refer to either the configuration or workspace agility function depending on context.
\begin{definition}
    The \emph{workspace agility function} is $g :\wspace \to \bbR_{> 0}$ defined as
    \begin{align}
        g(x) \defeq \sup_{\bq \in \projinv{x}} g(\bq)
    \end{align}
\end{definition}

\subsection{The Dynamic TSP and Dynamic Stochastic TSP}

The Dynamic TSP (DTSP) is then the following problem: given control system $\Pi$ on $\cspace$, and given $n$ targets $x_1, \dots, x_n \in \wspace$, find
\begin{align}
    \tsp_{\Pi}(\{x_i\}) \defeq \tsp_{\Pi}(x_1, \dots, x_n) \defeq \inf_{\pi \in \Pi} (\ell(\pi) : x_i \in \bar{\pi} \text{ for all } i)
\end{align}
that is, for the shortest valid path that passes through all the targets. We refer to any trajectory which visits all the target points $x_1, \dots, x_n$ as a \emph{TSP trajectory}.

The Dynamic Stochastic TSP (DSTSP) is the variant of the above problem in which the targets $X_1, \dots, X_n$ are randomly chosen; in this work we are concerned with $X_1, \dots, X_n$ chosen independently and identically distributed (iid) according to some distribution $f$ over $\wspace$; we write this as $X_i \simt{iid} f$. While $f$ can in principle be any probability distribution, we will be concerned with continuous probability distributions where $f$ represents a probability density function. Since $f$ represents a continuous probability distribution, it satisfies $f : \wspace \to \bbR_{\geq 0}$ such that for any subset $\wspace^* \subseteq \wspace$, we have $\bbE_{X \sim f} [X \in \wspace^*] = \int_{\wspace^*} f(x) \, dx$. 
We denote the support of $f$ as
\begin{align}
    \wspacef \defeq \{x \in \wspace : f(x) > 0\}
\end{align}
which we will assume is bounded (\Cref{assm:density-fn}).

\section{Main Results} \label{sec:results}

While our main results depend on a number of conditions, for clarity we will state the results first. The conditions are given in \Cref{sec:assumptions}. 

In the DSTSP, the length of the optimal TSP trajectory is a random variable dependent on the placement of the targets $X_i \simt{iid} f$, where $f$ is a probability density function on the workspace. Thus, our results take the form of bounds on $\tsp_\Pi(X_1, \dots, X_n)$ which hold with high probability as $n \to \infty$. We not only want to show that the probability that these bounds fail goes to $0$ as $n \to \infty$, we want to show that it approaches $0$ rapidly:

\begin{definition} \label{def:wvhp-def-1}
Let $\{A_n\}_{n \in \bbZ_{\geq 0}}$ be an infinite sequence of events parameterized by an integer $n$. Then we say that $A_n$ happens \emph{with very high probability} if there are constants $c_1, c_2, c_3 > 0$ such that, for all sufficiently large $n$,
\begin{align}
	\bbP[A_n] \geq 1 - c_1 e^{-c_2 n^{c_3}}  
\end{align}
\end{definition}
This implies that $\lim_{n \to \infty} \bbP[A_n] = 1$, and converges faster than any inverse polynomial (depending on definition it can even be said to converge to $1$ `exponentially'). The rapid convergence rate, aside from being of interest itself, allows bounds on the expected tour length to follow immediately from our probabilistic bounds.

\subsection{Parameters}

We first define some parameters needed to state our theorems. Since the formal definitions can be fairly involved, we will sketch the meaning of certain parameters which will be formally defined later in \Cref{sec:assumptions}; wherever we do we will refer to the specific place it is defined.

First, recall that $\gamma$ is the small time constraint factor and $g(x)$ is the agility function (\Cref{def:stcf-agility}), satisfying the rough relation
\begin{align}
    \sup_{\bq \in \projinv{x}} \vol_\wspace (\bar{R}_\varepsilon(\bq)) \approx g(x) \varepsilon^\gamma
\end{align}
Then we have the integer \emph{branching factor} $b$ which generally denotes how many $\varepsilon$-configuration reachable sets it takes to cover a $(2\varepsilon)$-configuration reachable set (see \Cref{assm:b-coverability}). We then use this to define a parameter $\beta$ which will be used in our lower bound:
\begin{definition} \label{def:beta}
        Let $b$ be the branching factor and $\gamma$ be the small-time constraint factor of $\Pi$, and $r$ be a constant where $r = 2$ when $\Pi$ is nonsymmetric and $r = 3/2$ when $\Pi$ is symmetric. Then we let the \emph{lower constant} be:
    \begin{align} \label{eq:xi2}
        \beta = (1 + \xi) r^\gamma ~~\text{ where }~~ \xi = \begin{cases} 3(\log(b)/r^\gamma) &\text{if } \log(b) > r^\gamma \\ 3 \sqrt{\log(b)/r^\gamma} &\text{if } \log(b) \leq r^\gamma \end{cases}
    \end{align}
\end{definition}
Note that $\beta = (1+\xi)2^\gamma \, \propto \, \max(\log(b),r^\gamma)$.

For the upper bound, we need to consider \emph{hierarchical cell structures} (HCS) (see \Cref{def:hcs}), which we sketch here. A hierarchical cell structure is a recursively-defined structure of nested cells, having an integer \emph{scaling parameter} $s$ and an \emph{efficiency parameter} $\alpha \leq 1$. An HCS at scale $\varepsilon$ rooted at some $\bq \in \cspace$ is defined by a cell contained in some $\bar{R}_\varepsilon(\bq)$ whose volume is at least $\alpha g(\proj{\bq}) \varepsilon^\gamma$ (roughly speaking, as there is an additional approximation term in the formal definition), which can covered by $s^\gamma$ HCS's at scale $\varepsilon/s$ (the next `level' down). The scaling parameter $s$ thus denotes how much smaller the HCS's get when going down a level, and the efficiency parameter $\alpha$ denotes how large the cells are relative to the largest reachable sets at the same scale.

Typically, HCS's can be constructed with scaling factor $s = 2$; for instance, control-affine systems satisfying common regularity conditions have HCS's with $s=2$ (see \Cref{lem:existence-of-hcs} and Appendix~\ref{appx:meta-assumption}). However, we make a more general definition because it can be extended to the Euclidean TSP with target points distributed on a set of fractal dimension, which in certain cases (such as the Menger Sponge) are naturally scaled using $s \neq 2$; see Appendix~\ref{appx:fractal-targets} for the relevant results.

\subsection{Main concentration bounds}

We first state a trivial, non-probabilistic bound which follows from \Cref{assm:not-inf-distance}:
\begin{proposition} \label{prop:dtsp-trivial-bound}
    For some constant $C > 0$, for all $x_1, x_2, \dots, x_n \in \wspacef$,
    \begin{align}
        \tsp_\Pi(x_1, \dots, x_n) \leq Cn \,.
    \end{align}
\end{proposition}

\begin{proof}
    For each $x_i$, we select (arbitrarily) some $\bq_i$ such that $\proj{\bq_i} = x_i$; then by \Cref{assm:not-inf-distance}, there is some $C$ such that $d_\Pi(\bq_i, \bq_{i+1}) \leq C$ for all $i = 1, \dots, n-1$, and hence we have a tour which goes through $\bq_1, \bq_2, \dots, \bq_n$ by the order of indices with length $\leq C(n-1)$.
\end{proof}

This will be useful for showing that our probabilistic bounds from \Cref{thm:dstsp-bounds} imply equivalent bounds in expectation (\Cref{cor:dstsp-expectation}).

We now state the main theorem, which gives a very high probability lower bound to the DSTSP, and a matching (up to a constant in $n$ and $f$) upper bound when $\Pi$ is symmetric. 
The lower bound uses the constant $\beta$, which in turn is based on the branching factor $b$ (and the small-time constraint factor $\gamma$); the upper bound, being derived from an algorithm reliant on hierarchical cell structures, uses $s$ and $\alpha$. Both upper and lower bounds feature the growth rate $n^{1-\frac{1}{\gamma}}$ and a term which demonstrates how density $f$ and agility $g$ interact to affect the tour length.
\begin{theorem} \label{thm:dstsp-bounds} 
    If the assumptions in \Cref{sec:assumptions} hold, then for any $\delta > 0$, the following holds for sufficiently large $n$:
    \begin{align} \label{eq:dstsp-bounds-lower}
        \tsp_{\Pi}(\{X_i\}) \geq (1-\delta) \beta^{-1}  n^{1-\frac{1}{\gamma}} \int_{\wspacef} f(x)^{1-\frac{1}{\gamma}} g(x)^{-\frac{1}{\gamma}} \, dx ~~~~ \wvhp
    \end{align}
    where $\beta$ is the lower constant (see \Cref{def:beta}).

    Furthermore, if $\Pi$ is symmetric and $\wspacef$ can be covered with hierarchical cell structures with scaling parameter $s \geq 2$ and efficiency parameter $\alpha \leq 1$ (see \Cref{def:hcs}), then for any $\delta > 0$ the following holds for sufficiently large $n$:
    \begin{align} \label{eq:dstsp-bounds-upper}
        \tsp_{\Pi}(\{X_i\}) \leq (1+\delta) \big(12 s \alpha^{-\frac{1}{\gamma}}\big)n^{1-\frac{1}{\gamma}} \int_{\wspacef} f(x)^{1-\frac{1}{\gamma}} g(x)^{-\frac{1}{\gamma}} \, dx ~~~~ \wvhp
    \end{align}
\end{theorem}

These results extend to the expected value (see Appendix~\ref{appx:wvhp} for the proof):
\begin{corollary} \label{cor:dstsp-expectation}
     If the assumptions in \Cref{sec:assumptions} hold, then for any $\delta > 0$,
     \begin{align}
         \bbE_{X_i \simt{iid} f}[\tsp_\Pi(\{X_i\})] \geq (1-\delta) \beta^{-1}  n^{1-\frac{1}{\gamma}} \int_{\wspacef} f(x)^{1-\frac{1}{\gamma}} g(x)^{-\frac{1}{\gamma}} \, dx 
     \end{align}
     for all sufficiently large $n$. Furthermore, if $\Pi$ is symmetric and $\wspacef$ can be covered with hierarchical cell structures with scaling parameter $s \geq 2$ and efficiency parameter $\alpha$ (see \Cref{def:hcs}), then for any $\delta > 0$,
    \begin{align} \label{eq:dstsp-bounds-upper}
        \bbE_{X_i \simt{iid} f}[\tsp_{\Pi}(\{X_i\})] \leq (1+\delta) \big(12 s \alpha^{-\frac{1}{\gamma}}\big)n^{1-\frac{1}{\gamma}} \int_{\wspacef} f(x)^{1-\frac{1}{\gamma}} g(x)^{-\frac{1}{\gamma}} \, dx
    \end{align}
    for all sufficiently large $n$.
\end{corollary}
Taken together, these results yield the general rule that for symmetric $\Pi$,
\begin{align} \label{eq:general-form-bound}
    \tsp_\Pi(\{X_i\}) = \Theta \Big(n^{1-\frac{1}{\gamma}} \int_{\wspacef} f(x)^{1-\frac{1}{\gamma}} g(x)^{-\frac{1}{\gamma}} \, dx \Big)
\end{align}
and the multiplicative gap between the upper and lower bounds is (up to the approximation factor)
\begin{align} \label{eq:main-growth}
    \cgap \defeq \cgap(\Pi) = 12 s \alpha^{-\frac{1}{\gamma}} \beta
\end{align}
where $\beta \, \propto \, \max(\log(b), (3/2)^\gamma)$ (since $\Pi$ is symmetric).

Note that this gap is a constant with regard to the density function $f$, since none of the parameters depend on $f$. Therefore, \eqref{eq:main-growth} can be viewed as a statement on how the TSP tour length varies as $f$ is changed (provided it still satisfies the assumptions given in \Cref{sec:assumptions}). While $\cgap$ has no direct dependence on $g$, both $g$ and $\cgap$ depend on $\Pi$; however, the integral $\int_{\wspacef} f(x)^{1-\frac{1}{\gamma}} g(x)^{-\frac{1}{\gamma}} \, dx$ is still a good indicator of how fast a system governed by $\Pi$ can visit many targets distributed iid by $f$.

\begin{remark} \label{rem:fractal-targets}
    Our techniques also allow us to achieve similar results to \Cref{thm:dstsp-bounds} for the case of the Euclidean TSP and targets distributed on a set of fractal dimension; see Appendix~\ref{appx:fractal-targets}.
\end{remark}

\begin{remark} \label{rem:nonsymmetric-dstsp-ub}
    When $\Pi$ is nonsymmetric, an upper bound of growth rate $\Theta(n^{1-\frac{1}{\gamma}})$ holds with very high probability; see \cite{adler-icra-16}. However, the dependence of the constant factor on the probability density function $f$ cannot be cleanly characterized in the same way as in \eqref{eq:general-form-bound}. 
\end{remark}

\subsection{DSTSP lower bound}

We now state more precise versions of the bounds from \Cref{thm:dstsp-bounds}, in particular giving the convergence rates.

\begin{proposition} \label{prop:dstsp-lower-bound}
    Let the assumptions in \Cref{sec:assumptions} hold, and let
    \begin{align}
        v_{f,g} &= \var_{f \sim X}\big[f(X)^{-\frac{1}{\gamma}} g(X)^{-\frac{1}{\gamma}} \big] 
        \\ &= \int_{\wspacef} f(x)^{1-\frac{2}{\gamma}} g(x)^{-\frac{2}{\gamma}} \, dx - \bigg(\int_{\wspacef} f(x)^{1-\frac{1}{\gamma}} g(x)^{-\frac{1}{\gamma}} \, dx \bigg)^2
    \end{align}
    Then for any $\delta > 0$, there is some $\lambda > 0$ such that for sufficiently large $n$,
    \begin{align}
    \bbP \Bigg[&\tsp_{\Pi}(\{X_i\}) \geq (1-\delta) \beta^{-1}  n^{1-\frac{1}{\gamma}} \int_{\wspacef} f(x)^{1-\frac{1}{\gamma}} g(x)^{-\frac{1}{\gamma}} \, dx \Bigg] 
    \\ \geq 1 &- \exp\bigg({-\frac{4}{5} \frac{\log(b) \lambda n^{\frac{1}{\gamma}}}{1-\delta/2}}\bigg) 
    \\ &- \exp\bigg({-\frac{n (1-\delta/2) \big(\int_{\wspacef} f(x)^{1-\frac{1}{\gamma}} g(x)^{-\frac{1}{\gamma}} \, dx \big)^2 \delta^2/32}{(1+\delta/2)v_{f,g} + (1-\delta/2) \big(\int_{\wspacef} f(x)^{1-\frac{1}{\gamma}} g(x)^{-\frac{1}{\gamma}} \, dx \big)^2 \delta/12}}\bigg)
\end{align}
\end{proposition}

\begin{proposition} \label{prop:dstsp-upper-bound}
If $\Pi$ is symmetric with hierarchical cell structure with scaling parameter $s \geq 2$ and efficiency parameter $\alpha$, for any $\delta > 0$: if $\gamma \geq 3$,
    \begin{align} \label{eq:dstsp-bounds-upper-01}
        \bbP\bigg[\tsp_{\Pi}(\{X_i\}) \leq (1+\delta) \big(12 s \alpha^{-\frac{1}{\gamma}}\big)n^{1-\frac{1}{\gamma}} \int_{\wspacef} f(x)^{1-\frac{1}{\gamma}} g(x)^{-\frac{1}{\gamma}} \, dx\bigg]
        \\ \geq 1 - e^{-\frac{1}{2} n^{1 - 2/\gamma}} \label{eq:dstsp-bound-upper-01}
        \\ \geq 1 - e^{-\frac{1}{2} n^{1/3}} \label{eq:dstsp-bound-upper-02}
    \end{align}
    for all sufficiently large $n$. For a more precise bound, let $\bp = (p_1,\dots,p_m)$ be the probability masses of the $m$ cells in the hierarchical cell structure. Then
    \begin{align} \label{eq:dstsp-bounds-upper-01}
        \bbP\bigg[\tsp_{\Pi}(\{X_i\}) \leq (1+\delta) \big(12 s \alpha^{-\frac{1}{\gamma}}\big)n^{1-\frac{1}{\gamma}} \int_{\wspacef} f(x)^{1-\frac{1}{\gamma}} g(x)^{-\frac{1}{\gamma}} \, dx\bigg]
        \\ \geq 1 - e^{-\frac{1}{2} n^{1 - 2/\gamma}(\sum_{j=1}^m p_j^{1-1/\gamma})^2} \label{eq:dstsp-bound-upper-03}
    \end{align}
    for all sufficiently large $n$. Note that $1 \leq \sum_{j=1}^m p_j^{1-1/\gamma} \leq m^{\frac{1}{\gamma}}$, and is maximized when all $m$ cells in the hierarchical structure have mass $= 1/m$. 
    
    If $\gamma = 2$, and (wlog) $p_1$ is the smallest nonzero value of $\bp = (p_1, \dots, p_m)$, then for all sufficiently large $n$,
    \begin{align}
        \bbP\bigg[\tsp_{\Pi}(\{X_i\}) \leq (1+\delta) \big(12 s \alpha^{-\frac{1}{\gamma}}\big)n^{1-\frac{1}{\gamma}} \int_{\wspacef} f(x)^{1-\frac{1}{\gamma}} g(x)^{-\frac{1}{\gamma}} \, dx\bigg]
        \\ \geq 1 - e^{- \frac{(2/9) p_1 n (\sum_{j=1}^m p_j^{1/2})^2 }{127 - \log(1/p_1) + \log(n) }}
    \end{align}
\end{proposition}

\begin{remark}
    Note that while the probability bound for $\gamma = 2$ converges to $1$ faster in the limit, in the sense that it is $1 - e^{-\Theta(n/\log(n))}$ as opposed to $1 - e^{-\Theta(n^{1-2/\gamma})}$, the $\gamma \geq 3$ bound avoids the $p_1$ term, which can be very small and can make it converge slower until $n$ is very small. However, for $\gamma \geq 3$, we can get even faster convergence of $1 - e^{-\Theta(n)}$; but the probability of failure includes some terms which potentially stay relatively large until $n$ is extremely big.
\end{remark}

\subsection{DTSP with adversarial targets}

\Cref{thm:dstsp-bounds} naturally leads to the following questions:
\begin{itemize}
    \item Given a control system $\Pi$ with agility function $g$ and some bounded $\wspace^* \subseteq \wspace$, what probability density function $f$ with support (contained in) $\wspace^*$ makes the TSP trajectory on $X_i \simt{iid} f$ the longest as $n \to \infty$?
    \item How does this `worst-case density function' compare to adversarial target points (i.e. chosen to maximize the length of the optimal TSP trajectory)?
\end{itemize}
We will show that the difference between these (both in expectation and with very high probability) is at most a constant factor which depends on certain characteristics of the dynamic constraints but not on the agility function $g$, resulting in the bound:

\begin{theorem} \label{thm:adv-dtsp-bounds}
    Let $\Pi$ be dynamic constraints satisfying the assumptions in \Cref{sec:assumptions} and $\wspace^* \subseteq \wspace$ be bounded. Then for any $\delta > 0$, for any sufficiently large $n$,
    \begin{align} \label{eq:adv-dtsp-bound-lower}
        \sup_{X_1,\dots,X_n \in \wspace^*} \big(\tsp_{\Pi}(\{X_i\})\big) \geq (1-\delta) \beta^{-1} n^{1-\frac{1}{\gamma}} \Big(\int_{\wspace^*} g(x)^{-1} \, dx\Big)^{\frac{1}{\gamma}}
    \end{align}
    If $\Pi$ is symmetric with a hierarchical cell structure with scaling parameter $s \geq 2$ and efficiency parameter $\alpha$, then for any $\delta > 0$,
    \begin{align} \label{eq:adv-dtsp-bound-upper}
        \sup_{X_1,\dots,X_n \in \wspace^*} \big(\tsp_{\Pi}(\{X_i\})\big) \leq (1+\delta) \big(6 s \alpha^{-\frac{1}{\gamma}}\big)n^{1-\frac{1}{\gamma}} \Big(\int_{\wspace^*} g(x)^{-1} \, dx\Big)^{\frac{1}{\gamma}}
    \end{align}
    for any sufficiently large $n$.
    
    This means that for symmetric $\Pi$ and any $\delta > 0$,
    \begin{align}
        (1-\delta) \beta^{-1} \leq \frac{\sup_{X_1, \dots, X_n \in \wspace^*} \tsp_\Pi(X_1,\dots,X_n)}{n^{1-\frac{1}{\gamma}} (\int_{\wspace^*} g(x)^{-1} \, dx)^{\frac{1}{\gamma}}} \leq (1+\delta) 6 s \alpha^{-\frac{1}{\gamma}}
    \end{align}
    for any sufficiently large $n$
\end{theorem}
Note that these are not probabilistic bounds since $X_1, \dots, X_n$ are not random anymore. The lower bound is directly derived from the `worst-case target distribution' $f_g(x) \, \propto \, g(x)^{-1}$ or, to normalize,
\begin{align}
    f_g(x) = \frac{g(x)^{-1}}{\int_{\wspace^*} g(y)^{-1} \, dy} \,.
\end{align}
The fact that the given lower bound holds at all (let alone with very high probability) when $X_1, \dots, X_n \simt{iid} f_g$ then implies that a non-random adversary can choose a set of target points which makes the shortest tour at least this large. The upper bound derives from our algorithm which covers the space using Hierarchical Cell Structures.

\begin{remark}
    The DSTSP bound's dependence on $f,g$ takes the form of an integral over $\wspacef$ of a function of $f,g$; this is because in some sense (when $n \to \infty$) the length of the TSP depends only on local conditions. Even if two regions are close, with enough target density the efficiency of visiting targets in each becomes roughly independent. However, the Adversarial DTSP has an exponent outside the integral. This is because varying $g$ locally will change where the targets go, and hence change the target density over the whole space.
\end{remark}

\begin{remark}
    One odd feature of this theorem is that the adversarial upper bound (equation \eqref{eq:adv-dtsp-bound-upper}) has a better absolute constant than the equivalent stochastic bound (equation \eqref{eq:dstsp-bounds-upper}), i.e. $6$ as opposed to $12$. This is because in the adversarial case we get a hard (deterministic) bound, but in the random case when $f$ does not maximize the integral, an `unlucky' random placement of targets might cause the tour to be larger than it was expected to be. To get our probabilistic bounds we multiply by a buffer constant which was not needed in the adversarial case.
    

\end{remark}

\section{Assumptions} \label{sec:assumptions}


Our main results follow from a few basic assumptions about the properties of the control system $\Pi$ at very small scales. Our assumptions largely follow from a key meta-assumption:
\begin{metaassumption} \label{meta:regularity}
    Our dynamics $\Pi$ are \emph{control-affine} and \emph{equiregular} (\cite{jean-14}, Definition 2.10) over $\projinv{\wspacef}$, which is contained in some compact $\cspace^* \subseteq \cspace$.
\end{metaassumption}
For symmetric control-affine dynamics, this yields important tools such as the Chow-Rashevskii theorem and especially the Uniform Ball-Box Theorem (\cite{jean-14}, Thm 2.4), which show that reachable sets around any $\bq \in \cspace$ always contain a rectilinear neighborhood of a certain size around $\bq$. We will note which assumptions follow from the meta-assumption and which have to be made on their own (in particular, any assumption concerning the distribution $f$ of the target points has nothing to do with the meta-assumption); for more details see Appendix~\ref{appx:meta-assumption}.

\begin{remark}
    Our results follow if the assumptions in this section are met, even if the meta-assumption itself does not hold.
\end{remark}

\subsection{$\zeta$-regularized approximations}

To show our results, we will often want to assume that certain functions are Lipschitz continuous or bounded away from $0$ or $\infty$, or both. However, in many important cases this may not hold. Thus, we want to use approximations of these functions that do have these properties, which we call \emph{$\zeta$-regularized approximations}; $\zeta > 0$ governs the degree of approximation and the Lipschitz constant and bound away from $0$ or $\infty$. Depending on whether we want our approximation to be an upper or lower bound, we define the \emph{upper} and \emph{lower} $\zeta$-regularized approximations.
\begin{definition} \label{def:alpha-reg-appx}
For any function $h$ and $\zeta > 0$ on domain $\cY$ with metric $d_\cY$, let
\begin{align}
    \widehat{\cF}(h,\zeta) &\defeq \{h^* : h^*(y) \geq \max(h(y),\zeta), \text{ and } h^* \text{ is } (1/\zeta)\text{-Lipschitz continuous} \}
    \\ \text{and }~~ \widecheck{\cF}(h,\zeta) &\defeq \{h^* : h^*(y) \leq \min(h(y),1/\zeta), \text{ and } h^* \text{ is } (1/\zeta)\text{-Lipschitz continuous} \} \,.
\end{align}
We define the \emph{$\zeta$-regularized upper and lower approximations} $\hat{h}^{(\zeta)}$ and $\check{h}^{(\zeta)}$ of $h$ respectively as
\begin{align}
    \hat{h}^{(\zeta)}(y) &\defeq \inf_{h^* \in \widehat{\cF}(h,\zeta)} (h^*(y))
    \\ \text{and }~~ \check{h}^{(\zeta)}(y) &\defeq \sup_{h^* \in \widecheck{\cF}(h,\zeta)} (h^*(y)) \,.
\end{align}
\end{definition}

We show that $\hat{h}^{(\zeta)}$ and $\check{h}^{(\zeta)}$ have desirable properties while not being far from $h$:
\begin{lemma}
    For any nonnegative $h$ which is continuous almost everywhere and bounded above by some finite $h_{\max} = \sup_y h(y)$ on domain $\cY$, the following hold for the upper $\zeta$-regularized function $\hat{h}^{(\zeta)}$:
    \begin{enumerate}[i.]
        \item $\hat{h}^{(\zeta)}(y) \geq h(y)$ and $\hat{h}^{(\zeta)}(y) \geq \zeta$ for all $y \in \cY$.
        \item $\lim_{\zeta \to 0} \hat{h}^{(\zeta)}(y) = h(y)$ almost everywhere.
        \item $\hat{h}^{(\zeta)}(y)$ is decreasing for any fixed $y$ as a function of $\zeta$.
        \item $\hat{h}^{(\zeta)}(y)$ is $(1/\zeta)$-Lipschitz continuous.
        \item If $h$ has finite integral on a bounded set $\cA$,
        \begin{align}
            \lim_{\zeta \to 0} \int_{\cA} \hat{h}^{(\zeta)}(x) \, dx = \int_{\cA} h(x) \, dx
        \end{align}
        \item For any $\zeta \leq h_{\max}$, we have $\max_y \hat{h}^{(\zeta)}(y) \leq h_{\max}$. (Or, in other words, $\sup_y \hat{h}^{(\zeta)}(y) = \max(\zeta,h_{\max})$).
    \end{enumerate}
    Analogous results hold for the lower $\zeta$-regularized function $\check{h}^{(\zeta)}$.
\end{lemma}

\begin{proof}
We show these as follows:

\emph{i:} This follows from the definition of $\hat{h}^{(\zeta)}(y)$.

\emph{ii:} This follows since it holds at all $y$ at which $h$ is continuous, which is almost everywhere.

\emph{iii:} This follows since $\cF(h,\zeta') \subseteq \cF(h,\zeta)$ if $\zeta' > \zeta$, and therefore an infimum over values in $\cF(h,\zeta)$ will be smaller than the same infimum over values in $\cF(h,\zeta')$.

\emph{iv.} This follows because the infimum of a set of $c$-Lipschitz continuous functions is also $c$-Lipschitz continuous. This is because of the following argument by contradiction. Suppose $\hat{h}^{(\zeta)}$ is not $(1/\zeta)$-Lipschitz continuous; then there is $y_1, y_2 \in \cY$ such that
\begin{align}
    |\hat{h}^{(\zeta)}(y_2) - \hat{h}^{(\zeta)}(y_1)| > (1/\zeta)d_{\wspace}(y_1, y_2) \,.
\end{align}
WLOG let $\hat{h}^{(\zeta)}(y_1) < \hat{h}^{(\zeta)}(y_2)$. Then there is some $h^* \in \cF(h,\zeta)$ such that
\begin{align}
    h^*(y_1) - \hat{h}^{(\zeta)}(y_1) &< |\hat{h}^{(\zeta)}(y_1) - \hat{h}^{(\zeta)}(y_2)| - (1/\zeta)d_{\cY}(y_1, y_2)
    \\ \implies h^*(y_2) - h^*(y_1) &\geq \hat{h}^{(\zeta)}(y_2) - h^*(y_1)
    \\ &> (\hat{h}^{(\zeta)}(y_2) - \hat{h}^{(\zeta)}(y_1)) - \big(|\hat{h}^{(\zeta)}(y_1) - \hat{h}^{(\zeta)}(y_2)| - (1/\zeta)d_{\cY}(y_1, y_2) \big)
    \\ &= (1/\zeta)d_{\cY}(y_1, y_2)
\end{align}
so $h^*$ is also not $(1/\zeta)$-Lipschitz continuous, which is a contradiction since $h^* \in \cF(h,\zeta)$.

\emph{v.} This follows from (ii) and the Dominated Convergence Theorem (where the dominating function is $f_{\max}$ over $\wspacef$, which is integrable since by the conditions on $f$, $\wspacef$ is bounded).

\emph{vi.} This follows because the constant function with value $\max(\zeta,h_{\max})$ is an upper bound to $h$, at least as large as $\zeta$ and Lipschitz continuous with any parameter (since it is constant), and hence is in $\widehat{\cF}(h,\zeta)$. Then by definition $\hat{h}^{(\zeta)}(y)$ must be at most $\max(\zeta,h_{\max})$ (and since $\hat{h}^{(\zeta)}(y) \geq \max(\zeta,h(y))$ for all $y$, this means $\sup_y \hat{h}^{(\zeta)}(y) = \max(\zeta,h_{\max})$).

Proofs for the lower $\zeta$-regularized function are analogous.
\end{proof}

The upper $\zeta$-regularized approximation will be important to showing the TSP lower bound (in general we can show a lower bound by assuming that the system is more agile than it really is, i.e. using an upper bound of $g$ in place of $g$ itself keeps the DSTSP lower bound valid) while the lower $\zeta$-regularized approximation will be important to showing the TSP upper bound.

\subsection{Assumptions about the target distribution}

We consider the support $\wspacef$ of density $f$ and how $f$ behaves on it.
\begin{assumption} \label{assm:density-fn}
    The function $f$ is a probability density function represents a full-dimensional continuous random variable, i.e. $f : \wspace \to \bbR_{\geq 0}$ is integrable, continuous almost everywhere, and satisfies $\int_\wspace f(x) \, dx = 1$. Then for any subset $\cA \subseteq \wspace$,
    \begin{align}
        \bbP_{X \sim f}[X \in \cA] = \int_{\cA} f(x) \, dx \,.
    \end{align}
    Furthermore, $f$ has a finite maximum value $f_{\max} \defeq \sup_{x \in \wspace} f(x) < \infty$.

    The target region $\wspacef = \{x : f(x) > 0\}$ is also compact.
\end{assumption}
Except for very contrived examples, assuming that $\wspacef$ is compact is really just to ensure it is bounded. If $\wspacef$ is allowed to be unbounded, it is easy to produce target point distributions $f$ for which the expected distance between $X_1, X_2 \simt{iid} f$ is infinite by using the St. Petersburg paradox: pick an origin point $\bzero \in \wspace$ and set $f$ to have support on a sequence of small regions $\wspace_1, \wspace_2, \dots \subset \wspace$, where $\bbP_{X \sim f}[X \in \wspace_i] = 2^{-i}$ but $d_\wspace(\bzero,X) \approx 2^i$ for all $X \in \wspace_i$; then a random $X \sim f$ will have infinite expected distance from any point in $\wspace$, so $X_2$ will always have expected infinite distance from $X_1$. The length of the TSP tour will then be infinite in expectation even for $2$ target points, let alone $n > 2$.

\subsection{Assumptions about the dynamics}

We now consider some assumptions we need to make about the dynamics $\Pi$. Let $d_{\wspace}(\cdot,\cdot)$ and $d_{\cspace}(\cdot,\cdot)$ denote metrics on the manifolds $\wspace$ and $\cspace$. These induce norms $\norm{\cdot}_\wspace$ and $\norm{\cdot}_\cspace$ on the tangent spaces at any $x \in \wspace$ and $\bq \in \cspace$.

Note that the lengths of the tours are \emph{not} necessarily measured in the metrics $d_\wspace$ or $d_\cspace$ but rather by the time it takes to execute the tour, i.e. by $d_\Pi$ -- these are not necessarily equivalent because different control inputs to the system might cause motions of different speeds relative to these metrics. We introduce these to aid in the analysis and in particular to make the following assumption well-defined:

\begin{assumption} \label{assm:x-not-one-dim}
    The manifold $\wspace$ has dimension $\geq 2$.
\end{assumption}
This is to avoid degenerate cases. Typically, when $\wspace$ is $1$-dimensional, the TSP tour length is constant since one can simply sweep the entire space.

\begin{assumption} \label{assm:speed-limits}
    There is some `speed limit' $c_\Pi$ of $\Pi$ such that for any $\bq \in \cspace$ and control $\bu \in \cU$, we have $\norm{\hcont(\bq,\bu)}_\wspace \leq c_\Pi$.
\end{assumption}

Note that \Cref{assm:x-not-one-dim} and \Cref{assm:speed-limits} together imply the following:
\begin{lemma}
    The small-time constraint factor $\gamma$ is at least $2$.
\end{lemma}

\begin{proof}
    Let $\cB_\varepsilon(x)$ denote the radius-$\varepsilon$ ball around $x \in \wspace$ under metric $d_{\wspace}$. Then
    \begin{align}
        \bar{R}_\varepsilon(\bq) &\subseteq \cB_{c_\Pi \varepsilon}(\proj{\bq})
        \\ \implies \vol_\wspace(\bar{R}_\varepsilon(\bq)) &\leq \vol_\wspace(\cB_{c_\Pi \varepsilon}(\proj{\bq}))
    \end{align}
    and we know that $\vol_\wspace(\cB_{c_\Pi \varepsilon}(\proj{\bq})) \, \propto \, \varepsilon^{\mathrm{dim}(\wspace)}$ for small $\varepsilon$. Thus, since $\log(\varepsilon) < 0$,
    \begin{align}
        \gamma = \lim_{\varepsilon \to 0} \frac{\log(\vol_\wspace(\bar{R}_\varepsilon(\bq)))}{\log(\varepsilon)} \geq \frac{\log(\vol_\wspace(\cB_{c_\Pi \varepsilon}(\proj{\bq})))}{\log(\varepsilon)} = \mathrm{dim}(\wspace) \geq 2 \,.
    \end{align}
\end{proof}

We also assume that any two configurations are connected by a finite path $\pi \in \Pi$:
\begin{assumption} \label{assm:not-inf-distance}
    For any $\bq,\bq' \in \cspace$, there is a valid $\pi \in \Pi$ with finite $\ell(\pi)$ such that $\pi(0) = \bq$ and $\pi(\ell(\pi)) = \bq'$. 
\end{assumption}
In other words, $d_{\Pi}(\bq,\bq') < \infty$. For symmetric equiregular control-affine dynamics satisfying Chow's condition~\cite{jean-14}, this assumption follows from the Chow-Rashevskii theorem. However, for nonsymmetric control-affine dynamics it must be explicitly assumed since there are simple examples in which it doesn't hold (e.g. a boat swept downstream by a current that can't return upstream).\footnote{To apply \Cref{thm:dstsp-bounds} for nonsymmetric systems, \Cref{assm:not-inf-distance} must be shown to hold explicitly.}

A key condition for our results to hold is that the agility function be well defined and that the limit it represents converges uniformly over the space:
\begin{assumption} \label{assm:reachability-volume}
The control system $\Pi$ has an \emph{agility function} $g: \cspace \to \bbR_{\geq 0}$ and a \emph{small-time reachability factor} $\gamma$ such that for any $\rho > 0$, there is some $\varepsilon^*_\rho > 0$ such that for all $0 < \varepsilon \leq \varepsilon^*_\rho$ and all $\bq \in \cspace$,
\begin{align}
    (1-\rho) g(\bq) \leq \frac{\vol_{\wspace}(\bar{R}_\varepsilon(\bq))}{\varepsilon^\gamma} \leq (1+\rho) g(\bq) 
\end{align}
\end{assumption}
Note that this assumption implies a similar bound on the workspace agility function:
\begin{lemma}
    For any $\rho > 0$, there is some $\varepsilon^*_\rho > 0$ such that for all $0 < \varepsilon \leq \varepsilon^*_\rho$ and all $x \in \wspace$,
\begin{align}
    (1-\rho) g(x) \leq \sup_{\bq : \proj{\bq} = x} \frac{\vol_{\wspace}(\bar{R}_\varepsilon(\bq))}{\varepsilon^\gamma} \leq (1+\rho) g(x) 
\end{align}
\end{lemma}

\begin{proof}
    We let $\varepsilon^*_\rho > 0$ be the same value as in \Cref{assm:reachability-volume}. Then, for any $\bq$ such that $\proj{\bq} = x$, we have
    \begin{align}
        \frac{\vol_{\wspace}(\bar{R}_\varepsilon(\bq))}{\varepsilon^\gamma} \leq (1+\rho) g(\bq) \leq (1+\rho) g(x)
    \end{align}
    Additionally, we have
    \begin{align}
        \sup_{\bq : \proj{\bq} = x} \frac{\vol_{\wspace}(\bar{R}_\varepsilon(\bq))}{\varepsilon^\gamma} \geq \sup_{\bq : \proj{\bq} = x} (1-\rho)g(\bq) = (1-\rho)g(x)
    \end{align}
    and we have shown both sides of the inequality.
\end{proof}
We also assume bounds on $g$ over the workspace: 
\begin{assumption}
    Letting $g_{\min} \defeq \inf_{x \in \wspace} g(x)$ and $g_{\max} \defeq \sup_{x \in \wspace} g(x)$,
    \begin{align}
        0 < g_{\min} \leq g_{\max} < \infty \,.
    \end{align}
\end{assumption}

\subsection{The coverage assumptions}

The conditions we need for our technique concern the ability to cover any $2\varepsilon$-radius reachable set with a constant number of $\approx \varepsilon$-radius reachable sets, and to cover the starting set $\cspace_0$ with a polynomially or even exponentially-increasing number of $\approx \varepsilon$-radius reachable sets.

We begin with the \emph{$b$-coverability} condition. The \emph{branching factor} $b > 1$ is an integer roughly corresponding to the number of $\varepsilon$-reachable sets needed to cover a $2\varepsilon$-reachable set in $\cspace$:
\begin{assumption} \label{assm:b-coverability}
There is some integer $b > 1$, called the \emph{branching factor}, such that for any $\rho > 0$, there is some $\varepsilon^*_\rho > 0$ such that for any $\bq \in \cspace$ and $0 < \varepsilon \leq \varepsilon^*_\rho$, there is a set $Q_{\rho,\varepsilon}(\bq)$ of configurations such that
\begin{align} \label{eq:branching-factor-appx}
    R_{2\varepsilon}(\bq) \subseteq \bigcup_{\bq' \in Q_{\rho,\varepsilon}(\bq)} R_{(1+\rho)\varepsilon}(\bq') \text{ and } |Q_{\rho,\varepsilon}(\bq)| \leq b
\end{align}
\end{assumption}

\begin{lemma} \label{lem:b-coverability}
    If $\Pi$ is symmetric and satisfies \Cref{meta:regularity}, \Cref{assm:b-coverability} holds.
\end{lemma}



We also have the \emph{starting coverability} condition, which says that the set $\projinv{\wspacef}$ (configurations on the support of $f$) can be covered with a polynomial number of $\varepsilon$-reachable sets:
\begin{assumption} \label{assm:starting-coverability}
There exists a polynomial $P$ such that there is some $\varepsilon^* > 0$ such that for any $\varepsilon < \varepsilon^*$ there is a set $Q^0_\varepsilon$ of configurations such that
\begin{align}
    \projinv{\wspacef} \subseteq \bigcup_{\bq \in Q^0_\varepsilon} R_{\varepsilon}(\bq) \text{ and } |Q^0_\varepsilon| \leq P(1/\varepsilon)
\end{align}
\end{assumption}
Note that this uses configuration reachable sets rather than workspace reachable sets. 

The real possibility that this guards against (given that $\wspacef$ is bounded by \Cref{assm:density-fn}) is that the set of configurations one can be in at any $x \in \wspacef$ is unbounded. For symmetric control-affine systems, reachable sets contain rectilinear sets with volumes polynomial in $\varepsilon$ as $\varepsilon \to 0$ (though the exponent is not $\gamma$ as we are discussing configuration reachable sets), so if $\projinv{\wspacef}$ is bounded, the above holds.

One quick way to check \Cref{assm:starting-coverability} holds is that if $\Pi$ is symmetric, control-affine, and equiregular (satisfying \Cref{meta:regularity}), then \Cref{assm:starting-coverability} is equivalent to the closure of $\projinv{\wspacef}$ being compact (which typically just means bounded since it's already closed): 

\begin{lemma} \label{lem:starting-coverability}
    If $\Pi$ is symmetric and satisfies \Cref{meta:regularity}, \Cref{assm:starting-coverability} holds.
\end{lemma}

This holds because, by the Uniform Ball-Box Theorem, for sufficiently small $\varepsilon$, each configuration $\varepsilon$-reachable set anchored at $\bq$ contains a box centered at $\bq$ whose with side lengths polynomial in $\varepsilon$, and therefore with volume polynomial in $\varepsilon$ (with the same polynomial for each $\bq$). Then, the compact $\projinv{\wspacef}$ can be tiled by polynomially many of these boxes (allowing for some overlap) since $\cspace$ is a manifold and is therefore approximately Euclidean on small scales. See Appendix~\ref{appx:meta-assumption} for the complete proof.

\subsection{Hierarchical Cell Structures}

There are some additional conditions which are needed to make our algorithm work (through which we prove our upper bounds on the TSP length), as we need to know that we can range over the whole space in an efficient manner. In particular, we assume the existence of (and ability to cover the whole space with) \emph{hierarchical cell structures}. The hierarchical cell structure generally reflects the intuition that a $\varepsilon$-workspace reachable set should be divisible into $2^\gamma$ $(\varepsilon/2)$-workspace reachable sets, which should each be divisible into $2^\gamma$ $(\varepsilon/4)$-workspace reachable sets, and so on, forming a sort of hierarchy of reachable sets of exponentially decreasing radius. We define:

\begin{definition} \label{def:hcs}
    A $\zeta$-regular \emph{hierarchical cell structure} $\hcs(\bq_0,\varepsilon_0,\alpha,s,\zeta)$ rooted at $\bq_0 \in \cspace$ with radius $\varepsilon_0 > 0$, efficiency parameter $\alpha \leq 1$, and scaling parameter $s$ is a structure consisting of the following elements:
    \begin{itemize}
        \item A Jordan-measurable \emph{cell} $S(\bq_0,\varepsilon_0) \subseteq \wspace$ such that
        \begin{align}
            S(\bq_0,\varepsilon_0) \subseteq \bar{R}_{ \varepsilon_0}(\bq_0) \text{ and } \vol_{\wspace} (S(\bq_0,\varepsilon_0)) \geq \alpha \check{g}^{(\zeta)}(\proj{\bq_0}) \varepsilon_0^\gamma
        \end{align}
        where $\check{g}^{(\zeta)}$ is the lower $\zeta$-regularized approximation of $g$. 
        \item A set of $s^\gamma$ disjoint $\zeta$-regular hierarchical cell structures with radius $\varepsilon_0/s$, efficiency parameter $\alpha$, scaling parameter $s$, and rooted at $\bq_1, \dots, \bq_{s^\gamma}$ such that
        \begin{align}
            S(\bq_0,\varepsilon_0) \subseteq \bigcup_{j=1}^{s^\gamma} S(\bq_j,\varepsilon_0/s) ~~~\text{ and }~~~ \bq_j \in \bar{R}_{ \varepsilon_0}(\bq_0) \text{ for all } j
        \end{align}
        We call these the \emph{sub-HCS}'s of the original hierarchical cell structure.
    \end{itemize}
\end{definition}
We will sometimes refer to a set of HCS's as $S_1, \dots, S_m$; in this case, we say $x \in S_j$ if $x$ is contained in the top-level cell. Note that this is a recursive definition, i.e. that to give $\hcs(\bq_0,\varepsilon_0,\alpha,s,\zeta)$ we also need to give its components $\hcs(\bq_j,\varepsilon_0/s,\alpha,s,\zeta)$ for $j$ from $1$ to $s^\gamma$, and then in turn \emph{their} components, and so forth. Thus, to know the hierarchical cell structure at $\bq_0, \varepsilon_0$ requires knowing a full hierarchy of cells which can be represented as an infinite tree with $s^\gamma$ branches at each node (as we will do when we define the \emph{hierarchical collection problem}). This is generally possible if we have some kind of regular structure to exploit, for instance the rectilinear sets inscribed within reachable sets implied by the Ball-Box Theorem (see \Cref{lem:existence-of-hcs}).

The assumption of Jordan-measurability of the cells is meant to exclude contrived cell structures. One important property of Jordan-measurability is that the volume of a Jordan-measurable set is the same as the volume of its interior, a fact which is used in the proof of \Cref{lem:hcs-cover-02}.

\begin{assumption} \label{assm:hcs}
    There is some $\alpha$ and $s$ such that for any $\zeta > 0$, there exists some $\varepsilon_0$ such that at any $x \in \wspacef$, there is a $\zeta$-regular HCS with radius $\varepsilon_0$, efficiency $\alpha$, and scaling parameter $s$ containing $x$ in its interior.
\end{assumption}

We can show this holds for our (well-behaved) control-affine dynamics:
\begin{lemma} \label{lem:existence-of-hcs}
    If $\Pi$ is symmetric and satisfies \Cref{meta:regularity} then \Cref{assm:hcs} holds with $s = 2$.
\end{lemma}
See Appendix~\ref{appx:meta-assumption} for proof. Note that this lemma doesn't show anything about how large $\alpha$ is. However, the existence of an $\alpha$ tells us a lot since $\alpha$ does not depend on $f$ (since the definition of the HCS makes no reference to the density function).

We now discuss covering $\wspacef$ with Hierarchical Cell Structures, which is the basis of our DTSP algorithm for symmetric dynamics:
\begin{definition}
    For $\rho \geq 0$, a \emph{$\rho$-accurate Hierarchical Cell Structure cover} (HCS cover) of $\wspacef$ is a set $S_1, \dots, S_m$ of HCS's such that:
    \begin{itemize}
        \item All HCS's have the same parameters ($\varepsilon_0$, $\alpha$, $\zeta$, $s$).
        \item $\wspacef \subseteq \bigcup_{j=1}^m S_j$.
        \item $\bbP_{X \sim f}[X \text{ is in more than one } S_j] \leq \rho$.
    \end{itemize}
\end{definition}
Any HCS with overlap parameter $\alpha$ necessarily satisfies the condition for any overlap parameter $\alpha' \leq \alpha$; thus, if the HCS's have different values of $\alpha$, we can use the minimum.

\Cref{assm:hcs} implies the existence of HCS covers of arbitrarily good accuracy; for proofs of these results, see Appendix~\ref{appx:hcs-subdivision}. First, we have the existence of a fixed-scale finite HCS cover (possibly with bad accuracy $\rho$):
\begin{lemma} \label{lem:hcs-cover-01}
    There exist $\alpha, s$ such that for all $\zeta > 0$, there is some $\varepsilon_0$ such that there exists a finite HCS cover (of some accuracy parameter $\rho$) of $\wspacef$.
\end{lemma}

Then, given a finite HCS covering, for any $\rho > 0$ we can produce a $\rho$-accurate covering at arbitrarily small scales by taking the sub-HCS's recursively until we reach the desired scale, and deleting any redundant ones. 
\begin{lemma} \label{lem:hcs-cover-02}
    If \Cref{assm:hcs} is satisfied with parameters $\alpha,s$, then for any $\rho > 0$ and $\zeta > 0$, there is some $\varepsilon_0$ such that a $\rho$-accurate $\zeta$-regularized HCS cover $S_1, \dots, S_m$ can be constructed. Furthermore, for any $\varepsilon^*_0 > 0$, there is some $\varepsilon_0 \leq \varepsilon^*_0$ for which it exists.
\end{lemma}

In our upper bound proofs in \Cref{sec:dstsp-ub} and \Cref{sec:worst-case}, we will assume that the HCS cover we base our algorithm on is $0$-accurate, i.e. there is no overlap at all between different HCS's. See Appendix~\ref{appx:good-target-bad-target} for a proof that this does not affect the main results (even without it the results hold) as long as \Cref{lem:hcs-cover-02} holds.

\begin{remark}
    For regular symmetric control-affine dynamics (on which we can invoke the Ball-Box Theorem) and continuous density functions $f$ we can find HCS's with $s = 2$. However, we make a broader definition allowing $s \neq 2$ so that targets distributed within sets of fractal dimension, which may scale more naturally with some other $s$ (e.g. the Menger Sponge naturally scales with $s = 3$) may be analyzed in future work.
\end{remark}

\subsection{Approximately everywhere} \label{sec:appx-everywhere}

There are certain important cases where the assumptions don't strictly hold over the entire space -- for instance, if the parameters of the dynamics have a discontinuity (say, a Dubins car whose turning radius sharply changes when it enters a given region), \Cref{assm:reachability-volume} may not hold because $\bq$ near the boundary may require $\varepsilon$ to be arbitrarily small, so no fixed $\varepsilon^*_\zeta$ will work. In order to deal with this instance, we define the notion of an assumption holding \emph{approximately everywhere} on some $\wspace^* \subseteq \wspace$ or $\cspace^* \subseteq \cspace$.

\begin{definition} \label{def:appx-everywhere}
    An assumption holds \emph{approximately everywhere} on a set $A$ in a space $\cA$ with metric $d_\cA$ and volume $\vol_\cA$ if, for every $\eta > 0$, there is some $A_\eta$ such that:
    \begin{itemize}
        \item The assumption holds on $A_\eta$.
        \item $\vol_\cA(A \backslash A'_\eta) \leq \eta$ where $A'_\eta \defeq \{a \in \cA : \inf_{a' \not \in A_\eta} d_\cA(a,a') \geq \eta \}$ (the $\eta$-interior of $A_\eta$).
    \end{itemize}
    We call $A_\eta$ the \emph{$\eta$-approximator} of $A$.
\end{definition}
In short, the assumption should hold on a set $A_\eta$ whose $\eta$-interior $A'_\eta$ (the set of points at least $\eta$ away from points outside the set) takes up most of $A$, i.e. $A \backslash A'_\eta$ has volume at most $\eta$. While we use the same $\eta > 0$ as the depth of the interior and the cap on the volume, they could be separated into $\eta_1, \eta_2 > 0$ and the definition would be equivalent (by using $\eta = \min(\eta_1, \eta_2)$).

\begin{remark} \label{rem:appx-everywhere}
    One thing to note is that as $\eta \to 0$, $A_\eta$ and $A'_\eta$ will generally expand to more completely fill $A$, and the condition $A_{\eta_1} \subseteq A_{\eta_2}$ if $\eta_1 \geq \eta_2$ can be enforced without affecting the definition. Let's define $A^* = \bigcup_{\eta \to 0} A_\eta$: noting that our assumption in question holds on all $A_\eta$ for $\eta > 0$, can we say that it must hold on $A^*$? No -- many of our assumptions take the form `for all $\zeta > 0$, there exists some $\varepsilon^*_\zeta > 0$ such that for all $0 < \varepsilon \leq \varepsilon^*_\zeta$ a certain condition holds for all $a$'; however, a different $\varepsilon^*_\zeta$ may be needed for each given $A_\eta$ (so really it should be $\varepsilon^*_{\zeta,\eta}$), and no $\varepsilon^*_\zeta > 0$ may work for all $\eta > 0$.
\end{remark}

\Cref{rem:appx-everywhere} shows why we need \Cref{def:appx-everywhere} rather than the more common `almost everywhere'. We then note that we only need our assumptions to hold approximately everywhere for our main results to hold:
\begin{proposition} \label{prop:appx-everywhere}
    If \Cref{prop:dstsp-lower-bound} and \Cref{prop:dstsp-upper-bound} are true when all the assumptions hold everywhere, they are also true when all the assumptions hold approximately everywhere on the support $\wspacef$. 
\end{proposition}
For the proof, see Appendix~\ref{appx:appx-everywhere}.



\section{The DSTSP Lower Bound}\label{sec:dstsp-lb}

We now show a \wvhp~lower bound for the DSTSP which not only includes the order of growth $\Theta(n^{1-\frac{1}{\gamma}})$ with regard to the number $n$ of target points, but which describes the relationship between the density $f$ of target points, the agility function $g$ of the system, and the small-time constraint factor $\gamma$. Following the techniques of \cite{adler-icra-16} and \cite{arias-ctd-05}, we wish to do this using the connection between Orienteering and the TSP; however, the need to be more precise regarding $f$ and $g$ causes a difficulty. An important difference between the Dynamic Stochastic TSP and Orienteering (on random targets) is that a TSP solution must visit all the targets, which may be spread throughout $\wspacef$, while an Orienteering path can choose to restrict itself to only a very small region within $\wspacef$ -- presumably one with a high density of target points and/or where the dynamic system has a larger range of motion. Thus, the TSP depends on $f$ and $g$ over all of $\wspacef$ while Orienteering essentially depends only on most advantageous or \emph{lucrative} area, in which target points can be visited most rapidly, while $f$ and $g$ elsewhere (with high probability) do not affect the solution at all.

\subsection{Lucrativity and the Adjusted Cost-Balancing Function} \label{sec:cbo_intuition}

In order to measure exactly how $f,g$ contribute to the `lucrativity' of a region, we consider the following: for small $\varepsilon > 0$, we have $\vol_{\wspace}(\bar{R}_\varepsilon(\bq)) \approx g(\bq) \varepsilon^\gamma$. Thus, we can estimate the probability that any given target falls into $\bar{R}_\varepsilon(\bq)$:
\begin{align}
    \bbP_{X \sim f}[X \in \bar{R}_\varepsilon(\bq)] &\approx f([\bq]_\wspace) g(\bq) \varepsilon^\gamma \\ &\leq f([\bq]_\wspace) g([\bq]_\wspace) \varepsilon^\gamma \label{eq:prob-mass-moral}
\end{align}
Thus, the expected number of targets that fall within a given radius-$\varepsilon$ reachable set anchored at $\bq \in [x]^\cspace$ can be bounded by
\begin{align}
    \bbE_{X_i \simt{iid} f}[|\{X_1, \dots, X_n\} \cap \bar{R}_\varepsilon(\bq) | ] \leq n f(x) g(x) \varepsilon^\gamma \,.
\end{align}
This means that if $\varepsilon = n^{-\frac{1}{\gamma}} (f(x) g(x))^{-\frac{1}{\gamma}}$, the expected number of targets in a $\varepsilon$-radius reachable set is at most $1$; we can then take $\varepsilon$ to be a rough measure of the average time to reach the nearest target from $x$, and hence $1/\varepsilon = n^{\frac{1}{\gamma}} (f(x) g(x))^{\frac{1}{\gamma}}$ is roughly the \emph{rate} at which we visit target points in the vicinity of $x$. Thus, we may define the \emph{lucrativity function} over $\wspace$ as
\begin{align} \label{eq::lucrativity}
    \cost^*(x) \defeq \cost^*_{f,g}(x) \defeq \big(f(x) g(x)\big)^{\frac{1}{\gamma}}
\end{align}
which is proportional to the rate at which the system can expect to encounter target points near $x$ (when $n$ is large).

Thus, we want to modify the Orienteering problem to balance out the lucrativity over the whole space and not have it favor any region over any other. Intuitively, this can be done by using lucrativity function as a cost function: the system is `charged' $\cost^*(x)$ cost per unit length for movement at $x$. This then means that the rate it can expect to encounter target points is roughly $1$ per unit cost, everywhere; we refer to this as the problem being `balanced'.

\begin{remark} \label{rem:cbo-failure}
    Unfortunately, while this intuition generally holds across the space under a very broad set of conditions, there are many cases of interest in which this will not hold everywhere. In particular, \eqref{eq:prob-mass-moral} may fail wherever $f([\bq]_\wspace) = 0$ or $g([\bq]_\wspace) = 0$, or wherever $f$ or $g$ has a discontinuity. This leads in particular to a problem where movement outside of $\wspacef$ (i.e. where $f(x) = 0$) is `free' (no cost). This makes the Orienteering problem with cost function $\cost^*$ too powerful to provide a useful lower bound to the TSP.
\end{remark}

To deal with the issue raised in \Cref{rem:cbo-failure}, we modify the lucrativity function to obtain a cost function for our dynamics:
\begin{definition} \label{def:cbo-cost}
The \emph{adjusted cost-balancing function} (which we will generally refer to as the \emph{cost function}) of density $f$ and agility $g$ with regularization factor $\zeta > 0$ is $\cost: \wspace \to \bbR$ is
\begin{align}
    \cost_\zeta(x) \defeq (\hat{f}^{(\zeta)}(x) \hat{g}^{(\zeta)}(x))^{\frac{1}{\gamma}}
\end{align}
i.e. $\cost_\zeta(x)$ is the product of the upper $\zeta$-regularizations of $f$ and $g$ (the minimal upper bounds of $f,g$ which are $\geq \zeta$ and $(1/\zeta)$-Lipschitz continuous, see \Cref{def:alpha-reg-appx}).

Then, for a trajectory $\pi \in \Pi$, its cost is
\begin{align}
    \ell^{\cost}(\pi) \defeq \ell_\zeta^{\cost}(\pi) \defeq \int_{0}^{\ell(\pi)} \cost\big(\pi(t)\big) dt \mathrm.
\end{align}
For convenience, we want to be able to reparameterize $\pi$ by cost. For any $t \in [0,\ell(\pi)]$, we define
\begin{align}
    t^{\cost} \defeq t^{\cost}_\pi \defeq \int_0^t \cost_\zeta(\pi(t')) \, dt'
\end{align}
and define $\pi^{\cost} : [0,\ell^{\cost}(\pi)] \to \cspace$ as the trajectory satisfying
\begin{align}
    \pi^{\cost}(t^{\cost}) = \pi(t)\,.
\end{align}
\end{definition}
We will use the `$\cost$' symbol in general to denote cost-denominated versions of definitions from the previous section, e.g. $\bar{R}^{\cost}_\varepsilon(\bq)$ for the region reachable in $\leq \varepsilon$ cost from $\bq$. In general, the value of $\zeta$ will be fixed and $\zeta$ will be left out of the notation.

We need to show that the cost function satisfies certain important properties, specifically being bounded above and below (away from $0$) and being Lipschitz continuous:
\begin{lemma}
    The cost function $\cost_\zeta(x) = (\hat{f}^{(\zeta)}(x)\hat{g}^{(\zeta)}(x))^{\frac{1}{\gamma}}$ satisfies the following:
    \begin{enumerate} [i.]
        \item $\cost_\zeta(x) \geq \cost^*(x)$ (where $\cost^*(x)$ is the lucrativity function $(f(x)g(x))^{\frac{1}{\gamma}}$) for all $x$.
        \item $\cost_\zeta(x)$ is uniformly bounded away from $0$ and is Lipschitz continuous. In particular, if $\gamma \geq 1$, then it is $\alpha$-Lipschitz continuous where
        \begin{align}
            \alpha = \frac{1}{\gamma} \zeta^{\frac{2}{\gamma}-3} (f_{\max}+g_{\max})
        \end{align}
        and $\cost_\zeta(x) \geq \zeta^{\frac{2}{\gamma}}$ everywhere.
        \item $\lim_{\zeta \to 0} \cost_\zeta(x) = \cost^*(x)$ almost everywhere.
    \end{enumerate}
\end{lemma}

\begin{proof}
    We prove these in order.

    \emph{i.} This follows from the fact that $\hat{f}^{(\zeta)}$ and $\hat{g}^{(\zeta)}$ are upper bounds for $f$ and $g$ (which are nonnegative) by construction, hence 
    \begin{align}
        \cost_\zeta(x) = (\hat{f}^{(\zeta)}(x)\hat{g}^{(\zeta)}(x))^{\frac{1}{\gamma}} \geq (f(x)g(x))^{\frac{1}{\gamma}} = \cost^*(x) \,.
    \end{align}

    \emph{ii.} This follows because by construction $\hat{f}^{(\zeta)}$ and $\hat{g}^{(\zeta)}$ are both $\geq \zeta$ everywhere, so 
    \begin{align}
        \cost_\zeta(x) = (\hat{f}^{(\zeta)}(x)\hat{g}^{(\zeta)}(x))^{\frac{1}{\gamma}} \geq \zeta^{\frac{2}{\gamma}} \,.
    \end{align}
    To show that they are Lipschitz continuous, we note that both $f$ and $g$ are bounded above (by $f_{\max}$ and $g_{\max}$ respectively) and that $\sup_x \hat{f}^{(\zeta)}(x) = f_{\max}$ and $\sup_x \hat{g}^{(\zeta)}(x) = g_{\max}$, and by construction $\hat{f}^{(\zeta)}$ and $\hat{g}^{(\zeta)}$ are $(1/\zeta)$-Lipschitz continuous. 
    
    However, the composition of two Lipschitz-continuous function is also Lipschitz-continuous (and the Lipschitz continuity factor of the composition is the product of the factors of the original two functions). Furthermore $(\cdot)^{\frac{1}{\gamma}}$ is Lipschitz-continuous if the domain is bounded away from $0$ below and upper bounded away from $\infty$; and when the input is bounded below by $\zeta^2$ and above by $f_{\max} g_{\max}$ (as in this case), it is $\alpha_1$-Lipschitz continuous where
    \begin{align}
        \alpha_1 = \begin{cases} \frac{1}{\gamma} \zeta^{2(\frac{1}{\gamma}-1)} &\text{if } \gamma \geq 1 \\ \frac{1}{\gamma} (f_{\max} g_{\max})^{\frac{1}{\gamma}-1} &\text{if } \gamma < 1 \end{cases}
    \end{align}
    We now analyze the Lipschitz-continuity of $\hat{f}^{(\zeta)}(x) \hat{g}^{(\zeta)}(x)$; the product of two bounded Lipschitz continuous functions is also a bounded Lipschitz continuous function. Specifically, if $h_1(x), h_2(x)$ are respectively $\zeta_1$- and $\zeta_2$-Lipschitz continuous nonnegative functions with respective fixed upper bounds $h_1^{\max}, h_2^{\max}$, then $h(x) = h_1(x) h_2(x)$ is also nonnegative, is bounded above by $h_1^{\max} h_2^{\max}$ and is $(\zeta_1 h_2^{\max} + \zeta_2 h_1^{\max})$-Lipschitz continuous. Thus, $\hat{f}^{(\zeta)}(x) \hat{g}^{(\zeta)}(x)$ is nonnegative, bounded above by $f_{\max} g_{\max}$ (and below by $\zeta^2$), and is $\alpha_2$-Lipschitz continuous where
    \begin{align}
        \alpha_2 = (1/\zeta)(f_{\max}+g_{\max}) \,.
    \end{align}
    Hence $\cost_\zeta(x)$ is $\alpha$-Lipschitz continuous where $\alpha = \alpha_1 \alpha_2$. In particular, when $\gamma \geq 2$ (the case we are most interested in) we have
    \begin{align}
        \alpha = \frac{1}{\gamma} \zeta^{\frac{2}{\gamma}-3} (f_{\max}+g_{\max}) \,.
    \end{align}

    \emph{iii.} This follows as $\hat{f}^{(\zeta)} \to f$ and $\hat{g}^{(\zeta)} \to g$ almost everywhere as $\zeta \to 0$; hence \emph{both} occur almost everywhere (the union of where they don't converge is measure $0$).
\end{proof}

We also define cost-denominated versions of distance and reachable sets:
\begin{definition} \label{def:cbo-reachable-set}
    Given $\bq, \bq' \in \cspace$, the cost-distance between them is
    \begin{align}
        d^{\cost}_{\Pi}(\bq, \bq') = \inf (\ell^{\cost}(\pi): \bq \tot{\pi} \bq', \pi \in \Pi)
    \end{align}
    Given $\varepsilon > 0$ and $\bq \in \cspace$, the \emph{$\varepsilon$-cost reachable set} (in both $\cspace$ and $\wspace$) are
    \begin{align}
        R^{\cost}_\varepsilon(\bq) \defeq \{\bq' \in \cspace : d^{\cost}_{\Pi}(\bq, \bq') \leq \varepsilon\} ~~~\text{and}~~~ \bar{R}^{\cost}_\varepsilon(\bq) = [R^{\cost}_\varepsilon(\bq)]_{\wspace}
    \end{align}
\end{definition}

Although intuitively on small scales the cost-reachable sets will resemble our original length-reachable sets (except scaled by the cost function $\cost$), they are not exactly the same shape. Thus, we need to show that this change does not alter our main assumptions, namely \Cref{assm:b-coverability} and \Cref{assm:starting-coverability}; we also need to show that (with sufficiently small $\zeta > 0$ regularization) it achieves the balancing effect we wanted.
\begin{lemma} \label{lem:cost-adjusting}
Let $b$ be the branching factor from \Cref{assm:b-coverability}. Then there are functions $\rho_i : \bbR_{> 0} \to \bbR_{\geq 0}$ for $i = 1,2,3,4$ satisfying $\lim_{\zeta \to 0} \rho_i(\zeta) = 0$ and a polynomial $P^{\cost}$ (which depends through the cost function $\cost \defeq \cost_\zeta$ on $\zeta$) such that for any $\zeta > 0, \bq \in \cspace$ and $0 < \varepsilon \leq \rho_1(\zeta)$ the following hold:
\begin{enumerate}[i.]
    \item There is a set $Q^{\cost}_\varepsilon(\bq)$ such that $|Q^{\cost}_\varepsilon(\bq)| \leq b$ and
    \begin{align}
        R^{\cost}_{2\varepsilon}(\bq) \subseteq \bigcup_{\bq' \in Q^{\cost}_\varepsilon(\bq)} R^{\cost}_{(1+\rho_2(\zeta))\varepsilon}(\bq') \,.
    \end{align}
    \item There is a set $Q^{\cost,0}_\varepsilon$ such that $|Q^{\cost,0}_\varepsilon| \leq P^{\cost}(1/\varepsilon)$ and
    \begin{align}
        \cspace_f \subseteq \bigcup_{\bq \in Q^{\cost,0}_\varepsilon} R^{\cost}_\varepsilon(\bq)
    \end{align}
    \item $\bbP_{X \sim f}[X \in \bar{R}^{\cost}_\varepsilon(\bq)] \leq (1+\rho_3(\zeta))\varepsilon^\gamma$.
    \item $\bbE_{X \sim f}[\cost_\zeta(X)^{-1}] \geq (1-\rho_4(\zeta)) \bbE_{X \sim f}[(f(X)g(X))^{-\frac{1}{\gamma}}]$.
\end{enumerate}
\end{lemma}

\begin{proof}
    We assume without loss of generality that $\zeta \leq 1$ (so $\alpha_1 = \zeta^{\frac{2}{\gamma}} \leq 1$) and $\varepsilon$ is sufficiently small so that \Cref{assm:b-coverability} and \Cref{assm:starting-coverability} hold.

    The lemma then holds due to the properties of $\cost_\zeta(x)$, namely that it is bounded below by $\alpha_1$ and $\alpha_2$-Lipschitz continuous; these two conditions show that on small scales, it cannot change too much in a multiplicative sense. First, we define
    \begin{align}
        \olra{R}_\varepsilon(\bq) \text{ and } \olra{R}^{\cost}_\varepsilon(\bq)
    \end{align}
    to be the sets reachable from $\bq$ using (possibly alternating) backwards and forwards trajectories in $\Pi$ of, respectively total length $\leq \varepsilon$ or total cost $\leq \varepsilon$ (note that if $\Pi$ is symmetric then $R_\varepsilon(\bq) = \olra{R}_\varepsilon(\bq)$ and $R^{\cost}_\varepsilon(\bq) = \olra{R}^{\cost}_\varepsilon(\bq)$).
    
    Then, letting $c_\Pi < \infty$ be the speed limit of $\Pi$ in $\wspace$ (i.e. $\norm{[\dot{\pi}(t)]_\wspace} \leq c_{\Pi}$ for any $\pi \in \Pi$), which of course also applies for backwards trajectories, we get
    \begin{align}
        \cost_\zeta([\bq]_\wspace) - (5c_\Pi (\alpha_2/\alpha_1))\varepsilon &\leq \cost_\zeta([\bq']_\wspace) \leq \cost_\zeta([\bq]_\wspace) + (5c_\Pi (\alpha_2/\alpha_1))\varepsilon 
        \\ &\text{ for any } \bq \in \cspace \text{ and } \bq' \in \olra{R}_{(5/\alpha_1)\varepsilon}(\bq)
    \end{align}
    since $\cost$ is $\alpha_2$-Lipschitz continuous and $\bq'$ is by definition reachable from $\bq$ using $\Pi$ (and backwards $\Pi$) in at most $(5/\alpha_1)\varepsilon$ time, which translates to at most $(5c_\Pi/\alpha_1)\varepsilon$ distance in the metric on $\wspace$, which finally translates to a change of at most $(5c_\Pi (\alpha_2/\alpha_1))\varepsilon$ in the value of $\cost$ by Lipschitz continuity. Then since $\cost_\zeta([\bq]_\wspace) \geq \alpha_1$, we can turn this into multiplicative bounds:
    \begin{align} \label{eq::cost-multiplicative-bound}
        (1 - (5c_\Pi(\alpha_2/\alpha_1^2))\varepsilon) \cost_\zeta([\bq]_\wspace) &\leq \cost_\zeta([\bq']_\wspace) \leq (1 + (5c_\Pi(\alpha_2/\alpha_1^2))\varepsilon) \cost_\zeta([\bq]_\wspace) 
        \\ &\text{ for any } \bq \in \cspace \text{ and } \bq' \in \olra{R}_{(5/\alpha_1)\varepsilon}(\bq)
    \end{align}
    We then let $\rho_\zeta^*(\varepsilon) \defeq (5c_\Pi(\alpha_2/\alpha_1^2))\varepsilon$ (noting that $\alpha_1, \alpha_2$ depend on $\zeta$, as well as $\gamma$ and $f$ and $g$ through $f_{\max},g_{\max}$); for any fixed $\zeta$, we have $\lim_{\varepsilon \to 0} \rho_\zeta^*(\varepsilon) = 0$.

    Furthermore, since $\cost_\zeta(x) \geq \alpha_1$ for all $x$, we know that
    \begin{align}
        \olra{R}^{\cost}_{5\varepsilon}(\bq) \subseteq  \olra{R}_{(5/\alpha_1)\varepsilon}(\bq)  
    \end{align}
    since any trajectory (allowing backwards movement) of cost $\leq 5\varepsilon$ must have length $\leq (5/\alpha_1)\varepsilon$. Thus, our bounds \eqref{eq::cost-multiplicative-bound} hold for all $\bq' \in \olra{R}^{\cost}_{5\varepsilon}(\bq)$ as well.

    We now fix $\bq \in \cspace$ and consider $R^{\cost}_{\varepsilon'}(\bq')$; if $R^{\cost}_{\varepsilon'}(\bq') \subseteq \olra{R}^{\cost}_{5\varepsilon}(\bq)$, then we can conclude
    \begin{align} \label{eq::cost-length-containment}
        R_{\varepsilon'/((1+\rho^*_\zeta(\varepsilon))\cost_\zeta([\bq]_\wspace))}(\bq') \subseteq R^{\cost}_{\varepsilon'}(\bq') \subseteq R_{\varepsilon'/((1-\rho^*_\zeta(\varepsilon))\cost_\zeta([\bq]_\wspace))}(\bq')
    \end{align}
    since at any $\bq'' \in R^{\cost}_{\varepsilon'}(\bq') \subseteq \olra{R}^{\cost}_{5\varepsilon}(\bq)$, we know that the cost is between $(1 - \rho^*_\zeta(\varepsilon)) \cost_\zeta([\bq]_\wspace)$ and $(1 + \rho^*_\zeta(\varepsilon)) \cost_\zeta([\bq]_\wspace)$. Similarly, if $\olra{R}^{\cost}_{\varepsilon'}(\bq') \subseteq \olra{R}^{\cost}_{5\varepsilon}(\bq)$, then
    \begin{align} \label{eq::cost-length-containment-symmetric}
        \olra{R}_{\varepsilon'/((1+\rho^*_\zeta(\varepsilon))\cost_\zeta([\bq]_\wspace))}(\bq') \subseteq \olra{R}^{\cost}_{\varepsilon'}(\bq') \subseteq \olra{R}_{\varepsilon'/((1-\rho^*_\zeta(\varepsilon))\cost_\zeta([\bq]_\wspace))}(\bq')
    \end{align}
    (Note: the distinction between $\varepsilon'$ and $\varepsilon$ is very important in the above).

    We now prove part (i). Equation \eqref{eq::cost-length-containment} gives us the following:
    \begin{align} 
        R^{\cost}_{2\varepsilon}(\bq) \subseteq R_{2\varepsilon/((1-\rho^*_\zeta(\varepsilon))\cost_\zeta([\bq]_\wspace))}(\bq)
    \end{align}
    
    We can then apply \Cref{assm:b-coverability} (since the right-hand side is the normal reachable set) to get a cardinality-$b$ set $Q_{\varepsilon/((1-\rho^*_\zeta(\varepsilon))\cost_\zeta([\bq]_\wspace))}(\bq)$ such that
    \begin{align}
        R_{2\varepsilon/((1-\rho^*_\zeta(\varepsilon))\cost_\zeta([\bq]_\wspace))}(\bq) \subseteq \bigcup_{\bq' \in Q_{\varepsilon/((1-\rho^*_\zeta(\varepsilon))\cost_\zeta([\bq]_\wspace))}(\bq)} R_{\varepsilon/((1-\rho^*_\zeta(\varepsilon))\cost_\zeta([\bq]_\wspace))}(\bq')
    \end{align}
    Note that without loss of generality, for any $\varepsilon'$ and any $\bq'$ (note: this $\bq'$ is not related to $Q_{\varepsilon/((1-\rho^*_\zeta(\varepsilon))\cost_\zeta([\bq]_\wspace))}(\bq)$, it is any $\bq' \in \cspace$), we can assume 
    \begin{align}
        Q_{\varepsilon'}(\bq') \subset \olra{R}_{3\varepsilon'}(\bq')
    \end{align}
    since WLOG we can assume that any $\bq'' \in Q_{\varepsilon'}(\bq')$ has the property
    \begin{align}
        R_{\varepsilon'}(\bq'') \cap R_{2\varepsilon'}(\bq') \neq \emptyset
    \end{align}
    as the points in $Q_{\varepsilon'}(\bq')$ are being used to cover $R_{2\varepsilon'}(\bq')$. This in turn yields
    \begin{align}
        R_{\varepsilon'}(\bq'') \subseteq \olra{R}_{4\varepsilon'}(\bq')
    \end{align}
    for any $\bq'' \in Q_{\varepsilon'}(\bq')$. Thus we can apply this to
    \begin{align}
        \varepsilon' = \varepsilon/((1-\rho^*_\zeta(\varepsilon))\cost_\zeta([\bq]_\wspace)) \text{ and } \bq' = \bq
    \end{align}
    which then yields for any $\bq' \in Q_{\varepsilon/((1-\rho^*_\zeta(\varepsilon))\cost_\zeta([\bq]_\wspace))}$ that
    \begin{align}
        R_{\varepsilon/((1-\rho^*_\zeta(\varepsilon))\cost_\zeta([\bq]_\wspace))}(\bq') \subset \olra{R}_{4\varepsilon/((1-\rho^*_\zeta(\varepsilon))\cost_\zeta([\bq]_\wspace))}(\bq)
    \end{align}
    Then, for $\varepsilon \leq \alpha_1^2/(45 \alpha_2 c_{\Pi})$ we have
    \begin{align}
        \rho^*_\zeta(\varepsilon) \leq 1/9
    \end{align}
    which in turn means
    \begin{align}
        4 (1+\rho^*_\zeta(\varepsilon))/(1-\rho^*_\zeta(\varepsilon)) \leq 4(10/9)/(8/9) = 5
    \end{align}
    and hence by \eqref{eq::cost-length-containment-symmetric} we have
    \begin{align}
        \olra{R}_{4\varepsilon/((1-\rho^*_\zeta(\varepsilon))\cost_\zeta([\bq]_\wspace))}(\bq) \subseteq \olra{R}^{\cost}_{5\varepsilon}(\bq)
    \end{align}
    This then finally yields by \eqref{eq::cost-length-containment} that for all $\bq' \in Q_{\varepsilon/((1-\rho^*_\zeta(\varepsilon))\cost_\zeta([\bq]_\wspace))}(\bq)$,
    \begin{align}
        R_{\varepsilon/((1-\rho^*_\zeta(\varepsilon))\cost_\zeta([\bq]_\wspace))}(\bq') \subseteq R^{\cost}_{((1+\rho^*_\zeta(\varepsilon))/(1-\rho^*_\zeta(\varepsilon)))\varepsilon}(\bq')
    \end{align}
    when $\varepsilon \leq \alpha_1^2/(45 \alpha_2 c_{\Pi})$. We can then define
    \begin{align}
        \rho_1(\zeta) = \zeta \alpha_1^2/(45 \alpha_2 c_{\Pi}) \text{ and } \rho_2(\zeta) = \frac{1 + \zeta/9}{1 - \zeta/9} - 1
    \end{align}
    in which case when $\varepsilon \leq \rho_1(\zeta)$ we get
    \begin{align}
        \frac{1+\rho^*_\zeta(\varepsilon)}{1-\rho^*_\zeta(\varepsilon)} \leq \frac{1+\rho^*_\zeta(\rho_1(\zeta))}{1-\rho^*_\zeta(\rho_1(\zeta))} \leq \frac{1 + \zeta/9}{1 - \zeta/9} = 1 + \rho_2(\zeta)
    \end{align}
    where $\lim_{\zeta \to 0} \rho_2(\zeta) = 0$. Thus, we finally have
    \begin{align}
        R_{\varepsilon/((1-\rho^*_\zeta(\varepsilon))\cost_\zeta([\bq]_\wspace))}(\bq') \subseteq R^{\cost}_{(1 + \rho_2(\zeta))\varepsilon}(\bq')
    \end{align}
    Putting this together with the above, we get
    \begin{align}
        R^{\cost}_{2\varepsilon}(\bq) &\subseteq R_{2\varepsilon/((1-\rho^*_\zeta(\varepsilon))\cost_\zeta([\bq]_\wspace))}(\bq) 
        \\ &\subseteq \bigcup_{\bq' \in Q_{\varepsilon/((1-\rho^*_\zeta(\varepsilon))\cost_\zeta([\bq]_\wspace))}(\bq)} R_{\varepsilon/((1-\rho^*_\zeta(\varepsilon))\cost_\zeta([\bq]_\wspace))}(\bq')
        \\ &\subseteq \bigcup_{\bq' \in Q_{\varepsilon/((1-\rho^*_\zeta(\varepsilon))\cost_\zeta([\bq]_\wspace))}(\bq)} R^{\cost}_{(1 + \rho_2(\zeta))\varepsilon}(\bq')
    \end{align}
    and $|Q_{\varepsilon/((1-\rho^*_\zeta(\varepsilon))\cost_\zeta([\bq]_\wspace))}(\bq)| \leq b$, and hence we may finally conclude that part (i) is true, with
    \begin{align}
        Q^{\cost}_\varepsilon(\bq) = Q_{\varepsilon/((1-\rho^*_\zeta(\varepsilon))\cost_\zeta([\bq]_\wspace))}(\bq) \,.
    \end{align}
    
    Part (ii) follows by considering $\cost^*_{\max} \defeq \sup_x (f(x)g(x))^{\frac{1}{\gamma}}$, which is a constant with regard to $\zeta$ and $\varepsilon$. Then, noting that $R_{\varepsilon/\cost^*_{\max}}(\bq') \subseteq R^{\cost}_\varepsilon(\bq')$ since $\cost_\zeta(x) \leq \cost^*_{\max}$, we can simply use $P^{\cost}$ such that $P^{\cost}(1/\varepsilon) = P(\cost^*_{\max}/\varepsilon)$ and $Q^{\cost,0}_\varepsilon = Q_{\varepsilon/\cost^*_{\max}}$ and we are done.

    Part (iii) follows because when $\varepsilon \leq \rho_1(\zeta)$ and $\bq' \in R^{\cost}_{\varepsilon}(\bq)$,
    \begin{align}
        (1 - (c_\Pi(\alpha_2/\alpha_1^2))\varepsilon) \cost_\zeta([\bq]_\wspace) &\leq \cost_\zeta([\bq']_\wspace) \leq (1 + (c_\Pi(\alpha_2/\alpha_1^2))\varepsilon) \cost_\zeta([\bq]_\wspace) 
    \end{align}
    (under the same logic as before but with radius $\varepsilon$ rather than $5\varepsilon$). Thus we know that
    \begin{align}
        R^{\cost}_\varepsilon(\bq) \subseteq R_{\varepsilon/((1 - (c_\Pi(\alpha_2/\alpha_1^2))\varepsilon) \cost_\zeta([\bq]_\wspace))}(\bq)
    \end{align}
    Then we set $\rho_3(\zeta) = (1-\zeta/45)^{-\gamma} - 1$. Then, when $\varepsilon \leq \rho_1(\zeta) = \zeta \alpha_1^2/(45\alpha_2 c_\Pi)$, we have 
    \begin{align}
        1 - (c_\Pi(\alpha_2/\alpha_1^2))\varepsilon \geq 1 - \zeta/45 
    \end{align}
    which thus means (taking the above and projecting to $\wspace$)
    \begin{align}
        \bar{R}^{\cost}_\varepsilon(\bq) \subseteq \bar{R}_{\varepsilon(1 - \zeta/45)^{-1} / \cost_\zeta([\bq]_\wspace)}(\bq)
    \end{align}
    We then consider the volume of the set above. Since reachable set volumes approach $g(\bq) \varepsilon^\gamma$, for any $\rho_{3^*}(\zeta)$ there is some $\rho_{1^*}(\zeta)$ such that when $\varepsilon \leq \rho_{1^*}(\zeta)$,
    \begin{align}
        \vol(\bar{R}_{\varepsilon}(\bq)) \leq (1 + \rho_{3^*}(\zeta)) g(\bq) \varepsilon^\gamma \leq (1 + \rho_{3^*}(\zeta)) g([\bq]_\wspace) \varepsilon^\gamma
    \end{align}
    Using $\varepsilon(1 - \zeta/45)^{-1} / \cost_\zeta([\bq]_\wspace)$ instead of $\varepsilon$ gives (for $\varepsilon \leq (1-\zeta/45) \cost_\zeta([\bq]_\wspace) \rho_{1^*}(\zeta) \leq  (1-\zeta/45) \zeta^{\frac{2}{\gamma}} \rho_{1^*}(\zeta)$)
    \begin{align}
        \vol(\bar{R}_{\varepsilon(1 - \zeta/45)^{-1} / \cost_\zeta([\bq]_\wspace)}(\bq)) &\leq (1 + \rho_{3^*}(\zeta)) (1-\zeta/45)^{-\gamma} \cost_\zeta([\bq]_\wspace)^{-\gamma} g([\bq]_\wspace) \varepsilon^\gamma
        \\ &= (1 + \rho_{3^*}(\zeta)) (1-\zeta/45)^{-\gamma} \hat{f}^{(\zeta)}([\bq]_\wspace)^{-1} \varepsilon^\gamma
    \end{align}
    Finally we note that the entire reachable set is (by definition) within a distance of at most $\varepsilon(1 - \zeta/45)^{-1} / \cost_\zeta([\bq]_\wspace)$ and hence the Lipschitz continuity (and boundedness away from $0$) implies that for any $\rho_{3^{**}}(\zeta)$, there is some $\rho_{1^{**}}(\zeta)$ such that for all $\varepsilon < \rho_{1^{**}}(\zeta)$,
    \begin{align}
        \hat{f}^{(\zeta)}(\bq') \leq (1 + \rho_{3^{**}}(\zeta)) \hat{f}^{(\zeta)}(\bq) \text{ for all } \bq' \text{ within distance } \varepsilon \text{ of } \bq
    \end{align}
    Hence, setting $\varepsilon < \rho_{1^{**}}(\zeta) (1 - \zeta/45) \cost_\zeta([\bq]_\wspace) \leq \rho_{1^{**}}(\zeta) (1 - \zeta/45) \zeta^{\frac{2}{\gamma}}$ then yields
    \begin{align}
        \bbE_{X \sim f}[X \in \bar{R}^{\cost}_\varepsilon(\bq)] &\leq \bbE_{X \sim f}[X \in \bar{R}_{\varepsilon(1 - \zeta/45)^{-1} / \cost_\zeta([\bq]_\wspace)}(\bq)]
        \\ &= \int_{\bar{R}_{\varepsilon(1 - \zeta/45)^{-1} / \cost_\zeta([\bq]_\wspace)}(\bq)} f(x) \, dx
        \\ &\leq \int_{\bar{R}_{\varepsilon(1 - \zeta/45)^{-1} / \cost_\zeta([\bq]_\wspace)}(\bq)} \hat{f}^{(\zeta)}(x) \, dx
        \\ &\leq \int_{\bar{R}_{\varepsilon(1 - \zeta/45)^{-1} / \cost_\zeta([\bq]_\wspace)}(\bq)} (1 + \rho_{3^{**}}(\zeta)) \hat{f}^{(\zeta)}(\bq) \, dx
        \\ &= (1 + \rho_{3^{**}}(\zeta)) \hat{f}^{(\zeta)}(\bq) \vol(\bar{R}_{\varepsilon(1 - \zeta/45)^{-1} / \cost_\zeta([\bq]_\wspace)}(\bq))
        \\ &\leq  (1 + \rho_{3^{**}}(\zeta)) \hat{f}^{(\zeta)}(\bq)(1 + \rho_{3^*}(\zeta)) (1-\zeta/45)^{-\gamma} \hat{f}^{(\zeta)}([\bq]_\wspace)^{-1} \varepsilon^\gamma
        \\ &= (1 + \rho_{3^{**}}(\zeta)) (1 + \rho_{3^*}(\zeta)) (1-\zeta/45)^{-\gamma} \varepsilon^\gamma
    \end{align}
    Thus, noting that $\rho_{3^*}(\zeta)$ and $\rho_{3^{**}}(\zeta)$ can be made arbitrarily small as $\zeta \to 0$ (and $(1-\zeta/45)^{-\gamma} \to 1$ as $\zeta \to 0$), we can let
    \begin{align}
        \rho_3(\zeta) = (1 + \rho_{3^{**}}(\zeta)) (1 + \rho_{3^*}(\zeta)) (1-\zeta/45)^{-\gamma} - 1
    \end{align}
    and the above will hold for all $\varepsilon$ such that
    \begin{align}
        \varepsilon \leq \rho_{1^{*}}(\zeta) (1 - \zeta/45) \zeta^{\frac{2}{\gamma}} \text{ and } \rho_{1^{**}}(\zeta) (1 - \zeta/45) \zeta^{\frac{2}{\gamma}}
    \end{align}
    We then set $\rho_1$ to be the minimum of the above values (of which there are only a fixed, finite number, so it remains positive).
    
    Finally, part (iv) follows from the Monotone Convergence Theorem since
    \begin{align}
        \bbE_{X \sim f}[(f(X)g(X))^{-\frac{1}{\gamma}}] = \int_{\wspacef} f(x)^{1-\frac{1}{\gamma}} g(x)^{-\frac{1}{\gamma}} \, dx < \infty
    \end{align}
    and $\lim_{\zeta \to 0} \cost_\zeta(x)^{-1} = (f(x)g(x))^{-\frac{1}{\gamma}}$ wherever $f(x)g(x)$ is continuous, which is almost everywhere.
\end{proof}

\Cref{lem:cost-adjusting}(i) is the cost-reachable equivalent to the $b$-coverability assumption (\Cref{assm:b-coverability}) for our original (length) reachable sets; (ii) is similarly equivalent to the starting coverability assumption (\Cref{assm:starting-coverability}); (iii) is a statement on the probability weights of the cost-reachable sets; and (iv) refers to the ability of $\cost_\zeta$ to approximate $\cost^*$ (while being bounded away from $0$ and Lipschitz continuous).

\subsection{Cost-Balanced Orienteering} \label{sec:cbo}

\begin{definition} \label{def:cbo}
The \emph{cost-bounded trajectory set} of dynamic system $\Pi$ is denoted
\begin{align}
    {\Pi}^{\cost}_{\lambda} \defeq \{\pi \in \Pi : \ell^{\cost}(\pi) \leq \lambda\}
\end{align} 
Then, the \emph{Cost-Balanced Orienteering} problem is defined by
\begin{align}
    \cbo_{\Pi}(X_1, X_2, \dots, X_n; \lambda) \defeq \max_{\pi \in {\Pi}^{\cost}_{\lambda}}(|\bar{\pi} \cap \{X_i\}_{i=1}^n|)
\end{align}
i.e. the maximum number of targets which can be visited by a trajectory of cost at most $\lambda$.
\end{definition}

Note that unlike normal Orienteering or TSP, even if we treat $X_1, \dots, X_n$ as fixed, the problem depends on the density function $f$ as that influences the cost function.

We now want to show a (with very high probability) lower bound on the CBO, which can then be turned into a corresponding upper bound on the Dynamic Stochastic TSP.

\begin{proposition}[CBO upper bound]\label{thm:cbo}
Let $\Pi$ be a symmetric dynamic system and $f$ be a probability density function satisfying the assumptions in \Cref{sec:assumptions}, and let $\lambda > 0$ be fixed and sufficiently small. Then, there exists $0 < \beta$ (dependent on $\Pi$ through branching factor $b$ and small-time constraint factor $\gamma$ but not directly on agility function $g$ or density $f$) such that if $X_1, X_2, \dots X_n \simt{iid} f$, then
\begin{align}
     \cbo_{\Pi}(X_1, X_2, \dots, X_n; \lambda, f) \leq \beta \lambda \, n^{\frac{1}{\gamma}}
\end{align}
with very high probability, where $\beta$ is a constant that does not depend on $f$, $g$, or $\lambda$. 
\end{proposition}
Note that $\beta$ depends on neither $f$ nor $g$ thanks to the cost balancing: changing $f$ or $g$ (by altering the dynamic system $\Pi$) also changes $\cost_\zeta$ to preserve the balance.

We make the constants and probability bound of \Cref{thm:cbo} more explicit:
\begin{proposition} \label{prop:cbo-more-precise}
If the assumptions in \Cref{sec:assumptions} hold, then for any $\delta > 0$,
\begin{align}
    \bbP[\cbo_{\Pi}(X_1, X_2, \dots, X_n; \lambda, f) \leq (1 + \delta) \beta \lambda \, n^{\frac{1}{\gamma}}] \geq 1 - e^{-\frac{4}{5} (1+\delta) \log(b) \lambda n^{\frac{1}{\gamma}}} 
\end{align}
for sufficiently large $n$ (where `sufficiently large' can depend on $\delta$).
\end{proposition}

\begin{definition}
    Given a cost bound $\lambda > 0$, scale $\varepsilon > 0$, and approximation factor $\rho > 0$, we define a \emph{$\lambda$-cost, $\varepsilon$-scale, $\rho$-approximate representation sequence} (to be referred to in general as a \emph{representation sequence}) to be any sequence of configurations
    \begin{align}
        \bpsi = (\psi_0, \psi_1, \dots, \psi_{\lceil \lambda/((1-\rho)\varepsilon) \rceil}) ~~~&\text{where}~~~ \psi_k \in \cspace \text{ for all } k
        \\ \text{such that }~ \psi_0 \in Q^{\cost,0}_{(1-\rho)\varepsilon} \text{ and }& \psi_k \in Q^{\cost}_{(1-\rho/2)\varepsilon}(\psi_{k-1}) \text{ for all } k \geq 1
    \end{align}
    Note that we scale back by $(1-\rho)$ for the initial covering set $Q^{\cost,0}_{(1-\rho)\varepsilon}$ but only by $(1-\rho/2)$ for the $b$-covering sets $Q^{\cost}_{(1-\rho/2)\varepsilon}$.
    
    We say $\bpsi$ \emph{represents} trajectory $\pi$, denoted $\pi \reps \bpsi$, if
    \begin{align}
        \pi^{\cost}(k(1-\rho)\varepsilon) \in R^{\cost}_\varepsilon(\psi_k) \text{ for all } k \in \{0,1, \dots, n\}
    \end{align}
    We denote the set of all representation sequences (with parameters $\lambda,\varepsilon,\rho$) as $\Psi^{\cost}_{\lambda, \varepsilon,\rho}$. Furthermore, we denote the cost-reachable set of $\bpsi$ (with radius $\varepsilon$) as the union of the reachable sets of $\psi_0, \psi_1, \dots, \psi_{\lceil \lambda/((1-\rho)\varepsilon) \rceil}$, which we denote
    \begin{align}
        R^{\cost}_\varepsilon(\bpsi) \defeq \bigcup_{k=0}^{\lceil \lambda/((1-\rho)\varepsilon) \rceil} R^{\cost}_\varepsilon(\psi_k)
    \end{align}
    and we denote its workspace projection as
    \begin{align}
        \bar{R}^{\cost}_\varepsilon(\bpsi) \defeq \bigcup_{k=0}^{\lceil \lambda/((1-\rho)\varepsilon) \rceil} \bar{R}^{\cost}_\varepsilon(\psi_k) \,.
    \end{align}
\end{definition}
Note that any representation sequence may represent many (even infinitely many) different valid trajectories, and any trajectory may be represented by multiple representation sequences. We now show a few of lemmas concerning representation sequences:

\begin{lemma} \label{lem:path-representation}
    For any $\rho > 0$, there is some sufficiently small $\zeta > 0$ (which defines the cost function) and $\varepsilon^{(\rho)} > 0$ such that for all $0 < \varepsilon \leq \varepsilon^{(\rho)}$ such that any trajectory $\pi \in \Pi^{\cost}_\lambda$ such that $\pi(0) \in [\wspacef]^\cspace$ is represented by some $\bpsi \in \Psi^{\cost}_{\lambda,\varepsilon,\rho}$.
\end{lemma}

\begin{proof}
    First, we define $\zeta > 0$ sufficiently small so that $1 + \rho_2(\zeta) \leq 1/(1-\rho/2)$; we then use $\varepsilon^{(\rho)} = \rho_1(\zeta)$ (see \Cref{lem:cost-adjusting} for the definitions of $\rho_2(\zeta)$ and $\rho_1(\zeta)$). This in particular means that for any $0 < \varepsilon \leq \varepsilon^{(\rho)}$ and any $\bq \in \cspace$, there is a set $Q^{\cost}_{(1-\rho)\varepsilon}(\bq)$ such that $|Q^{\cost}_\varepsilon(\bq)| \leq b$ and
    \begin{align} \label{eq:cost-adjustment-applied}
        R^{\cost}_{(2-\rho)\varepsilon}(\bq) = R^{\cost}_{2(1-\rho/2)\varepsilon}(\bq) 
        \\ \subseteq \bigcup_{\bq' \in Q^{\cost}_{(1-\rho/2)\varepsilon}(\bq)} R^{\cost}_{(1+\rho_2(\zeta))(1-\rho/2)\varepsilon}(\bq') \subseteq \bigcup_{\bq' \in Q^{\cost}_{(1-\rho/2)\varepsilon}(\bq)} R^{\cost}_{\varepsilon}(\bq') \,.
    \end{align}
    We then construct $\bpsi \in \Psi^{\cost}_{\lambda,\varepsilon,\rho}$ such that $\bpsi \reps \pi$ inductively.
    
    \emph{Base case:} We need to choose $\psi_0$ such that $\pi(0) = \pi^{\cost}(0) \in R^{\cost}_\varepsilon(\psi_0)$. This must exist because by definition the $\varepsilon$-radius reachable sets from points in $Q^{\cost,0}_{(1-\rho)\varepsilon}$ cover $[\wspacef]^\cspace$.

    \emph{Inductive step:} For any $k > 0$, given $\psi_{k-1}$ such that $\pi^{\cost}((k-1)(1-\rho)\varepsilon) \in R^{\cost}_\varepsilon(\psi_{k-1})$, we need to choose $\psi_k \in Q^{\cost}_{(1-\rho/2)\varepsilon}(\psi_{k-1})$ such that $\pi^{\cost}(k(1-\rho)\varepsilon) \in R^{\cost}_\varepsilon(\psi_k)$. We do this by noting that by definition $\pi^{\cost}(k(1-\rho)\varepsilon) \in R^{\cost}_\varepsilon(\pi^{\cost}((k-1)(1-\rho)\varepsilon))$ (since they are only cost $(1-\rho)\varepsilon$ apart on valid trajectory $\pi$). Thus, $\pi^{\cost}(k(1-\rho)\varepsilon) \in R^{\cost}_{(2-\rho)\varepsilon}(\psi_{k-1})$ (since we can get from $\psi_{k-1}$ to $\pi((k-1)(1-\rho)\varepsilon)$ with a $\varepsilon$-cost valid trajectory, and then from $\pi((k-1)(1-\rho)\varepsilon)$ to $\pi(k(1-\rho)\varepsilon)$ with an additional $(1-\rho)\varepsilon$-cost valid trajectory, thus in total a $(2-\rho)\varepsilon$-cost valid trajectory). But then by \eqref{eq:cost-adjustment-applied} we know that $R^{\cost}_{(2-\rho)\varepsilon}(\psi_{k-1})$ is covered by the $\varepsilon$-radius cost reachable sets from points in $Q^{\cost}_{(1-\rho/2)\varepsilon}(\psi_{k-1})$, and hence we can choose the one containing $\pi^{\cost}(k(1-\rho)\varepsilon) \in R^{\cost}_{(2-\rho)\varepsilon}(\psi_{k-1})$ to be $\psi_k$, thus maintaining $\psi_k \in Q^{\cost}_{(1-\rho/2)\varepsilon}(\psi_{k-1})$ and $\pi^{\cost}(k(1-\rho)\varepsilon) \in R^{\cost}_\varepsilon(\psi_k)$.

    Thus, we can build $\psi_0, \psi_1, \dots$ like this until we have covered all of $\pi$. Since $\pi^{\cost}$ only takes in inputs from $0$ to $\lambda$, we only need to do this up to $\pi^{\cost}(\lfloor \lambda/((1-\rho)\varepsilon) \rfloor \varepsilon)$ and we are done (we may even have to add another point to the representation sequence if $\lfloor \lambda/((1-\rho)\varepsilon) \rfloor = \lceil \lambda/((1-\rho)\varepsilon) \rceil - 1$, which we can do from $Q^{\cost}_\varepsilon(\psi_{\lfloor \lambda/((1-\rho)\varepsilon) \rfloor})$). 
\end{proof}

\begin{lemma} \label{lem:path-containment}
    For any $\bpsi \reps \pi \in \Pi$, we have $\pi \subset R^{\cost}_{2\varepsilon}(\bpsi)$; if $\Pi$ is symmetric, then $\pi \subset R^{\cost}_{(3/2)\varepsilon}(\bpsi)$.
    Furthermore, $\bar{\pi} \subset \bar{R}^{\cost}_{2\varepsilon}(\bpsi)$ holds for any $\Pi$ and $\bar{\pi} \subset \bar{R}^{\cost}_{(3/2)\varepsilon}(\bpsi)$ holds for symmetric $\Pi$.
\end{lemma}

\begin{proof}
    For any $\pi^{\cost}(t)$, we can write $t = k(1-\rho)\varepsilon + \tau$ where $k$ is an integer (and is between $0$ and $\lceil \lambda/\varepsilon \rceil$) and $\tau < (1-\rho)\varepsilon < \varepsilon$. Then $\pi^{\cost}(t) \in R^{\cost}_{2\varepsilon}(\psi_k)$ since $\pi^{\cost}(k(1-\rho)\varepsilon)$ is within $\varepsilon$ cost from $\psi_k$ and $\pi^{\cost}(t)$ is within $\tau < (1-\rho)\varepsilon < \varepsilon$ cost from $\pi^{\cost}(k(1-\rho)\varepsilon)$.

    If $\Pi$ is symmetric, then we divide into two cases: (i) $\tau \leq \varepsilon/2$; (ii) $\tau > \varepsilon/2$. In case (i), we have that $\pi^{\cost}(t)$ is at most $\tau \leq \varepsilon/2$ cost (moving forwards) from $\pi^{\cost}(k(1-\rho)\varepsilon)$, which is at most $\varepsilon$ cost from $\psi_k$, hence 
    \begin{align}
        \pi^{\cost}(t) \in R^{\cost}_{(3/2)\varepsilon}(\psi_k) \subseteq R^{\cost}_{(3/2)\varepsilon}(\bpsi).
    \end{align} 
    In case (ii) we have that $\pi^{\cost}(t)$ is at most $(1-\rho)\varepsilon-\tau < \varepsilon/2$ from $\pi^{\cost}((k+1)(1-\rho)\varepsilon)$ (moving backward along $\pi$, which is allowed by symmetric $\Pi$), and $\pi^{\cost}((k+1)(1-\rho)\varepsilon)$ is at most $\varepsilon$ cost from $\psi_{k+1}$, hence
    \begin{align}
        \pi^{\cost}(t) \in R^{\cost}_{(3/2)\varepsilon}(\psi_{k+1}) \subseteq R^{\cost}_{(3/2)\varepsilon}(\bpsi).
    \end{align} 
    In both cases, we have $\pi^{\cost}(t) \in R^{\cost}_{(3/2)\varepsilon}(\bpsi)$ for any $t \in [0,\lambda]$ and hence $\pi \subseteq R^{\cost}_{(3/2)\varepsilon}(\bpsi)$.

    The statements for these sets projected into $\wspace$ then follow trivially.
\end{proof}

\begin{lemma} \label{lem:bigpsi-cardinality}
    Let $b$ be the branching factor and $P^{\cost}$ the polynomial from \Cref{lem:cost-adjusting}(ii),
    \begin{align}
        |\Psi^{\cost}_{\lambda,\varepsilon,\rho}| \leq P^{\cost}(1/((1-\rho)\varepsilon)) b^{\lceil \lambda/((1-\rho)\varepsilon) \rceil}
    \end{align}
    Additionally, for any fixed $\delta > 0$, there is some $\rho^{(\delta)} > 0$ and $\varepsilon^{(\delta)} > 0$ such that for all $0 < \varepsilon \leq \varepsilon^{(\delta)}$,
    \begin{align}
        |\Psi^{\cost}_{\lambda,\varepsilon,\rho^{(\delta)}}| \leq  b^{(1+\delta)\lambda/\varepsilon} \,.
    \end{align}
\end{lemma}

\begin{proof}
    This follows from the fact that for any $\bpsi = (\psi_0, \psi_1, \dots, \psi_{\lceil \lambda/((1-\rho)\varepsilon) \rceil}) \in \Psi^{\cost}_{\lambda,\varepsilon,\rho}$, we have $\psi_0 \in Q^{\cost,0}_\varepsilon$ and $\psi_k \in Q^{\cost}_\varepsilon(\psi_{k-1})$ for all $k > 0$. Thus, we have $|Q^{\cost,0}_{(1-\rho)\varepsilon}| \leq P^{\cost}(1/((1-\rho)\varepsilon))$ choices for $\psi_0$, and then for each $k > 0$ (iterating from $1$ to $\lceil \lambda/((1-\rho)\varepsilon) \rceil$) we have $|Q^{\cost}_{(1-\rho/2)\varepsilon}(\psi_{k-1})| \leq b$ choices. Thus the total number of ways to construct $\bpsi \in \Psi^{\cost}_{\lambda,\varepsilon,\rho}$ is at most $P^{\cost}(1/((1-\rho)\varepsilon)) b^{\lceil \lambda/((1-\rho)\varepsilon) \rceil}$.

    Finally, the approximation holds because for any $\delta > 0$, we can select $\rho^{(\delta)}$ such that $1/(1-\rho^{(\delta)}) \leq 1 + \delta/2$. We note that
    \begin{align}
        b^{(1+\delta)\lambda/\varepsilon} / b^{\lceil \lambda/((1-\rho^{(\delta)})\varepsilon) \rceil} \geq b^{ ((1+\delta) - 1/(1-\rho^{(\delta)})) \lambda/\varepsilon - 1} \geq b^{(\delta/2)\lambda/\varepsilon-1}
    \end{align}
    grows faster than any polynomial in $1/\varepsilon$, and specifically it must grow faster than $P^{\cost}(1/((1-\rho^{(\delta)})\varepsilon))$. Thus, for sufficiently small $\varepsilon$,
    \begin{align}
        |\Psi^{\cost}_{\lambda,\varepsilon,\rho^{(\delta)}}| \leq P^{\cost}(1/((1-\rho)\varepsilon)) b^{\lceil \lambda/((1-\rho^{(\delta)})\varepsilon) \rceil} \leq b^{(1+\delta)\lambda/\varepsilon}
    \end{align}
    and we are done.
\end{proof}

Furthermore, given any constant $r \geq 0$, the probability mass of any $\bar{R}^{\cost}_{r\varepsilon}(\bpsi)$ (and thus the expected number of targets that fall within it) is easy to bound:

\begin{lemma} \label{lem:prob-in-coverage}
    For any fixed $\lambda > 0$ and $\delta > 0$, there is a sufficiently small $\rho^{(\delta)} > 0$ such that for any sufficiently small $\varepsilon > 0$ and $\bpsi \in \Psi^{\cost}_{\lambda,\varepsilon,\rho^{(\delta)}}$,
    \begin{align}
        \bbP_{X \sim f}[X \in \bar{R}^{\cost}_{r\varepsilon}(\bpsi)] \leq (1 + \delta) r^\gamma \lambda \varepsilon^{\gamma-1} \,.
    \end{align}
\end{lemma}

\begin{proof}
We know that for any $\psi_k \in \cspace$,
\begin{align}
    \bbP_{X \sim f}[X \in \bar{R}^{\cost}_\varepsilon(\psi_k)] \leq (1 + o(1)) \varepsilon^\gamma
\end{align}
Therefore, using $r\varepsilon$ as the radius (which doesn't affect the fact that the $o(1)$ error term still goes to $0$ as $\varepsilon \to 0$ and only depends on $\varepsilon$) and letting $\rho^{(\delta)}$ (as in the previous lemma) satisfy $1/(1-\rho^{(\delta)}) < 1-\delta/2$, we can take a union bound. When $\varepsilon < \delta \lambda/8$, there are 
\begin{align}
    \lceil \lambda/((1-\rho^{(\delta)})\varepsilon) \rceil + 1 \leq (1+\delta/2)\lambda/\varepsilon + 2 \leq (1+(3/4)\delta)\lambda/\varepsilon 
\end{align}
possible values of $k$. We can then take $\varepsilon$ sufficiently small so that $\bbP_{X \sim f}[X \in \bar{R}^{\cost}_{r\varepsilon}(\psi_k)] \leq (1 + \delta^*) r^\gamma \varepsilon^\gamma$, where $\delta^*$ is the value (dependent on $\delta$ only) such that $(1+\delta^*)(1+(3/4)\delta) = 1+\delta$); then by the union bound, we have
\begin{align}
        \bbP_{X \sim f}[X \in \bar{R}^{\cost}_{r\varepsilon}(\bpsi)] &\leq \bigcup_{k=0}^{\lceil \lambda/((1-\rho^{(\delta)})\varepsilon) \rceil} \bbP_{X \sim f}[X \in \bar{R}^{\cost}_{r\varepsilon}(\psi_k)]
        \\ &\leq (\lceil \lambda/((1-\rho^{(\delta)})\varepsilon) \rceil + 1) (1 + \delta^*) r^\gamma \varepsilon^\gamma
        \\ &\leq (1+\delta) r^\gamma \lambda \varepsilon^{\gamma-1} 
\end{align}
as we wanted.
\end{proof}

We now define a new problem in which the goal is to find a representation sequence whose $2\varepsilon$-cost-reachable set (if $\Pi$ is nonsymmetric) or $(3/2)\varepsilon$-cost-reachable set (if $\Pi$ is symmetric) contains as many target points as possible. We formally define:

\begin{definition}
    Given length bound $\lambda > 0$, scale $\varepsilon > 0$, and approximation factor $\delta > 0$, we define the \emph{Sequence Containment Problem} ($\scp$) as follows. First, let $\rho \defeq \rho^{(\delta)}$ be sufficiently small so that \Cref{lem:bigpsi-cardinality,lem:prob-in-coverage} hold (we can take the minimum of the values necessary for each), and let $\zeta > 0$ (the cost function regularization factor) be sufficiently small so that \Cref{lem:path-representation} holds (thus defining the cost function $\cost$). If $\Pi$ is nonsymmetric, let $r = 2$; if $\Pi$ is symmetric, let $r = 3/2$; then
    \begin{align}
        \scp_{\Pi}(X_1, \dots, X_n; \lambda,\varepsilon,\delta) \defeq \max_{\bpsi \in \Psi^{\cost}_{\lambda,\varepsilon,\rho}} |\{X_1, \dots, X_n\} \cap \bar{R}^{\cost}_{r \varepsilon}(\bpsi)|
    \end{align}
\end{definition}

This replaces the optimization problem over the tricky and uncountably infinite $\Pi$ (regarded as a set of trajectories) with an optimization problem over the finite set $\Psi^{\cost}_{\lambda,\varepsilon,\rho}$. Furthermore, the new problem is an upper bound for the old one:

\begin{lemma} \label{lem:cbo-to-scp}
    For any dynamics $\Pi$, density function $f$ (which influences the cost function), cost constraint $\lambda > 0$ and scale $\varepsilon > 0$,
    \begin{align}
        \cbo_{\Pi}(X_1,\dots,X_n;\lambda) \leq \scp_{\Pi}(X_1, \dots, X_n; \lambda,\varepsilon,\delta)
    \end{align}
\end{lemma}

\begin{proof}
    This follows from \Cref{lem:path-representation,lem:path-containment}: for any $\pi \in \Pi^{\cost}_\lambda$, there exists $\bpsi \in \Psi^{\cost}_{\lambda,\varepsilon,\rho}$ such that $\bpsi \reps \pi$, and this implies $\bar{\pi} \subseteq \bar{R}^{\cost}_{r\varepsilon}(\bpsi)$. Thus,
    \begin{align}
        |\{X_1, \dots, X_n\} \cap \bar{\pi}| \leq |\{X_1, \dots, X_n\} \cap \bar{R}^{\cost}_{r\varepsilon}(\bpsi)|
    \end{align}
    and so the maximum of the former over $\pi \in \Pi$ is at most the maximum of the latter over $\bpsi \in \Psi^{\cost}_{\lambda,\varepsilon,\rho}$.
\end{proof}

We now prove \Cref{prop:cbo-more-precise}. In order to find an upper bound (\wvhp) to the $\cbo$, we want to obtain a (\wvhp) upper bound to $\scp_{\Pi}(X_1, \dots, X_n; \lambda,\varepsilon,\delta)$, which by \Cref{lem:cbo-to-scp} will then hold for the $\cbo$. Since we can set $\varepsilon > 0$ to any (sufficiently small) value, we use $\varepsilon = n^{-\frac{1}{\gamma}}$ (which will get arbitrarily small as $n \to \infty$). 

We now consider first fixing $\bpsi$ (arbitrarily) before $X_1,\dots,X_n \simt{iid} f$ are chosen and then looking at $|\{X_1, \dots, X_n\} \cap \bar{R}^{\cost}_{r\varepsilon}(\bpsi) |$ as a random variable (since it takes random inputs $X_1,\dots,X_n$). We let $p^{(\bpsi)} \defeq \bbP_{X \sim f}[X \in \bar{R}^{\cost}_{r\varepsilon}(\bpsi)]$. Plugging into \Cref{lem:prob-in-coverage} yields
\begin{align}
    p^{(\bpsi)} = \bbP_{X \sim f}[X \in \bar{R}^{\cost}_{r\varepsilon}(\bpsi)] &\leq (1 + \delta) \lambda r^\gamma n^{-(1-\frac{1}{\gamma})}
    \\ \implies \bbE_{X_i \simt{iid} f}[|\{X_1,\dots,X_n\} \cap \bar{R}^{\cost}_{r\varepsilon}(\bpsi)|] &\leq (1 + \delta) \lambda r^\gamma n^{\frac{1}{\gamma}}
\end{align}
Additionally, with $\bpsi \in \Psi^{\cost}_{\lambda,\varepsilon,\rho}$ fixed, we can let $Z_i \defeq Z^{(\bpsi)}_i \defeq \bbone\{X_i \in \bar{R}^{\cost}_{r\varepsilon}(\bpsi)\}$ and
\begin{align}
    Z \defeq Z^{(\bpsi)} \defeq \sum_{i=1}^n Z^{(\bpsi)}_i = |\{X_1,\dots,X_n\} \cap \bar{R}^{\cost}_{r\varepsilon}(\bpsi)|
\end{align}
We define $p^* = (1 + \delta) r^\gamma n^{-(1-\frac{1}{\gamma})}$. Then we know that $Z_i \simt{iid} \bern(p^{(\bpsi)})$, and that for sufficiently small $\varepsilon$ (which translates to sufficiently large $n$ since $\varepsilon \to 0$ as $n \to \infty$) we have $p^{(\bpsi)} \leq p^*$ for all $\bpsi$; thus, we can WLOG assume $p^{(\bpsi)} \leq p^*$ for all $\bpsi \in \Psi^{\cost}_{\lambda,\varepsilon,\rho}$.

We then define $Z^*_1, \dots, Z^*_i \simt{iid} \bern(p^*)$ and $Z^* = \sum_{i=1}^n Z^*_i$. Since $p^{(\bpsi)} \leq p^*$, for any $A > 0$,
\begin{align} \label{eq:idealizing-z-psi}
    \bbP[Z^{(\bpsi)} \geq A] \leq \bbP[Z^* \geq A]
\end{align}
Let $\mu^* \defeq \bbE[Z^*] = (1+\delta) r^\gamma \lambda n^{\frac{1}{\gamma}}$.

Therefore we use \Cref{lem:cbo-to-scp}, the union bound, equation \eqref{eq:idealizing-z-psi}, and \Cref{lem:bigpsi-cardinality} (in that order) to get that for all $A > 0$ (where the probabilities are over $X_i \simt{iid} f$):
\begin{align}
    \bbP\big[\cbo_{\Pi}(X_1, \dots, X_n; f,\lambda) \geq A] &\leq \bbP\big[\scp_{\Pi}(X_1, \dots, X_n; \lambda,\varepsilon,\delta) \geq A\big]
    \\ &= \bbP[\exists \, \bpsi \in \Psi^{\cost}_{\lambda,\varepsilon,\rho} : Z^{(\bpsi)} \geq A ]
    \\ &\leq \sum_{\bpsi \in \Psi^{\cost}_{\lambda,\varepsilon,\rho}} \bbP[Z^{(\bpsi)} \geq A]
    \\ &\leq |\Psi^{\cost}_{\lambda,\varepsilon,\rho}| \, \bbP[Z^* \geq A]
    \\ &\leq b^{(1+\delta)\lambda/\varepsilon} \, \bbP[Z^* \geq A]
    \\ &= b^{(1+\delta)\lambda n^{\frac{1}{\gamma}}} \, \bbP[Z^* \geq A]\label{eq:prob-of-failure}
\end{align}
for any sufficiently small $\varepsilon > 0$, i.e. sufficiently large $n$.

Since $Z^*$ is a sum of iid Bernoulli random variables, we can apply the Chernoff bound. In particular, let $A = (1+\xi) \mu^*$. Then the upper Chernoff bound says
\begin{align}
    \bbP[Z^* \geq A] &= \bbP[Z^* \geq (1+\xi) \mu^*] 
    \\ &\leq e^{-\frac{\xi^2}{2+\xi} \mu^* }
    \\ &= e^{-\frac{\xi^2}{2+\xi} (1+\delta) r^\gamma \lambda n^{\frac{1}{\gamma}} }
\end{align}
We then note that equation \eqref{eq:prob-of-failure}, which is an upper bound on the probability that the CBO problem yields a result $\geq A$, is
\begin{align}
    b^{(1+\delta)\lambda n^{\frac{1}{\gamma}}} \, \bbP[Z^* \geq A] &\leq e^{\log(b) \, (1+\delta) \, \lambda n^{\frac{1}{\gamma}} -\frac{\xi^2}{2+\xi} (1+\delta) r^\gamma \lambda n^{\frac{1}{\gamma}} }
    \\ &=  e^{\big(\log(b)  -\frac{\xi^2}{2+\xi} r^\gamma \big) \lambda (1+\delta)  n^{\frac{1}{\gamma}}}
\end{align}

We note that since $\lambda (1+\delta) > 0$, as long as $\log(b) - \frac{\xi^2}{2+\xi} r^\gamma < 0$, the above goes to $0$ as $n \to \infty$ (and does so according to $e^{-c n^{\frac{1}{\gamma}}}$, i.e. with very high probability). We now analyze $\xi$ as defined in \eqref{eq:xi2}: since $\log(b)/r^\gamma > \sqrt{\log(b)/r^\gamma} \iff \log(b)/r^\gamma > 1 \iff \log(b) > r^\gamma$, if $\log(b) > r^\gamma$ we get
\begin{align}
    \log(b) - \frac{\xi^2}{2+\xi} r^\gamma &= \log(b) - \frac{9(\log(b)/r^\gamma)^2}{2+3(\log(b)/r^\gamma)} r^\gamma
    \\ &\leq \log(b) - \frac{9(\log(b)/r^\gamma)^2}{5(\log(b)/r^\gamma)} r^\gamma
    \\ &= -\frac{4}{5} \log(b)
\end{align}
Similarly, if $\log(b) \leq r^\gamma$ we get
\begin{align}
    \log(b) - \frac{\xi^2}{2+\xi} r^\gamma &= \log(b) - \frac{9(\log(b)/r^\gamma)}{2+3\sqrt{\log(b)/r^\gamma}} r^\gamma
    \\ &\leq \log(b) - \frac{9(\log(b)/r^\gamma)}{5} r^\gamma
    \\ &= -\frac{4}{5} \log(b)
\end{align}
and hence we have in either case the bound
\begin{align}
    \log(b) - \frac{\xi^2}{2+\xi} r^\gamma \leq -\frac{4}{5} \log(b)
\end{align}

Then for sufficiently large $n$ (since $\varepsilon = n^{-\frac{1}{\gamma}}$ this is equivalent to `for sufficiently small $\varepsilon$') and letting $\{X_i\} := \{X_1, \dots, X_n\}$ we have
\begin{align}
    \bbP[\cbo_{\Pi}(\{X_i\}; f,\lambda) \geq (1+\delta) \beta \lambda n^{\frac{1}{\gamma}}] &\leq \bbP[\scp_{\Pi}(\{X_i\}; \lambda,\varepsilon,\rho) \geq (1+\delta) \beta \lambda n^{\frac{1}{\gamma}}]
    \\ &\leq |\Psi^{\cost}_{\lambda,\varepsilon,\rho}| ~ \bbP[Z^* \geq (1+\xi) \mu^*]
    \\ &\leq b^{(1 + \delta) \lambda n^{\frac{1}{\gamma}}} \, e^{\big(-\frac{\xi^2}{2+\xi} (1+\delta) r^\gamma \lambda \varepsilon^{\gamma-1} n \big)}
    \\ &\leq e^{-\frac{4}{5} (1+\delta) \log(b) \lambda n^{\frac{1}{\gamma}}}
\end{align}
thus proving \Cref{thm:cbo,prop:cbo-more-precise}.

\subsection{DSTSP lower bound}\label{sec:dstsp-from-cbo}

While the Cost-Balanced Orienteering problem can be bounded cleanly, it replaces the length of the trajectory with a cost function. Thus, the direct correspondence between Orienteering and TSP doesn't hold with CBO and we need a new technique. Let's denote
\begin{align}
    \Pi^{\{X_i\}} \defeq \{\pi \in \Pi : X_i \in \bar{\pi} \text{ for all } i\}
\end{align}
i.e. the set of all TSP solution trajectories (not necessarily the minimum length). We then take some $\pi \in \Pi^{\{X_i\}}$ and want to show that with very high probability it must be at least a certain length. We do this by chopping $\pi$ into segments of $\lambda$ cost; if $\lambda$ is sufficiently short, the cost function will be roughly constant over the span of any of these segments (since $\cost_\zeta$ is Lipschitz continuous). Thus, the length of such a segment will be at least roughly $\lambda/\cost_\zeta(x)$ where $x$ is any point on the segment. Thus, we can let $\ell_j$ be the cost of the $j$th such segment, and let $j(i)$ be the segment that $X_i$ falls in; we can then consider the sum over all $i$ of the length of the segment that $X_i$ falls in, which is
\begin{align}
    \sum_{i=1}^n \ell_{j(i)} \approx \sum_{i=1}^n \lambda/\cost_\zeta(X_i) \approx n \lambda \bbE_{X \sim f}[\cost_\zeta(X)^{-1}] \approx (1-\delta)n \lambda \int_{\wspacef} f(x)^{1-\frac{1}{\gamma}} g(x)^{-\frac{1}{\gamma}} \, dx  \,.
\end{align}
However, this may well overcount the total length since each segment is counted once per $X_i$ that falls on it; but by \Cref{prop:cbo-more-precise} we know that no such segment can have more than $(1+\delta)\beta \lambda n^{\frac{1}{\gamma}}$ targets on it. Thus, the total length is at least this sum divided by $(1+\delta)\beta \lambda n^{\frac{1}{\gamma}}$; combining the two approximation factors $\delta$ (they can both be set arbitrarily small), we end with the conclusion that for any $\delta > 0$, for all sufficiently large $n$,
\begin{align}
    \tsp_{\Pi}(X_1,\dots,X_n) \geq (1-\delta)\beta^{-1} n^{1-\frac{1}{\gamma}} \int_{\wspacef} f(x)^{1-\frac{1}{\gamma}} g(x)^{-\frac{1}{\gamma}} \, dx
\end{align}
with very high probability.

Let us fix $\lambda > 0$ generate $X_1, X_2, \dots \simt{iid} f$ (an infinite sequence of targets, of which we will look at trajectories visiting the first $n$ and let $n \to \infty$). For simplicity we define
\begin{align}
    \cbo^{(n)} := \cbo_{\Pi}(X_1, X_2, \dots, X_n; \lambda, f)
\end{align}
Let $\beta$ and $\xi$ be as defined in \Cref{def:beta}.
We will be using $(1-\delta/2)^{-1}$ instead of $1 + \delta$ for the approximation (this will make the TSP bounds more elegant to state), which are not equal but for any $\delta_1 > 0$, there is some $\delta_2 > 0$ such that $1+\delta_1 = (1-\delta_2/2)^{-1}$ and vice versa. Thus, with this alteration, \Cref{prop:cbo-more-precise} holds for all $\delta > 0$ for all sufficiently large $n$ (where `sufficiently large' may depend on $\delta$),
\begin{align}
    \bbP[\cbo^{(n)} \leq (1-\delta/2) \beta \lambda \, n^{\frac{1}{\gamma}}] \geq 1 - e^{-\frac{4}{5} \frac{\log(b) \lambda n^{\frac{1}{\gamma}}}{1-\delta/2}} \geq 1 - e^{-\frac{4}{5} \log(b) \lambda n^{\frac{1}{\gamma}}}
\end{align}
For the remainder of this section, we assume the bound on $\cbo^{(n)}$ holds; at the end we will incorporate the probability that it fails into our bound. 
Let $\pi := \pi^{(n)} \in \Pi$ be a TSP trajectory for targets $\{X_1, \dots, X_n\}$ and let $t_1, t_2, \dots, t_n \in [0,\ell(\pi)]$ satisfy $\bar{\pi}(t_i) = X_i$ ($t_i$ is the time when $\pi$ visits $X_i$, if there's more than one then choose arbitrarily). Without loss of generality we assume that $\max_i t_i = \ell(\pi)$ (a TSP trajectory has no need to continue once it has visited all $n$ targets).

We want to partition $\pi$ into cost-$\lambda$ segments. However, $\pi$ might not divide evenly into cost-$\lambda$ segments, so we define the following values:
\begin{align}
    k \defeq \lceil \ell^{\cost}(\pi)/\lambda \rceil ~~\text{ and }~~ \lambda' \defeq \ell^{\cost}(\pi)/k 
\end{align}
Thus, $\lambda - 1/k \leq \lambda' \leq \lambda$. Furthermore, since the bound on $\cbo^{(n)}$ holds, we know that any cost-$\lambda'$ trajectory can have at most $(1-\delta/2)^{-1} \beta \lambda n^{\frac{1}{\gamma}}$ targets in it, and hence to get all $n$ points we need
\begin{align}
    k \geq \frac{n}{(1-\delta/2)^{-1} \beta \lambda n^{\frac{1}{\gamma}}} = (1-\delta/2) \beta^{-1} \lambda^{-1} n^{1 - \frac{1}{\gamma}}
\end{align}
such segments. Thus, as $n \to \infty \implies k \to \infty \implies \lambda' \to \lambda$. We define $0 = t'_0 < t'_1 < \dots < t'_k = \ell(\pi)$ such that
\begin{align}
    t'_j := \min \Big(t : \int_0^t \cost_\zeta(\pi(\tau)) \, d\tau = j \lambda' \Big)
\end{align}
i.e. the trajectory reaches cost $j \lambda'$ at time $t'_j$. This also means that
\begin{align} \label{eq:lambda-segments}
    \ell^{\cost}(\pi_{[t'_{j-1}, t'_j]}) = \lambda'
\end{align}
i.e. the cost $\pi$ accumulates between any $t'_{j-1}$ and $t'_j$ is $\lambda'$. For any $j$, let $A_j := \{i : t_i \in [t'_{j-1}, t'_j)\}$ (and $A_k = \{i : t_i \in [t'_{k-1}, t'_k]\})$ so the whole interval $[0,\ell(\pi)]$ is included), i.e. $\pi$ visits $X_i$ in the time interval $[t'_{j-1}, t'_j)$. Because $\cbo^{(n)}$ is bounded and by \eqref{eq:lambda-segments}, we know that for sufficiently large $n$, all $j \in [k]$ satisfy
\begin{align} \label{eq:procrustes-error}
    |A_j| \leq (1-\delta/2)^{-1} \beta \lambda n^{\frac{1}{\gamma}} \,.
\end{align}
We also let $\ell_j \defeq \ell_j^{(\pi,\lambda)} = t'_j - t'_{j-1}$, corresponding to the length of the $j$th cost-$\lambda'$ trajectory, and define $j(i)$ such that $i \in A_{j(i)}$ ($\pi$ visits $X_i$ in the $j(i)$th cost-$\lambda'$ interval); note that it takes $\lambda$ and not $\lambda'$ as an input (and $\lambda'$ is determined by $\lambda$ and $\ell^{\cost}(\pi)$ as described above). Note that the collection $\{A_j\}_{j \in [k]}$ partitions $[n]$.

Since $\cost_\zeta(x) \geq \alpha_1$ everywhere and $\cost_\zeta$ is $\alpha_2$-Lipschitz continuous, for any $\eta > 0$ we can find $\lambda$ such that for all $\lambda' \leq \lambda$, if $\pi'$ is a trajectory with cost $\ell^{\cost}(\pi') = \lambda'$ and $X \in \bar{\pi}'$ then $\ell(\pi') \geq (1-\eta) \lambda'/\cost_\zeta(X)$. We then derive the following:
\begin{align}
    \ell(\pi) &= \sum_{j=1}^k \ell_j
    \\ &= \frac{1-\delta/2}{\beta \lambda n^{\frac{1}{\gamma}}} \sum_{j=1}^k (1-\delta/2)^{-1} \beta \lambda n^{\frac{1}{\gamma}} \ell_j
    \\ &\geq \frac{1-\delta/2}{\beta \lambda n^{\frac{1}{\gamma}}} \sum_{j=1}^k |A_j| \ell_j
    \\ &= \frac{1-\delta/2}{\beta \lambda n^{\frac{1}{\gamma}}} \sum_{i=1}^n \ell_{j(i)} \label{eq:cbo-to-dstsp-first-steps}
\end{align}
This then leads to the intuitive steps (note that we use $\lambda \approx \lambda'$ and remove the $\delta$ terms):
\begin{align}
    \ell(\pi) &\gtrsim \frac{1}{\beta \lambda n^{\frac{1}{\gamma}}} \sum_{i=1}^n \lambda/\cost_\zeta(X_i) \text{ with very high probability} \label{eq:approx-1}
    \\ &\gtrsim \frac{1}{\beta} n^{1-\frac{1}{\gamma}} \bbE_{X \sim f}[\cost_\zeta(X)^{-1}] \text{ with very high probability} \label{eq:approx-2}
    \\ &\approx \frac{1}{\beta} n^{1-\frac{1}{\gamma}} \int_{\wspacef} f(x)^{1 - \frac{1}{\gamma}} g(x)^{-\frac{1}{\gamma}} \, dx \label{eq:approx-3}
\end{align}
The two `$\gtrsim$' steps \eqref{eq:approx-1} and \eqref{eq:approx-2} need to be precisely stated and justified: \eqref{eq:approx-1} happens because the cost-$\lambda'$ segment $\pi_{[t'_{j(i)-1}, t'_{j(i)}]}$ contains $X_i$ but requires care in handling the approximation, while \eqref{eq:approx-2} is shown with a concentration bound (and because when $n$ is large $\lambda' \approx \lambda$). The approximation in \eqref{eq:approx-3} is justified by using a regularization factor $\zeta$ which is sufficiently small to yield an approximation as close as we desire.

Before we precisely state and prove the results that will give us steps \eqref{eq:approx-1} and \eqref{eq:approx-2}, we discuss our main tool for showing them, which is the \emph{one-sided Bernstein condition} and \emph{Bernstein's inequality} (see \cite{rebeschini-20}). We will not define the one-sided Bernstein condition since we don't directly use the definition; instead, we give a (known) sufficient condition, namely that a random variable which is bounded above satisfies it:
\begin{proposition} \label{prop:bern-condition}
    If $Y - \bbE[Y] \leq a'$ (guaranteed) for some $a' > 0$, then $Y$ satisfies the one-sided Bernstein condition with parameter $a = a'/3$.
\end{proposition}
This is useful because it allows us to use the (one sided) \emph{Bernstein's inequality}:
\begin{proposition} \label{prop:bern-inequality}
    Let $Y_1, \dots Y_n$ be iid random variables (each with expectation $\bbE[Y]$ and variance $\var[Y]$) satisfying the one-sided Bernstein condition with parameter $a > 0$. Then, for any $\delta > 0$,
    \begin{align}
        \bbP \Big[\frac{1}{n} \sum_{i=1}^n Y_i - \bbE[Y] \geq \delta \Big] \leq e^{-\frac{n \delta^2 / 2}{\var[Y] + a \delta}}
    \end{align}
\end{proposition}
This is usually given in terms of $Y$ bounded above because the one-sided Bernstein inequality is given as an upper tail bound (as it's typically used in that fashion); here we are trying to derive a lower tail bound of a sum of iid random variables which are bounded below (because they are all nonnegative) so we rephrase them as:
\begin{proposition} \label{prop:bern-restated}
    Let $Y_1, \dots, Y_n$ be iid nonnegative random variables (each with expectation $\bbE[Y]$ and variance $\var[Y]$). Then for any $\delta > 0$,
    \begin{align}
        \bbP \Big[ \sum_{i=1}^n Y_i \leq (1 - \delta) n \bbE[Y]  \Big] \leq e^{-\frac{n \bbE[Y]^2 \delta^2 / 2}{\var[Y] + \bbE[Y]^2 \delta/2}}
    \end{align}
\end{proposition}

\begin{proof}
    By \Cref{prop:bern-condition}, if $Y$ is nonnegative then $-Y$ satisfies the Bernstein condition with parameter $a = \bbE[Y]/3$. Then we can re-write
    \begin{align}
        \sum_{i=1}^n Y_i \leq (1 - \delta) n \bbE[Y] &\iff \sum_{i=1}^n (-Y_i) \geq (1 - \delta) n \bbE[-Y]
        \\ &\iff \frac{1}{n} \sum_{i=1}^n (-Y_i) - \bbE[-Y] \geq \delta \, \bbE[Y]
    \end{align}
    Plugging this into \Cref{prop:bern-inequality} then yields the result.
\end{proof}

Let $Y_i \defeq \cost_\zeta(X_i)^{-1} \leq 0$, and if $X \sim f$ we define $Y \defeq \cost_\zeta(X)^{-1} \leq 0$ (the generic version of $Y_i$). We need to establish a few things:
\begin{lemma} \label{lem:EY-finite}
    For any approximation factor $\delta < 0$, there is a sufficiently small regularization factor $\zeta \defeq \zeta^{(\delta)} > 0$ so that $\bbE[Y]$ and $\var[Y]$ satisfy
    \begin{align}
        (1-\delta)\int_{\wspacef} f(x)^{1-\frac{1}{\gamma}} g(x)^{-\frac{1}{\gamma}} \, dx  \leq \bbE[Y] \leq \int_{\wspacef} f(x)^{1-\frac{1}{\gamma}} g(x)^{-\frac{1}{\gamma}} \, dx 
        \\ \text{and }~~ \var[Y] \leq \int_{\wspacef} f(x)^{1-\frac{2}{\gamma}} g(x)^{-\frac{2}{\gamma}} \, dx - (1-\delta)^2 \bigg(\int_{\wspacef} f(x)^{1-\frac{1}{\gamma}} g(x)^{-\frac{1}{\gamma}} \, dx \bigg)^2
    \end{align}
    which are both finite when $\gamma \geq 2$. Additionally, even when $\gamma < 2$, we can bound the expected value and the variance above using the regularization factor itself:
    \begin{align}
        (1-\delta)\int_{\wspacef} f(x)^{1-\frac{1}{\gamma}} g(x)^{-\frac{1}{\gamma}} \, dx  \leq \bbE[Y] &\leq \alpha_1^{-1}
        \\ \text{and }~~ \var[Y] &\leq \alpha_1^{-2}/4 \,.
    \end{align}
\end{lemma}

\begin{proof}
    We recall that the lucrativity function (of which the cost function $\cost_\zeta$ is an approximation) is $\cost^*(x) = (f(x)g(x))^{\frac{1}{\gamma}}$ and that $\gamma \geq 2$. Let $Y^* = \cost^*(X)$ where $X \sim f$. Then we have
    \begin{align}
        \bbE[Y^*] &= \int (\cost^*)^{-1} \, df
        \\ &= \int_{\wspacef} f(x) (f(x)g(x))^{-\frac{1}{\gamma}}\, dx
        \\ &= \int_{\wspacef} f(x)^{1-\frac{1}{\gamma}} g(x)^{-\frac{1}{\gamma}} \, dx
        \\ &\leq \int_{\wspacef} (1 + f(x)) g_{\min}^{-\frac{1}{\gamma}} \, dx \label{eq:this-equation-001}
        \\ &= (\vol_{\wspace}(\wspacef) + 1) g_{\min}^{-\frac{1}{\gamma}}
        \\ &< \infty
    \end{align}
    where \eqref{eq:this-equation-001} follows because $g(x) \geq g_{\min}$ for all $x$ and
    \begin{align}
        f(x)^{1-\frac{1}{\gamma}} \leq \max(1, f(x)) \leq 1 + f(x)
    \end{align}
    (and $\int_{\wspacef} f(x) \, dx = 1$ by definition). Similarly,
    \begin{align}
        \bbE[(Y^*)^2] &= \int_{\wspacef} f(x)^{1-\frac{2}{\gamma}} g(x)^{-\frac{2}{\gamma}} \, dx
        \\ &\leq \int_{\wspacef} (1 + f(x)) g_{\min}^{-\frac{2}{\gamma}} \, dx \label{eq:this-equation-002}
        \\ &= (\vol_{\wspace}(\wspacef) + 1) g_{\min}^{-\frac{2}{\gamma}}
        \\ &< \infty
    \end{align}
    Thus, both $\bbE[Y^*]$ and $\bbE[(Y^*)^2]$ are finite when $\gamma \geq 2$ (since we rely on $1-\frac{2}{\gamma} \geq 0$); but by definition $\cost_\zeta(x) \geq \cost^*(x)$ and hence if $\gamma \geq 2$, we have $\cost_\zeta(x)^{-1} \leq \cost^*(x)^{-1}$ and so $\bbE[Y] \leq \bbE[Y^*] < \infty$ and $\bbE[Y^2] \leq \bbE[(Y^*)^2] < \infty$ and hence both $\bbE[Y]$ and $\var[Y] = \bbE[Y^2]-\bbE[Y]^2 \leq \bbE[(Y^*)^2] - (1-\delta)^2 \bbE[Y^*]^2$ are finite.

    When $\gamma < 2$, we then use the fact that by construction $\cost_\zeta(x) \geq \alpha_1$, and hence $\bbE[\cost_\zeta(X)^{-1}] \leq \alpha_1^{-1}$ and $\var[\cost_\zeta(X)^{-1}] \leq \alpha_1^{-2}/4$ (since $\cost_\zeta(X)^{-1} \in (0,\alpha_1^{-1}]$ its variance is at most $\alpha_1^{-2}/4$). Thus, we are done.
\end{proof}

Returning to the main proof, we first address step \eqref{eq:approx-1}:
\begin{lemma}
For any $\delta_1 > 0$, there is a sufficently small $\lambda > 0$ such that (when $X_i \simt{iid} f$),
\begin{align}
    \bbP \Big[\exists \, \pi \in \Pi^{\{X_i\}} \text{ s.t. } \sum_{i=1}^n \ell_{j(i)} < (1 - \delta_1) \lambda n \bbE[Y] \Big] \leq e^{-\frac{n (1-\delta_1) \bbE[Y]^2 \delta_1^2/8}{(1+\delta_1)\var[Y] + (1-\delta_1) \bbE[Y]^2 \delta_1/6}}
\end{align}
for all sufficiently large $n$.
\end{lemma}

\begin{proof}
For any $x \in \wspacef$ let
\begin{align}
    \ell^*(x;\lambda) = \min_{\pi' \in \Pi} (\ell(\pi') : \ell^{\cost}(\pi') = \lambda, x \in \bar{\pi}')
\end{align}
i.e. the length of the shortest possible cost-$\lambda$ trajectory through $x$. 

Recall that for any $\eta > 0$, there is some $\lambda_{\eta}^* > 0$ such that for any $0 < \lambda \leq \lambda_{\eta}^*$,
\begin{align}
    \bbP_{X \sim f}[\exists \, \pi' \in \Pi : X \in \bar{\pi}', ~\ell^{\cost}(\pi') = \lambda \text{ and } \ell(\pi') < (1 - \eta) \lambda/\cost_\zeta(X) ] \leq \eta
\end{align}
We define $\wspace(\eta) \subseteq \wspacef$ to be the region in which
\begin{align}
    x \in \bar{\pi}', ~\ell^{\cost}(\pi') = \lambda \implies \ell(\pi') \geq (1 - \eta) \lambda/\cost_\zeta(x)
\end{align}
for $\lambda \leq \lambda^*_{\eta}$ breaks down, i.e.
\begin{align}
    \wspace(\eta) = \{x \in \wspacef : \exists \, \pi' \in \Pi : x \in \bar{\pi}', ~\ell^{\cost}(\pi') \leq \lambda^*_{\eta} \text{ and } \ell(\pi') < (1 - \eta) \ell^{\cost}(\pi')/\cost_\zeta(x)\}
\end{align}
Then, if we have some $x \in \wspacef$ and a trajectory $\pi'$ s.t. $x \in \bar{\pi}'$ and $\ell^{\cost}(\pi') = \lambda < \lambda^*_{\eta}$,
\begin{align}
    \ell(\pi') \geq \begin{cases} (1-\eta) \lambda'/\cost_\zeta(x) &\text{if } x \not \in \wspace(\eta) \\ 0 &\text{if } x \in \wspace(\eta) \end{cases}
\end{align}
Therefore, we define the random variables
\begin{align}
    Z_i \defeq Z_i^{(\eta)} = \begin{cases} (1-2\eta) \lambda/\cost_\zeta(X_i) &\text{if } X_i \not \in \wspace(\eta) \\ 0 &\text{if } X_i \in \wspace(\eta) \end{cases}
\end{align}
Since these are iid (depending only on iid $X_i$) let $Z$ be the generic version. We know from \eqref{eq:procrustes-error} that
\begin{align}
    \lambda' &\geq \lambda - 1/k \geq (1 - \beta n^{-(1-\frac{1}{\gamma})}) \lambda
    \\ \implies (1-\eta)\lambda' &\geq (1-2\eta)\lambda ~~\text{ for }~~ n \geq (\beta/\eta)^{1 + \frac{1}{\gamma-1}}
\end{align}
Since $\ell_{j(i)}$ is a cost-$\lambda'$ trajectory passing through $X_i$, we know (fixing $\eta$) that for any sufficiently large $n$
\begin{align}
    \ell_{j(i)} \geq Z_i \text{ for all } i
\end{align}
Additionally, since $\bbP_{X \sim f}[X \in \wspace(\eta)] \leq \eta$ and because $\bbE[Y] = \bbE_{X \sim f}[\cost_\zeta(X)^{-1}]$ and $\bbE[Y^2] = \bbE_{X \sim f}[\cost_\zeta(X)^{-2}]$ are finite, we have
\begin{align}
    \lim_{\eta \to 0} \bbE[Z^{(\eta)}] = \lambda \bbE[Y] ~~\text{ and }~~ \lim_{\eta \to 0} \var[Z^{(\eta)}] = \lambda \var[Y]
\end{align}
Therefore, for any $\delta_1 > 0$ we can select some $\eta > 0$ such that
\begin{align}
    \bbE[Z^{(\eta)}] \geq (1-\delta_1/2) \lambda \bbE[Y] \text{ and } \bbE[Z^{(\eta)}] \leq (1+\delta_1) \lambda \var[Y]
\end{align}
Fixing such $\eta$, we apply \Cref{prop:bern-restated} (noting that $Z_i$'s are nonnegative iid) to get
\begin{align}
    \bbP \Big[ \sum_{i=1}^n Z_i \leq (1 - \delta_1/2) n \bbE[Z]  \Big] &\leq e^{-\frac{n \bbE[Z]^2 (\delta_1/2)^2 / 2}{\var[Z] + \bbE[Z]^2 (\delta_1/2)/3}}
    \\ \implies \bbP \Big[ \sum_{i=1}^n Z_i \leq (1 - \delta_1) \lambda n \bbE[Y]  \Big] &\leq e^{-\frac{n (1-\delta_1) \bbE[Y]^2 \delta_1^2/8}{(1+\delta_1)\var[Y] + (1-\delta_1) \bbE[Y]^2 \delta_1/6}}
    \\ \implies \bbP \Big[ \sum_{i=1}^n \ell_{j(i)} \leq (1 - \delta_1) \lambda n \bbE[Y]  \Big] &\leq e^{-\frac{n (1-\delta_1) \bbE[Y]^2 \delta_1^2/8}{(1+\delta_1)\var[Y] + (1-\delta_1) \bbE[Y]^2 \delta_1/6}}
\end{align}
for all $\lambda < \lambda^*_{\eta}$, which holds since we can apply
\begin{align}
    \bbE[Z]^2 \geq (1-\delta_1/2)^2 \lambda \bbE[Y] \geq (1 - \delta_1) \lambda \bbE[Y] ~~\text{ and }~~ \var[Z] \leq (1+\delta_1) \var[Y]
\end{align}
(because of the $\var[Z] > 0$ term, the first substitution on the numerator and the denominator makes the fraction smaller, hence the exponential of the negative of the fraction larger).
\end{proof}

But this means that with very high probability, given the CBO bound and applying \eqref{eq:cbo-to-dstsp-first-steps} (and using $1+\delta_1 < (1+\delta_1/2)^2$)
\begin{align}
    &\sum_{i=1}^n \ell_{j(i)} > (1-\delta_1/2) \lambda n \bbE[Y]
    \\ \implies &\ell(\pi) \geq \frac{1-\delta_1/2}{\beta \lambda n^{\frac{1}{\gamma}}} \sum_{i=1}^n \ell_{j(i)}  > (1-\delta_1) \beta^{-1} n^{1-\frac{1}{\gamma}} \bbE[Y]
\end{align}
with very high probability. Specifically (taking into account the probability that the CBO bound holds) we have
\begin{align}
    \bbP \big[\tsp_{\Pi}(\{X_i\}) \geq (1-\delta_1) \beta^{-1} n^{1-\frac{1}{\gamma}} \bbE[Y] \big] 
    \\ \geq 1 - e^{-\frac{4}{5} \frac{\log(b) \lambda n^{\frac{1}{\gamma}}}{1-\delta_1/2}} - e^{-\frac{n (1-\delta_1/2) \bbE[Y]^2 \delta_1^2/32}{(1+\delta_1/2)\var[Y] + (1-\delta_1/2) \bbE[Y]^2 \delta_1/12}}
\end{align}
We then do a few simplifications on the exponent of the final term (ignoring the constant multiple of $n$ and the negative to make the expression cleaner):
\begin{align}
    \frac{(1-\frac{\delta_1}{2}) \bbE[Y]^2 \frac{\delta_1^2}{32}}{(1+\frac{\delta_1}{2})\var[Y] + (1-\frac{\delta_1}{2}) \bbE[Y]^2 \frac{\delta_1}{12}} = \frac{(1-\frac{\delta_1}{2}) \bbE[Y]^2 \frac{\delta_1^2}{32}}{(1+\frac{\delta_1}{2})(\bbE[Y^2] - \bbE[Y]^2) + (1-\frac{\delta_1}{2}) \bbE[Y]^2 \frac{\delta_1}{12}}
    \\ = \bigg(\frac{(1+\frac{\delta_1}{2})(\bbE[Y^2] - \bbE[Y]^2) + (1-\frac{\delta_1}{2}) \bbE[Y]^2 \frac{\delta_1}{12}}{(1-\frac{\delta_1}{2}) \bbE[Y]^2 \frac{\delta_1^2}{32}}\bigg)^{-1}
    \\ = \bigg(\frac{1+\frac{\delta_1}{2}}{(1-\frac{\delta_1}{2}) \frac{\delta_1^2}{32}} \frac{\bbE[Y^2]}{\bbE[Y]^2} + \frac{(1-\frac{\delta_1}{2}) \frac{\delta_1}{12}-(1+\frac{\delta_1}{2})}{(1-\frac{\delta_1}{2}) \frac{\delta_1^2}{32}}\bigg)^{-1} \label{eq:big-ol-exponent}
\end{align}
This can be left as it is, but if we have $\bbE[Y^2]/\bbE[Y]^2$ and $\delta_1$ is very small, we can write it in its most significant terms as
\begin{align}
    \bigg(\frac{1+\frac{\delta_1}{2}}{(1-\frac{\delta_1}{2}) \frac{\delta_1^2}{32}} \frac{\bbE[Y^2]}{\bbE[Y]^2} + \frac{(1-\frac{\delta_1}{2}) \frac{\delta_1}{12}-(1+\frac{\delta_1}{2})}{(1-\frac{\delta_1}{2}) \frac{\delta_1^2}{32}}\bigg)^{-1} &= \frac{\delta_1^2}{32(\bbE[Y^2]/\bbE[Y]^2 - 1)} + o(\delta_1^3)
    \\ &= \frac{\delta_1^2}{32(\var[Y]/\bbE[Y]^2)} + o(\delta_1^3)
\end{align}
Then, given a desired approximation factor $1-\delta$, we choose $\delta = \delta_1/2$ and a regularization term $\zeta > 0$ producing 
\begin{align}
    \bbE[Y] \geq (1-\delta_2) \int_{\wspacef} f(x)^{1-\frac{1}{\gamma}} g(x)^{-\frac{1}{\gamma}} \, dx
\end{align}
where $\delta_2 = 1 - (1-\delta)/(1-\delta/2)$, i.e. such that
\begin{align}
    1 - \delta = (1-\delta/2)(1-\delta_2) = (1-\delta_1)(1-\delta_2)
\end{align}
Thus, our bound becomes
\begin{align}
    (1-\delta_1) \beta^{-1} n^{1-\frac{1}{\gamma}} \bbE[Y] = (1-\delta) \beta^{-1} n^{1-\frac{1}{\gamma}} \int_{\wspacef} f(x)^{1-\frac{1}{\gamma}} g(x)^{-\frac{1}{\gamma}} \, dx
\end{align}
and substituting $\delta_1 = \delta/2$ into the very-high-probability bound yields
\begin{align}
    \bbP \bigg[\tsp_{\Pi}(\{X_i\}) \geq (1-\delta) &\beta^{-1} n^{1-\frac{1}{\gamma}} \int_{\wspacef} f(x)^{1-\frac{1}{\gamma}} g(x)^{-\frac{1}{\gamma}} \, dx \bigg] 
    \\ &\geq 1 - e^{-\frac{4}{5} \frac{\log(b) \lambda n^{\frac{1}{\gamma}}}{1-\delta/4}} - e^{-\frac{n (1-\delta/4) \bbE[Y]^2 \delta^2/128}{(1+\delta/4)\var[Y] + (1-\delta/4) \bbE[Y]^2 \delta/24}}
    \\ &\geq 1 - e^{-\frac{4}{5} \log(b) \lambda n^{\frac{1}{\gamma}}} - e^{-\big(\frac{\delta^2}{128(\var[Y]/\bbE[Y]^2)} + o(\delta^3)\big)n}
\end{align}
when $n$ is sufficiently large. When $\gamma \geq 2$, this can be made to be arbitrarily close to
\begin{align}
    \bbP \bigg[\tsp_{\Pi}(\{X_i\}) \geq (1-\delta) &\beta^{-1} n^{1-\frac{1}{\gamma}} \int_{\wspacef} f(x)^{1-\frac{1}{\gamma}} g(x)^{-\frac{1}{\gamma}} \, dx \bigg] 
    \\ &\geq 1 - e^{-\frac{4}{5} \frac{\log(b) \lambda n^{\frac{1}{\gamma}}}{1-\delta/2}} - e^{-\frac{n (1-\delta/2) \bbE[Y^*]^2 \delta^2/32}{(1+\delta/2)\var[Y^*] + (1-\delta/2) \bbE[Y^*]^2 \delta/12}}
    \\ &\geq 1 - e^{-\frac{4}{5} \log(b) \lambda n^{\frac{1}{\gamma}}} - e^{-\big(\frac{\delta^2}{32(\var[Y^*]/\bbE[Y^*]^2)} + o(\delta^3)\big)n}
\end{align}
where $\bbE[Y^*]$ and $\var[Y^*]$ are $\int_{\wspacef} f(x)^{1-\frac{1}{\gamma}} g(x)^{-\frac{1}{\gamma}} \, dx$ and $\int_{\wspacef} f(x)^{1-\frac{2}{\gamma}} g(x)^{-\frac{2}{\gamma}} \, dx$ respectively, with the downside that this may increase how large $n$ needs to be in order to be `sufficiently large' (in particular, it's achieved by taking $\delta_1 \to \delta$ and $\delta_2 \to 0$ to compensate, which requires cost regularization factor $\zeta$ to go to $0$). For sufficiently large $n$ the dominant error term is always $-e^{-\frac{4}{5}\log(b)\lambda n^{\frac{1}{\gamma}}}$ since the other error term's exponent scales with $n$, but when $\delta$ is very small the value of $n$ at which this term takes over may be extremely large.

Finally, we can state it in the following maximally simple terms: for any $\delta > 0$, there is some sufficiently small $\lambda > 0$ and sufficiently large $n^{(\delta)}$ such that for all $n \geq n^{(\delta)}$,
\begin{align}
    \bbP \bigg[\tsp_{\Pi}(\{X_i\}) \geq (1-\delta) \beta^{-1} n^{1-\frac{1}{\gamma}} \int_{\wspacef} f(x)^{1-\frac{1}{\gamma}} g(x)^{-\frac{1}{\gamma}} \, dx \bigg] &\geq 1 - (1+\delta) e^{-\frac{4}{5} \frac{\log(b) \lambda n^{\frac{1}{\gamma}}}{1-\delta/2}}
    \\ &\geq 1 - (1+\delta) e^{-\frac{4}{5} \log(b) \lambda n^{\frac{1}{\gamma}}}
\end{align}
with the caveat that as $\delta \to 0$, $n^{(\delta)}$ potentially goes to $\infty$.



\section{DSTSP upper bound for symmetric systems}\label{sec:dstsp-ub}

As in \cite{adler-icra-16}, we show an upper bound via an algorithm which, with very high probability, creates a trajectory $\pi \in \Pi$ which successfully visits the targets $X_1, \dots, X_n$ with a trajectory of length at most $\Theta (n^{1-\frac{1}{\gamma}})$.
However, while the $\Theta(n^{1-\frac{1}{\gamma}})$ growth rate holds in general, the more precise growth rate of 
\begin{align}
    \tsp_{\Pi}(X_1, \dots, X_n) \approx c n^{1-\frac{1}{\gamma}} \Big (\int_{\wspacef} f(x)^{1-\frac{1}{\gamma}} g(x)^{-\frac{1}{\gamma}} \, dx\Big)  ~~~~\wvhp
\end{align}
does not hold in general for non-symmetric systems. Hence, in this section we deal exclusively with the case where $\Pi$ is symmetric.


\subsection{The Hierarchical Collection Problem}

Unlike in the general DSTSP algorithms, however, the symmetric property of the control system allows a tile to be completely cleared of target points before the algorithm moves on to the next one, since after visiting the one target the system can return to the anchor and go to the next one and so on.

For any $0 < \varepsilon_0 < \varepsilon^*$ (where $\varepsilon^*$ is the value given in \Cref{assm:b-coverability}), $\bq_0 \in \cspace$, and $\delta > 0$, we define the \emph{$\varepsilon_0$-scale hierarchical cell structure rooted at $\bq_0$} as an infinite rooted tree $T(\bq_0,\varepsilon_0)$ in the following way:
\begin{enumerate}
    \item The nodes represent reachable sets / cells, each with an anchor $\bq$ and a radius $\varepsilon$. We denote the node as $v_{(\bq,\varepsilon)}$, which represents the reachable set $\bar{R}_{\varepsilon}(\bq)$.
    \item The children of $v_{(\bq,\varepsilon)}$ are
    \begin{align}
        C(v_{(\bq,\varepsilon)}) = \{v_{(\bq^{(1)},\varepsilon/2)}, \dots, v_{(\bq^{(b)}, \varepsilon/2)}\} ~~~\text{satisfying}~~~ R_{\varepsilon}(\bq) \subseteq \bigcup_{j=1}^{b} R_{\varepsilon/2}(\bq^{(j)})
    \end{align}
    as given in \Cref{assm:b-coverability}. We also assume WLOG that
    \begin{align}
        R_{\varepsilon}(\bq) \cap R_{\varepsilon/2}(\bq^{(j)}) \neq \emptyset ~~~\text{for all}~~ j
    \end{align}
    (if not, we remove $v_{(\bq^{(j)},\varepsilon/2)}$ from the tree).
\end{enumerate}
If $v' \in C(v)$ we say $v = P(v')$ (the parent). 

We define the \emph{layers} of the tree in the normal way, with the root $v_{(\bq_0,\varepsilon_0)}$ being layer $0$, its children being layer $1$, and so forth (note that all nodes at layer $k$ represent reachable sets with radius $\varepsilon_0/2^k$). We denote these layers as
\begin{align}
    T_k \defeq \{v_{(\bq,\varepsilon)} \in T(\bq_0,\varepsilon_0) : \varepsilon = \varepsilon_0/2^k\}.
\end{align}
For any $x \in \wspace$ and $v_{(\bq,\varepsilon)} \in T(\bq_0,\varepsilon_0)$, we say `$x \in v_{(\bq,\varepsilon)}$' if $x \in \bar{R}_\varepsilon(\bq)$ (slightly abusing notation by conflating the node $v$ and the region it represents).

\begin{lemma}
    For any $T(\bq_0,\varepsilon_0)$ the following hold:
    \begin{enumerate}[i.]
        \item $|T_k| \leq b^k$.
        \item For any $v_{(\bq,\varepsilon)} \in T(\bq_0,\varepsilon_0)$ and any $v_{(\bq', \varepsilon/2)} \in C(v_{(\bq,\varepsilon)})$ (any parent-child pair),
        \begin{align} \label{eq:parent-child-connection}
            d_{\Pi}(\bq, \bq') \leq 3\varepsilon/2 \,.
        \end{align} 
        \item For any $v_{(\bq,\varepsilon)} \in T(\bq_0,\varepsilon_0)$,
        \begin{align}
            d_{\Pi}(\bq_0, \bq) \leq 3 \varepsilon_0
        \end{align}
    \end{enumerate}
\end{lemma}

\begin{proof}
We show these in order:

\emph{(i)} $T_k = \bigcup_{v \in T_{k-1}} C(v)$ and $|C(v)| \leq b$, so $|T_k| \leq b |T_{k-1}|$; recursing back to $|T_0| = |\{\bq_0\}| = 1$ yields the result.

\emph{(ii)} Noting that $R_\varepsilon(\bq) \cap R_{\varepsilon/2}(\bq') \neq \emptyset$ (by \eqref{eq:parent-child-connection}, let $\bq'' \in R_\varepsilon(\bq) \cap R_{\varepsilon/2}(\bq')$. Then by definition $d_{\Pi}(\bq, \bq'') \leq \varepsilon$ and $d_{\Pi}(\bq', \bq'') \leq \varepsilon/2$ and (since the control system $\Pi$ is symmetric, $d_\Pi$ is a metric) we get by the triangle inequality that
\begin{align}
    d_{\Pi}(\bq, \bq') \leq d_{\Pi}(\bq, \bq'') + d_{\Pi}(\bq', \bq'') \leq 3\varepsilon/2
\end{align}

\emph{(iii)} This follows from (ii): let $v_{(\bq,\varepsilon)} \in T_k$, so $\varepsilon = \varepsilon_0/2^k$. This means that, letting $\bq_k \defeq \bq$, we can find $\bq_1, \dots, \bq_{k-1}$ such that $v_j \defeq v_{(\bq_j, \varepsilon_0/2^j)} \in C(v_{(\bq_{j-1},\varepsilon_0/2^{j-1})}) = C(v_{j-1})$ for all $j = 1, 2, \dots, k$ (tracing the path from the root $v_0 = v_{(\bq_0,\varepsilon_0)}$ to $v_{(\bq,\varepsilon)}$). Then by (ii) we have
\begin{align}
    d_{\Pi}(\bq_0,\bq_k) &\leq \sum_{j=1}^k d_{\Pi}(\bq_{j-1},\bq_j) \leq \sum_{j=1}^k 3 \varepsilon_0 2^{-j} \leq 3 \varepsilon_0
\end{align}
Hence we are done.
\end{proof}
Let $n$ target points $X_1,\dots,X_n$ be in $\bar{R}_{\varepsilon_0}(\bq_0)$ (the area covered by this tree essentially). Each target point $X_1, \dots, X_n$ is contained (at least) one of the nodes at level $k$ (at each $k$). Specifically, for any $i \in [n], k \in \bbZ_{\geq 0}$ let $v_k(X_i) \defeq v_{(\bq_k(X_i),\varepsilon_0/2^k)} \in T(\bq_0,\varepsilon_0)$ satisfying the following properties:
\begin{itemize}
    \item $X_i \in v_k(X_i)$ for all $i,k$ (i.e. $X_i \in \bar{R}_{\varepsilon_0/2^k}(\bq_k(X_i))$);
    \item $v_0(X_i) = v_{(\bq_0,\varepsilon_0)}$ is the root for all $i$;
    \item $v_k(X_i) \in C(v_{k-1}(X_i))$ for all $i \in [n]$ and $k \in \bbZ_{> 0}$.
\end{itemize}
Note that this means $v_k(X_i) \in T_k$. Such a set of nodes exists because of the structure of $T(\bq_0,\varepsilon_0)$, specifically that the reachable sets of the nodes in $C(v)$ cover the reachable set of $v$.

This suggests a `depth-first-search' type algorithm for visiting $X_1, \dots, X_n$. For any $v \in T_k$ let
\begin{align}
    n(v) \defeq |\{i : v_k(X_i) = v \}|
\end{align}
i.e. the number of target points $X_i$ whose layer-$k$ node is $v$. We travel from node to node along the tree $T(\bq_0,\varepsilon_0)$. If we describe the system as being `at' node $v_{(\bq,\varepsilon_0/2^k)} \in T_k$, it refers to being at $\bq$, from which we can do three things:
\begin{enumerate}
    \item travel to its parent $P(v)$ (taking at most $3 \varepsilon_0 2^{-k}$ time);
    \item travel to a child $v' \in C(v)$ (taking at most $3 \varepsilon_0 2^{-(k+1)}$ time);
    \item visit some target point $X_i \in v$ and returning (taking at most $\varepsilon_0 2^{-(k-1)}$ time since $X_i \in \bar{R}_{\varepsilon_0/2^k}(\bq)$).
\end{enumerate}
This allows us to state our TSP as an abstract problem similar to the classic TSP on graphs (but slightly different); since all distance upper bounds above are multiples of $\varepsilon_0$, we ignore it (and re-insert it after we find the solution).

\begin{definition}
The \emph{hierarchical collection problem (HCP)} with cell branching factor $\bar{b}$ and scaling factor $s$ is an abstract problem on an infinite rooted tree $T = (V,E)$ where every $v \in V$ has $\bar{b}$ children. We define:
\begin{enumerate}
    \item $\gamma = \log_s(\bar{b})$, which defines the relationship between $s$ and $\bar{b}$ (in the Hierarchical Cell Structure this problem models, scaling the cells down by a factor of $s$ results in a $\bar{b}$-factor increase in the number needed to cover the same region).
    \item The root of $T$ is $v_0$; the children of $v \in V$ are denoted as the set $C(v)$ and its parent is $P(v)$ and the edges incident to $v$ is denoted $E_v$.
    \item The \emph{level} of a node is its distance from $v_0$ and is denoted $k(v)$ (so $k(v_0) = 0$ and for all $v \in V$ and $v' \in C(v)$ we have $k(v') = k(v)+1$); the set of nodes of level $k$ is denoted $L_k \defeq \{v \in V : k(v) = k\}$.
    \item Each edge $e = (v,v') \in E$ has a cost
    \begin{align}
        w(e) = s^{-\min(k(v),k(v'))}
    \end{align}
    i.e. if $e$ connects level $k$ to level $k+1$ then its cost is $s^{-k}$.
\end{enumerate}
Finally, we have $n$ \emph{targets} $\tau_1, \dots, \tau_n$; these are not nodes on the tree but are infinitely long paths starting at the root and continuing down the tree. We say $v \in \tau_i$ if $v$ is in the path $\tau_i$, and $v_k(\tau_i)$ is the level-$k$ node of $\tau_i$, and we assume that $\tau_i \neq \tau_j$ (i.e. they don't represent the same infinite path) for all $i,j \in [n]$.

The problem is solved by a \emph{player} who moves around $T$ and collects targets. When at $v \in V$, the player chooses one of two actions, for an associated cost:
\begin{enumerate}
    \item move along some $e \in E_v$ for cost $w(e)$ (i.e. going up costs $s^{-(k-1)}$, going down costs $s^{-k}$);
    \item collect a target $\tau_i$ such that $v \in \tau_i$ for cost $2 \cdot s^{-k}$.
\end{enumerate}
Then, starting at $v_0$, the goal is to collect all $n$ targets $\tau_i$ and return to $v_0$ for minimum total cost. We denote the minimum cost to collect $\tau_1, \dots, \tau_n$ as $\hcp(\tau_1, \dots, \tau_n; \bar{b},s)$, and the maximum cost to collect $n$ targets (i.e. $\tau_1, \dots, \tau_n$ are adversarially chosen) as
\begin{align}
    \hcp^*(n; \bar{b},s) \defeq \max_{\tau_1, \dots, \tau_n} \hcp(\tau_1, \dots, \tau_n; \bar{b},s) \,.
\end{align}
\end{definition}

The targets $\tau_1, \dots, \tau_n$ in the HCP represent which cells the targets $X_1, \dots, X_n$ fall into in the symmetric DTSP (we refer here to the DTSP and not the DSTSP because the targets will be treated as given, not randomly distributed, for now).

\begin{remark} \label{rmk:bhcp-01}
The HCP isn't exactly equivalent to the original symmetric DTSP because
\begin{enumerate}[a.]
    \item all the costs are derived from upper bounds to the equivalent costs in the DTSP; 
    \item the DTSP doesn't require starting or ending at any particular point; and (most importantly)
    \item it restricts possible solutions to only those that follow the hierarchical structure.
\end{enumerate}
\end{remark}

The general tradeoff that this problem explores is that collecting targets gets cheaper as the player moves down the tree $T$, but this movement has costs and the targets can get more dispersed as the tree branches out (note that all the targets are present at the root, though collecting them is the most expensive there). This mirrors the TSP on the cell structure in the following way:

\begin{proposition}
    Let $S : \cspace \times (0,\varepsilon^*]$ be a hierarchical cell structure with cell branching factor $\bar{b}$, scaling factor $s$, and efficiency factor $\alpha$ (see \Cref{def:hcs}) and let $X_1, \dots, X_n$ be targets in $S(\bq_0, \varepsilon_0)$. We then set up an infinite rooted tree $T = (V,E)$ with $\bar{b}$ children at every node with the following correspondence: every node $v \in V$ represents a cell $S(\bq,\varepsilon)$ for appropriate values of $\bq, \varepsilon$, such that
    \begin{enumerate}[i.]
        \item the root $v_0$ represents $S(\bq_0,\varepsilon_0)$;
        \item if $v$ represents $S(\bq,\varepsilon)$, then its children $v_1, \dots, v_{\bar{b}} \in C(v)$ represent (in some order) the sets $S(\bq^{(1)},\dots,\bq^{(\bar{b})},\varepsilon/s)$ which cover $S(\bq,\varepsilon)$ (as given in \Cref{def:hcs}).
    \end{enumerate}
    Let $\bq(v)$ be the anchor of the cell represented by $v$, and $\varepsilon(v)$ be its radius.
    
    Each target $\tau_i$ corresponding to $X_i$ is a path down the tree starting at the root $v_0$ and always choosing a node representing $S(\bq,\varepsilon)$ containing $X_i$ (if there are multiple such paths, choose one arbitrarily).

    Then, the following bound holds:
    \begin{align}
        \tsp_{\Pi}(X_1, \dots, X_n) \leq \varepsilon_0 \hcp(\tau_1, \dots, \tau_n; b,s) \leq \varepsilon_0 \hcp^*(n;b,s) \,.
    \end{align}
\end{proposition}

\begin{proof}
The first inequality follows because $\varepsilon(v) = \varepsilon_0 s^{-k(v)}$ (at each level the scale is divided by $s$). Thus, we can equate a solution to the HCP instance (moving on the tree and collecting targets) with an equivalent solution on the $\tsp$: at each step where we are at $v \in V$ in the HCP, we are at $\bq(v) \in \cspace$ in the $\tsp$. If $v' \in C(v)$, then 
\begin{align}
    d_{\pi}(\bq(v),\bq(v')) \leq \varepsilon(v) = \varepsilon_0 s^{-k(v)}
\end{align}
i.e. moving from $\bq(v)$ to $\bq(v')$ takes time at most $ \varepsilon_0 w(e)$ where $e = (v,v') \in E$. Furthermore,
\begin{align}
    v \in \tau_i \implies X_i \in S(\bq(v),\varepsilon(v)) \implies d_{\pi}(\bq(v),X_i) \leq \varepsilon_0 s^{-k(v)}
\end{align}
and hence visiting $X_i$ from $\bq(v)$ and then returning takes at most $2 \varepsilon_0 s^{-k(v)}$ time, or, in other words at most $ \varepsilon_0$ times the equivalent cost in the HCP. Thus every solution to the HCP yields a corresponding $\tsp$ trajectory whose length is at most $\varepsilon_0$ times the cost, thus showing the first inequality.

The second inequality then follows from the definition of $\hcp^*(n; \bar{b},s)$.
\end{proof}

We now define a few notions which will help us discuss the HCP.
\begin{definition}
    For $v \in T$, $n_v \defeq | \{ i : v \in \tau_i \} |$, i.e. the number of targets $\tau_i$ passing through $v \in T$.

    A \emph{plan} $\theta$ for a HCP instance $\tau_1, \dots, \tau_n$ is a list of actions which, when performed starting at the root $v_0$, collects all the targets $\tau_i$ and returns to $v_0$. The cost of the plan is $\costtwo(\theta)$, the set of vertices entered is $V_\theta$, and $\theta$ is \emph{optimal} if there is no plan $\theta'$ such that
    \begin{enumerate}[i.]
        \item $\costtwo(\theta') < \costtwo(\theta)$; or
        \item $\costtwo(\theta') = \costtwo(\theta)$ and $|V_{\theta'}| < |V_{\theta}|$.
    \end{enumerate}
\end{definition}
The extra condition for `optimality' (to enter as few vertices as possible) is not really a feature of the optimization problem represented by the HCP in general but is made for convenience, particularly for \Cref{lem:bhcp-optimal-02}.
We now consider two questions: (i) what is the optimal algorithm for the HCP?; (ii) what is the upper bound $\hcp^*(n; b,s)$? 

\begin{lemma} \label{lem:bhcp-optimal-01}
    In any optimal plan, every edge $e \in E$ is crossed either $0$ or $2$ times.
\end{lemma}

\begin{proof}
    Let $e = (v,v')$, where $v$ is on the same side of $e$ as the root $v_0$ and $v'$ is on the opposite side.
    First, since the player must start and end at $v_0$, every edge is crossed an even number of times. Then, if an edge is crossed more than twice, it must happen an even number $2m$ times. Let these happen at steps $s_1 < s_2 < \dots < s_{2m}$, and let $S_i$ denote the actions taken between $s_i$ and $s_{i+1}$ (and $S_0, S_{2m}$ denote respectively the actions before $s_1$ beginning at $v_0$ and the actions after $s_{2m}$ ending at $v_0$). Note that $S_1, S_3, \dots, S_{2m-1}$ all begin and end at $v'$ while $S_2, S_4, \dots, S_{2m-2}$ all begin and end at $v$, and finally $S_0$ begins at $v_0$ and ends at $v$ and $S_{2m}$ begins at $v$ and ends at $v_0$. Then we can produce a modified solution which performs $S_0$, crosses $e$, does $S_1, S_3, \dots, S_{2m-1}$ without crossing $e$, then crosses back over $e$ and performs $S_2, S_4, \dots, S_{2m}$ and ends at $v_0$. Since costs do not depend on the step at which they are performed, the costs of $S_0, S_1, \dots, S_{2m}$ all remain the same, and all targets are collected, but instead of crossing $e$ a total of $2m$ times it is crossed only twice, thus reducing the cost of the plan.
\end{proof}


\begin{lemma} \label{lem:bhcp-optimal-02}
A vertex $v \neq v_0$ is entered in an optimal plan iff $n_v \geq \frac{s}{s-1}$.
\end{lemma}

\begin{proof}
We first prove that if $n_v < \frac{s}{s-1}$ then an optimal plan never enters $v$. Suppose we have a plan $\theta$ that enters $v$; it must do so from $P(v)$, and let $k = k(P(v))$. We then consider a modification $\theta'$ of this plan which, rather than entering $v$, collects all $\tau_i$ going through $v$ at $P(v)$. The only difference between $\theta$ and $\theta'$ is that $\theta$ travels to $v$ and around its subtree and makes its collections there, while $\theta'$ makes all collections at $P(v)$, so we can compare only the costs of these actions. Then:
\begin{itemize}
    \item $\theta'$ incurs costs of $2 s^{-k} n_v$ to collect all the relevant targets;
    \item letting $k' \geq 1$ be the number of levels deeper than $k$ that $\theta$ goes, the costs incurred are at least
    \begin{align} \label{eq:theta-sum}
        2 s^{-(k+k')} n_v + 2 \sum_{j=0}^{k'-1} s^{-(k+j)}
    \end{align}
    (collection costs, minimized if they are maximally deep, and minimum possible movement costs to reach level $k+k'$ and return).
\end{itemize}
But then
\begin{align} \label{eq:theta-vs-theta'}
    n_v &< \frac{s}{s-1} \implies n_v < 1 + \frac{n_v}{s}
\end{align}
and we can then collapse \eqref{eq:theta-sum} because:
\begin{align}
    2 s^{-(k+k')} n_v + 2 \sum_{j=0}^{k'-1} s^{-(k+j)} &= 2 s^{-k} \Big(s^{-k'} n_v + \sum_{j=0}^{k'-1} s^{-j}  \Big) 
    \\ &> 2 s^{-k} \Big(s^{-(k'-1)} n_v + \sum_{j=0}^{k'-2} s^{-j}  \Big)
    \\ > \dots &> 2 s^{-k} n_v
\end{align}
where each step takes the last entry in the sum and the $n_v$ term and applies \eqref{eq:theta-vs-theta'}. But this means that $\theta'$ is a lower-cost plan than $\theta$, so any plan that enters $v$ can be improved by avoiding $v$ and collecting earlier.

Now suppose $n_v \geq \frac{s}{s-1}$, and suppose we have a plan $\theta$ which does not enter $n_v$, and let $k = k(v)$. Let $v'$ be the closest ancestor to $v$ that $\theta$ enters, and let $k' = k(v) - k(v')$ (how many levels above $v$ plan $\theta$ gets). Then all the targets $\tau_i$ that go through $v$ are most efficiently collected at $v'$ and incur cost
\begin{align}
    2 s^{-k(v')} = 2 s^{-(k-k')}
\end{align}
Alternatively, we can consider plan $\theta'$ which instead of collecting at $v'$ go to $v$ and collect there. This incurs cost
\begin{align}
    2 s^{-k(v)}n_v + 2 \sum_{j=0}^{k'-1} s^{-(k(v')+j} &= 2 s^{-k} n_v + 2 \sum_{j=0}^{k'-1} s^{-(k-k'+j)}
    \\ &= 2 s^{-k} \Big(n_v + \sum_{j=1}^{k'} s^{-j}\Big)
    \\ &\leq 2 s^{-k} s^{k'} = 2 s^{-(k-k')}
\end{align}
where the sum collapses because at each step we apply
\begin{align}
    n_v \geq \frac{s}{s-1} \implies s \, n_v \geq s + n_v
\end{align}
Hence, as before, the modified plan $\theta'$ costs at most as much as $\theta$ (and visits more vertices) so $\theta$ cannot be optimal.
\end{proof}

Thus, we have our optimal algorithm: assuming $n \geq \frac{s}{s-1}$ (if not, the optimal algorithm is to simply collect all targets at the root), we consider the subset of vertices
\begin{align}
    V_{\geq \frac{s}{s-1}} \defeq \Big\{v \in V : n_v \geq \frac{s}{s-1} \Big\}
\end{align}
which must include the root and must be connected since
\begin{align}
    n_v \geq \frac{s}{s-1} \implies n_{P(v)} \geq n_v \geq \frac{s}{s-1} \,.
\end{align}
Then the optimal plan is to do a depth-first-search tour of this subgraph (which is a finite tree) using every edge exactly twice and collecting every target $\tau_i$ at the deepest vertex at which it is present.

\begin{proposition} \label{prop::hcp-upper-bound}
For any $n, \bar{b}, s$ such that $s \geq 2$ and $\gamma = \log_s(\bar{b}) \geq 2$,
\begin{align}
    \hcp^*(n; \bar{b},s) \leq 6 s n^{1-\frac{1}{\gamma}}
\end{align}
\end{proposition}

\begin{proof}
We consider a plan that does the following: it takes a tour to each vertex at some level $k^*$; at each $v \in L_{k^*}$, it collects all targets $\tau_i \in v$. Note that this plan has a fixed cost: it must traverse all edges within the first $k^*$ levels twice, and each target collected costs $2 s^{-k^*}$.

Thus the total cost of movement is
\begin{align}
    2 \sum_{k=0}^{k^*-1} \bar{b}(\bar{b}/s)^k = 2 \bar{b} \frac{(\bar{b}/s)^{k^*}-1}{(\bar{b}/s)-1} = 2 \frac{(\bar{b}/s)^{k^*}-1}{(1/s)-(1/\bar{b})}
\end{align}
because between levels $k$ and $k+1$ there are $\bar{b}^{k+1}$ edges, each with cost $s^{-k}$, which must be traversed twice, and the total cost of collection is $2ns^{-k^*}$ since we need to collect $n$ targets at a cost of $2 s^{-k^*}$ each.

Now we let $k^* = \lceil \log_{\bar{b}} n \rceil - 1$. This is the deepest level at which there must still be at least two targets at the same vertex. In this case, we have 
\begin{align}
    (\bar{b}/s)^{k^*} \leq (\bar{b}/s)^{\log_{\bar{b}} n} = n s^{-\log_{\bar{b}} n} = n^{1-\frac{1}{\gamma}}
\end{align}
as $s^{-\log_{\bar{b}} n} = s^{-\log_{s^\gamma} n} = n^{-\frac{1}{\gamma}}$. Furthermore, since $s \geq 2$ and $\gamma \geq 2$, we know that $\bar{b} \geq 2s$ and so $(1/s) - (1/\bar{b}) \geq 1/(2s)$ and so our movement cost is bounded by
\begin{align}
    2 \frac{(\bar{b}/s)^{k^*}-1}{(1/s)-(1/\bar{b})} \leq 4 s n^{1-\frac{1}{\gamma}} \,.
\end{align}
Our collection cost is additionally
\begin{align}
    2ns^{-k^*} \leq 2sn s^{-\log_{\bar{b}} n} = 2s n^{1-\frac{1}{\gamma}} \,. 
\end{align}
Adding these together gives a total cost of $\leq 6s n^{1-\frac{1}{\gamma}}$.
\end{proof}

\begin{remark}
While the abstraction provided by the HCP is useful for providing an upper bound and an algorithm for the Symmetric DTSP, a number of common-sense improvements can be made for the most common control systems. These include the following:
\begin{itemize}
    \item The abstraction ignores the possibility that cells which are far from each other in the structure (i.e. to go from one to the other requires moving far back up the tree and then back down) might be very close in the space. These `horizontal' connections can make it possible to traverse through a number of cells of the same level $k$ without having to return to larger cells close to the root.
    \item As discussed in \Cref{rmk:bhcp-01}, you can ignore the requirement to start and end at the root (which we made in order to have nicer theorems and proofs).
    \item If you have a good point-to-point motion planner (specifically, configuration-to-configuration), you can first use an HCP algorithm to determine the order in which to visit the targets and the configuration $\bq \in [X_i]^{\cspace}$ in which to visit each target $X_i$ and then use the point-to-point motion planner for each pair of consecutive configurations. If the point-to-point motion planner is computationally efficient enough you can even use the cell structure to determine the configurations for each target point (depending on which cell it ends up in, each target point needs to be in a certain set of configurations to be reachable from the anchor of the cell) and then use a standard TSP approximation algorithm on the complete graph where the edge weights are the point-to-point distances.
\end{itemize}
\end{remark}

\subsection{Hierarchical Collection Problems and the TSP}

We now consider the following related problem: suppose there are $m$ bins, and $n$ balls will be thrown in them iid according to some probability vector $\bp = (p_1, \dots, p_m)$; let $n_j$ be the number of balls thrown into the $j$th bin, and note that its distribution is $\bin(p_j, n)$. Let
\begin{align}
    \zetterm \defeq 1 - \frac{1}{\gamma} \in (0,1) ~~\text{ and }~~ Y_j \defeq n_j^\zetterm \text{ and } Y = \sum_{j=1}^m Y_j
\end{align}
This models an upper bound to the Symmetric DSTSP in the following way: we cover the space with $m$ $\varepsilon_0$-scale hierarchical cell structures $S_1, \dots, S_m$ with scaling factor $s$ and cell branching factor $\bar{b} = s^\gamma$; for each $S_j$, let $S'_j \subseteq S_j$ so that $S'_1, \dots, S'_m$ partition $\wspacef$ (each $x \in \wspacef$ is in exactly one $S'_j$). Then let $p_j := \bbP_{X \sim f}[X \in S'_j]$; thus, each target $X_i \sim f$ falls into one of the $m$ `bins' according to probability vector $\bp$, so we can let $n_j$ be the number of targets in $S'_j \subseteq S_j$. Then, by \Cref{prop::hcp-upper-bound}, the time it takes to visit all the $n_j$ targets in $S'_j$ is at most 
\begin{align}
    6 s \varepsilon_0 n^{1-\frac{1}{\gamma}} = 6 s \varepsilon_0 n_j^\zetterm = 6 s \varepsilon_0 Y_j
\end{align} 
To visit all $n$ targets thus takes at most
\begin{align}
    C + \sum_{j=1}^m 6 s \varepsilon_0 Y_j = C + 6 s \varepsilon_0 Y
\end{align}
where $C$ is a constant (in $n$) which depends on the choices of the $S_j$, representing the time it takes for the system to tour the roots of the hierarchical cell structures. This is an upper bound for the Symmetric DSTSP trajectory length because it represents (an upper bound to) the length of the path generated by the hierarchical cell algorithm, but the algorithm is not necessarily optimal.

We now show a few properties of $Y$:
\begin{lemma} 
$\bbE[Y_j] \leq (p_j n)^\zetterm$ and $\bbE[Y] \leq \sum_{j=1}^m (p_j n)^\zetterm = \big(\sum_{j=1}^m p_j^\zetterm \big) n^\zetterm$
\end{lemma}

\begin{proof}
Note that $\cdot^\zetterm$ is a concave function because $\zetterm \in (0,1)$, and that $\bbE[n_j] = p_j n$. Thus, by Jensen's Inequality,
\begin{align}
    \bbE[Y_j] = \bbE[n_j^\zetterm] \leq (\bbE[n_j])^\zetterm = (p_j n)^\zetterm
\end{align}
and the result from $\bbE[Y]$ follows by summing the $\bbE[Y_j]$.
\end{proof}

Given the logic above (and some analysis on what $m$ and $\bp$ are as $\varepsilon_0 \to 0$), this alone gives an upper bound to the expected value of expected trajectory length in the Symmetric DSTSP. However, we also want concentration bounds.

We will consider $Y$ as a Doob martingale: for any $i = 0, 1, \dots, n$, let $Z_i \in [m]$ be the bin that ball $i$ falls into, and consider placing them one-by-one in index order (i.e. deciding $Z_1$ first, then $Z_2$, etc). We then define our (centered) Doob martingale:
\begin{align}
    Y_j(i) = \bbE[Y_j \, | \, Z_1, \dots, Z_i] - \bbE[Y_j] ~~~\text{and}~~~ Y(i) = \sum_{j=1}^m Y_j(i) = \bbE[Y \, | \, Z_1, \dots, Z_i] - \bbE[Y]
\end{align}
Then note that by definition $Y_j(0) = 0$ and $Y_j(n) = Y_j - \bbE[Y_j]$ (and similarly with $Y$), and that $Y(i)$ is a martingale with respect to $i$, i.e.
\begin{align}
    \bbE[Y(i) \, | \, Y(i-1)] = Y(i-1)
\end{align}
(as is well-known from the construction as a conditional expectation of a process, which is known as a Doob martingale). For convenience, we also let $n_j(i)$ denote the number of balls in bin $j$ after $Z_i$ has been decided, i.e.
\begin{align}
    n_j(i) = \sum_{i'=1}^i \bone\{Z_{i'} = j\} \,.
\end{align}
We then show a few results concerning $Y(i)$:
\begin{lemma} \label{lem::Y-bounded-differences}
    $|Y(i)-Y(i-1)| \leq 1$.
\end{lemma}

\begin{proof}
    Consider first what happens when $Z_i$ is changed to $Z'_i$. Let $Z^{(n)} = (Z_1, \dots, Z_n)$ and let $Z^{(n)}_{-i}(Z'_i) = Z_1,\dots,Z_{i-1}, Z'_i, Z_{i+1}, Z_n $ i.e. $Z^{(n)}$ with $Z_i$ replaced by $Z_i$. Then let $Y(Z^{(n)})$ be the value of $Y$ given $Z_1, \dots, Z_n$. For simplicity let $Z_i = j$ and $Z'_i = j'$.
    \begin{align}
        |Y(Z^{(n)}) - Y(Z^{(n)}_{-i}(Z'_i))| &= |n_j^\zetterm + n_{j'}^\zetterm - (n_j-1)^\zetterm - (n_{j'}+1)^\zetterm|
        \\ &= |(n_j^\zetterm - (n_j-1)^\zetterm) - ((n_{j'}+1)^\zetterm - n_{j'}^\zetterm)| \leq 1
    \end{align}
    Each of the two differences are in $[0,1]$ and hence their difference is in $[-1,1]$. Thus changing the value of any $Z_i$ can only change the final value by $1$ (in either direction). Thus, for any $Z_i, Z'_i$,
    \begin{align}
        |\bbE[Y \, | \, Z_1, \dots, Z_i] - \bbE[Y \, | \, Z_1, \dots, Z_{i-1}, Z'_i]|  &= \bbE_{Z_{i+1},\dots,Z_n}[|Y(Z^{(n)}) - Y(Z^{(n)}_{-i}(Z'_i))|]
        \\ &\leq 1
    \end{align}
    But then we get that
    \begin{align}
        |Y(i) - Y(i-1)| &= |\bbE[Y \, | \, Z_1, \dots, Z_i] - \bbE[Y \, | \, Z_1, \dots, Z_{i-1}]|
        \\ &= |\bbE[Y \, | \, Z_1, \dots, Z_i] - \bbE_{Z'_i} \big[\bbE[Y \, | \, Z_1, \dots, Z_{i-1}, Z'_i]\big] |
        \\ &\leq \bbE_{Z'_i}\big[|\bbE[Y \, | \, Z_1, \dots, Z_i] - \bbE[Y \, | \, Z_1, \dots, Z_{i-1}, Z'_i]| \big]
        \\ &\leq 1
    \end{align}
\end{proof}

Thus, we can immediately apply Azuma-Hoeffding to get a bound; unfortunately, to get a good result this requires that $\gamma > 2$ (i.e. $\zetterm = 1 - \frac{1}{\gamma} > \frac{1}{2}$). However, we will give the result and sharpen it afterwards:

\begin{lemma} \label{lem::Y-azuma-hoeffding}
    Given the definitions above,
    \begin{align}
        \bbE \bigg[Y(n) \geq n^\zetterm \sum_{j=1}^m p_j^\zetterm \bigg] \leq e^{-\frac{n^{2\zetterm} (\sum_{j=1}^m p_j^\zetterm)^2 }{2n}} = e^{-\frac{1}{2} n^{2\zetterm - 1} (\sum_{j=1}^m p_j^\zetterm)^2}
    \end{align}
    Furthermore, $\sum_{j=1}^m p_j^\zetterm \geq 1$, so we can even find a bound independent of $\bp = (p_1,\dots,p_m)$:
    \begin{align}
        \bbE \bigg[Y(n) \geq n^\zetterm \sum_{j=1}^m p_j^\zetterm \bigg] \leq e^{-\frac{1}{2}n^{2\zetterm - 1}}
    \end{align}
\end{lemma}

\begin{proof}
    This is a direct application of the Azuma-Hoeffding Inequality with the bounded differences as shown in \Cref{lem::Y-bounded-differences}.

    We know that $\sum_{j=1}^m p_j^\zetterm \geq 1$ because $\bp$ is a probability vector and can therefore be expressed as a convex mixture of the basis vectors $\bone_1, \dots, \bone_m$. Then we note that the function $\theta(\bp) = \sum_{j=1}^m p_j^\zetterm$ is concave and $\theta(\bone_j) = 1$ for all $j$, and hence by Jensen's Inequality
    \begin{align}
        \theta(\bp) \geq \sum_{j=1}^m p_j \theta(\bone_j) = 1
    \end{align}
    and thus we have the result we wanted.
\end{proof}

\begin{proposition} \label{prop::conc-bounds-azhoef}
    Given $Y = \sum_{j=1}^m n_j^\zetterm$, we have
    \begin{align}
        \bbP \bigg[Y \geq 2 n^\zetterm \sum_{j=1}^m p_j^\zetterm \bigg] \leq e^{-\frac{1}{2} n^{2\zetterm - 1} (\sum_{j=1}^m p_j^\zetterm)^2}
    \end{align}
    which implies the following bound independent of $\bp$:
    \begin{align}
        \bbP \bigg[Y \geq 2 n^\zetterm \sum_{j=1}^m p_j^\zetterm \bigg] \leq e^{-\frac{1}{2} n^{2\zetterm - 1}}
    \end{align}
\end{proposition}

\begin{proof}
    Using the definition of the centered martingale $Y(i)$, we have $Y(n) = Y - \bbE[Y]$, thus giving us:
    \begin{align}
        Y \geq 2 n^\zetterm \sum_{j=1}^m p_j^\zetterm &\implies Y - n^\zetterm \sum_{j=1}^m p_j^\zetterm \geq n^\zetterm \sum_{j=1}^m p_j^\zetterm
        \\ &\implies Y - \bbE[Y] \geq n^\zetterm \sum_{j=1}^m p_j^\zetterm
        \\ &\implies Y(n) \geq n^\zetterm \sum_{j=1}^m p_j^\zetterm
    \end{align}
    which in turn means by \Cref{lem::Y-azuma-hoeffding}
    \begin{align}
        \bbP \bigg[ Y \geq 2 n^\zetterm \sum_{j=1}^m p_j^\zetterm \bigg] \leq \bbP \bigg[ Y(n) \geq n^\zetterm \sum_{j=1}^m p_j^\zetterm \bigg] \leq e^{-\frac{1}{2} n^{2\zetterm - 1} (\sum_{j=1}^m p_j^\zetterm)^2}
    \end{align}
    The second bound independent of $\bp$ follows as in the proof of \Cref{lem::Y-azuma-hoeffding}.
\end{proof}

Since we have $-\frac{1}{2} n^{2\zetterm-1}$ in the exponent, this is a very high probability concentration bound when $\zetterm > 1/2$, which happens iff $\gamma > 2$. When $\gamma \leq 2$, this bound is not high probability (let alone very high probability), so we need more precise methods.

One place where we lost some potential benefit is in \Cref{lem::Y-bounded-differences}: while this is the best single constant bound that holds under all circumstances, it ignores the fact that when $i$ is not very close to $n$, there are a lot of $Z_{i'}$ for $i < i'$ which still need to be allocated; this in turn means that each bin has some expected number of balls still to be added, so adding a ball to any given bin \emph{now} won't change $\bbE[n_j^\zetterm \, | \, Z_1, \dots, Z_i]$ by very much (as the differences between $k^\zetterm$ and $(k+1)^\zetterm$ get smaller and smaller as $k$ gets larger). First, we need the following lemma:

\begin{lemma} \label{lem::diff-of-W-zeta}
    Let $W \sim \bin(p,n)$ for some $p \in (0,1)$ and $n$, and $\zetterm \in (0,1)$; then
    \begin{align}
        \bbE[(W+1)^\zetterm] - \bbE[W^\zetterm] \leq e^{-\frac{3}{28} np} + 2\zetterm (np)^{\zetterm-1}
    \end{align}
\end{lemma}

\begin{proof}
    Since $\zetterm \in (0,1)$, $\cdot^\zetterm$ is concave, and so the first-order approximation is an upper bound, yielding
    \begin{align}
        (W+1)^\zetterm - W^\zetterm &\leq W^\zetterm + \zetterm W^{\zetterm-1} - W^\zetterm = \zetterm W^{\zetterm-1}\label{eq::diff-of-powers}
    \end{align}
    While this is fine for $W > 0$, we have to be careful for the case of $W = 0$, since the exponent $\zetterm-1$ is negative; however, when $W = 0$, we have $(W+1)^\zetterm - W^\zetterm = 1$, so we have the bound
    \begin{align}
        (W+1)^\zetterm - W^\zetterm \leq \min(\zetterm W^{\zetterm-1},1)
    \end{align}
    since for $W \geq 1$, we have $\zetterm W^{\zetterm-1} \leq \zetterm < 1$.

    We then note the following: for $W \leq np/2$, we have $(W+1)^\zetterm - W^\zetterm \leq 1$, and for $W > np/2$, we have $(W+1)^\zetterm - W^\zetterm \leq \zetterm (np/2)^{\zetterm-1}$. Thus, we have
    \begin{align}
        \bbE[(W+1)^\zetterm] - \bbE[W^\zetterm] &= \bbE[(W+1)^\zetterm - W^\zetterm]
        \\ &\leq \bbP[W \leq np/2] + \bbP[W > np/2] (\zetterm (np/2)^{\zetterm-1})
        \\ &\leq \bbP[W \leq np/2] + 2 \zetterm (np)^{\zetterm-1}
    \end{align}
    We can then use Bernstein's Inequality to bound $\bbP[W \leq np/2]$, as $W$ is a binomial random variable and hence a sum of iid Bernoulli random variables. In particular, we let these iid Bernoullis be $W_i$ ($i \in [n]$) and we let
    \begin{align}
        W'_i = p - W_i ~~~\text{(which also means } W_i = p - W'_i \text{)}
    \end{align}
    which are zero-mean and $|W'_i| \leq 1$ always. Thus,
    \begin{align}
        W = \sum_{i=1}^n W_i = np - \sum_{i=1}^n W'_i
    \end{align}
    i.e. $W \leq np/2 \iff \sum_{i=1}^n W'_i \geq np/2$. Thus, since the $W'_i$ have variance $p(1-p)$, we have
    \begin{align}
        \bbP\bigg[\sum_{i=1}^n W'_i \geq \frac{np}{2} \bigg] \leq e^{-\frac{(np)^2/8}{np(1-p) + np/6}} = e^{-\frac{np/8}{(1-p) + 1/6}} \leq e^{-\frac{3}{28} np}
    \end{align}
    and hence we can put it all together and get
    \begin{align}
        \bbE[(W+1)^\zetterm] - \bbE[W^\zetterm] \leq e^{-\frac{3}{28} np} + 2 \zetterm (np)^{\zetterm-1}
    \end{align}
\end{proof}

\begin{lemma} \label{lem::Y-bounded-differences-redux}
    For the martingale defined above, WLOG $\bp$ be sorted, i.e. $0 < p_1 \leq \dots \leq p_m$ (any $0$ entries can be ignored and $\bp$ reduced). Then
    \begin{align}
        |Y(i) - Y(i-1)| \leq \min(e^{-\frac{3}{28} (n-i) p_1} + 2\zetterm ((n-i)p_1)^{\zetterm-1}, 1)
    \end{align}
\end{lemma}

\begin{proof}
    We consider the situation when $Z_i$ is about to be decided, i.e. $Z_1, \dots, Z_{i-1}$ are known and $Z_{i+1}, \dots, Z_n$ are in the future. How much can $\bbE[Y \, | \, Z_1, \dots, Z_{i-1}]$ differ from $\bbE[Y \, | \, Z_1, \dots, Z_i]$? We note that
\begin{align}
    \big | \bbE[Y \, | \, Z_1, \dots, Z_i] - \bbE[Y \, | \, Z_1, \dots, Z_{i-1}] \big | 
    \\ = \big | \bbE[Y \, | \, Z_1, \dots, Z_i] - \bbE_{Z \sim \bp} [\bbE[Y \, | \, Z_1, \dots, Z_{i-1}, Z]] \big |
    \\ \leq \max_{Z \in [m]} \bbE[Y \, | \, Z_1, \dots, Z_{i-1}, Z] - \min_{Z' \in [m]} \bbE[Y \, | \, Z_1, \dots, Z_{i-1}, Z'] 
\end{align}
(note that we are working with $Y$ which is not centered, but the centering subtracts $\bbE[Y]$ from both sides and can be ignored). That is, we can bound $|Y(i) - Y(i-1)|$ given $Z_1, \dots, Z_{i-1}$ by taking the difference between the choice of $Z_i$ that maximizes $Y(i)$ and the choice of $Z_i$ that minimizes $Y(i)$. Since $Z_1, \dots, Z_{i-1}$ are unknown, we also have to maximize over them, i.e. we want to bound
\begin{align}
    &\max_{Z_1,\dots,Z_{i-1}} (\max_{Z \in [m]} \bbE[Y \, | \, Z_1, \dots, Z_{i-1}, Z] - \min_{Z' \in [m]} \bbE[Y \, | \, Z_1, \dots, Z_{i-1}, Z'])
\end{align}
We now break this down further: setting $Z = j$ really adds $1$ to the final $n_j$, and $Z' = j'$ does the same to $n_{j'}$; the remaining $j'' \neq j,j'$ are unaffected and cancel out. Note also that $Z_1, \dots, Z_{i-1}$ only matter insofar as they affect $n_1(i-1), \dots, n_m(i-1)$, so we will condition on those instead. Thus, letting $W_j \sim \bin(p_j, n-i)$, we have :
\begin{align}
    \max_{Z_1,\dots,Z_{i-1}} (\max_{Z \in [m]} \bbE[Y \, | \, Z_1, \dots, Z_{i-1}, Z] - \min_{Z' \in [m]} \bbE[Y \, | \, Z_1, \dots, Z_{i-1}, Z'])
    \\=  \max_{n_j(i), n_{j'}(i)} \big(\bbE[(W_j + n_j(i) + 1)^\zetterm ] - \bbE[(W_j + n_j(n))^\zetterm] 
    \\- \big(\bbE[(W_{j'} + n_{j'}(i) + 1)^\zetterm ] - \bbE[(W_{j'} + n_{j'}(n))^\zetterm] \big)
    \\ \leq \max_{n_j(i)} \bbE[(W_j + n_j(i) + 1)^\zetterm ] - \bbE[(W_j + n_j(n))^\zetterm]
    \\ = \max_{j} \bbE[(W_j + 1)^\zetterm ] - \bbE[W_j^\zetterm]
\end{align}
where the expectation is over the $n-i$ values $Z_{i+1}, \dots, Z_n$ yet to be determined. This holds because $(a_1+1)^\zetterm-a_1^\zetterm \geq (a_2+1)^\zetterm-a_2^\zetterm$ if $a_1 \leq a_2$; and that also means that this is maximized if $j$ is selected so $p_j$ is minimized, i.e. we use $p_1$.

Then, by \Cref{lem::diff-of-W-zeta} (and the previously-derived upper bound of $1$), the result follows, with $n-i$ because this is the number of $Z_{i'}$ yet to be determined.
\end{proof}

We now adjust the Azuma-Hoeffding argument for the new difference bound:

\begin{proposition} \label{prop::Y-azuma-hoeffding-redux}
    Given the definitions above, let $\zetterm \leq 2/3$ and (wlog) let $p_1$ be the smallest nonzero value in $\bp = (p_1, \dots, p_m)$. Then, for all
    \begin{align}
        n \geq \frac{280}{3} \log(1/\zetterm) p_1^{-1}
    \end{align}
    we have the following concentration bounds:
    \begin{align}
        \bbE \bigg[Y(n) \geq 2 n^\zetterm \sum_{j=1}^m p_j^\zetterm \bigg] \leq 
        \begin{cases} e^{-\frac{p_1 n^{2\zetterm} (\sum_{j=1}^m p_j^\zetterm)^2}{13 + \frac{8}{2\zetterm-1} n^{2\zetterm-1} }}
            & \text{for } \zetterm \in (1/2,2/3)
            \\ e^{- \frac{(2/9) p_1 n (\sum_{j=1}^m p_j^{1/2})^2 }{127 - \log(1/p_1) + \log(n) }} &\text{for } \zetterm = 1/2
            \\ e^{-\frac{p_1 n^{2\zetterm} (\sum_{j=1}^m p_j^\zetterm)^2}{\frac{560}{3} \log(1/\zetterm) + 4 + 18 \zetterm^2 \frac{1}{1-2\zetterm} (\frac{280}{3} \log(1/\zetterm) )^{2\zetterm-1}  }} &\text{for } \zetterm \in (0,1/2)
        \end{cases}
    \end{align}
    As before, $\sum_{j=1}^m p_j^\zetterm \geq 1$, so we can remove it and the bound remains valid (if potentially somewhat looser). Writing in asymptotic notation (but noting that we have firm constants by the above) this is
    \begin{align}
        \bbE \bigg[Y(n) \geq n^\zetterm \sum_{j=1}^m p_j^\zetterm \bigg] \leq 
        \begin{cases} e^{- \Theta(p_1 n (2\zetterm-1) (\sum_{j=1}^m p_j^\zetterm)^2)}
            & \text{for } \zetterm \in (1/2,2/3)
            \\ e^{- \Theta(p_1 (n/\log(n)) (\sum_{j=1}^m p_j^\zetterm)^2)} &\text{for } \zetterm = 1/2
            \\ e^{- \Theta(p_1 n^{2\zetterm} (\sum_{j=1}^m p_j^\zetterm)^2 (\log(1/\zetterm) + (1-2\zetterm)^{-1})^{-1})} &\text{for } \zetterm \in (0,1/2)
        \end{cases}
    \end{align}
\end{proposition}

\begin{proof}
    Let 
    \begin{align}
        c_i = \min(e^{-\frac{3}{28} (n-i) p_1} + 2\zetterm ((n-i)p_1)^{\zetterm-1}, 1)
    \end{align} 
    We now want to compute $\sum_{i=1}^n c_i^2$. We first define 
    \begin{align}
        a_i = c_{n-i} = \min(e^{-\frac{3}{28} i p_1} + 2\zetterm (i \, p_1)^{\zetterm-1}, 1)
    \end{align}
    where $i$ now goes from $0$ to $n-1$. Then we know that
    \begin{align}
        \sum_{i=0}^{n-1} a_i^2 = \sum_{i=1}^n c_i^2
    \end{align}
    Let $z = i p_1$; then we want to compare $e^{-\frac{3}{28} z}$ to $2\zetterm z^{\zetterm-1}$. In particular we want to find $z^*$ sufficiently large such that
    \begin{align}
        e^{-\frac{3}{28} z} \leq 2\zetterm z^{\zetterm-1} \text{ for all } z \geq z^*
    \end{align}
    We compute:
    \begin{align}
        e^{-\frac{3}{28} z} &\leq 2\zetterm z^{\zetterm-1}
        \\ \iff -\frac{3}{28} z &\leq (\zetterm-1)\log(z) + \log(2\zetterm)
        \\ \iff \frac{3}{28} z &\geq (1-\zetterm)\log(z) - \log(2\zetterm)
    \end{align}
    Then note that if $z \geq \frac{28}{3} \frac{1}{1-\zetterm}$,
    \begin{align}
        \frac{d}{dz} \frac{3}{28} z = \frac{3}{28} \geq (1-\zetterm)\frac{1}{z} = \frac{d}{dz} (1-\zetterm)\log(z) - \log(2\zetterm)
    \end{align}
    Thus, if $z^* \geq \frac{28}{3} \frac{1}{1-\zetterm}$ and
    \begin{align}
        \frac{3}{28} z^* &\geq (1-\zetterm)\log(z^*) - \log(2\zetterm)
    \end{align}
    then we can conclude that for all $z \geq z^*$,
    \begin{align}
        e^{-\frac{3}{28} z} &\leq 2\zetterm z^{\zetterm-1} \,.
    \end{align}
    Noting that $1 < \frac{28}{3} \frac{1}{1-\zetterm} \leq 28$ (as $\zetterm \leq 2/3$), we let $z^* = \frac{280}{3} \log(1/\zetterm)$. Then
    \begin{align}
        \frac{3}{28} z^* = 10 \log(1/\zetterm) \geq (1-\zetterm)\log \Big(\frac{280}{3} \log(1/\zetterm) \Big) - \log(2\zetterm)
    \end{align}
    which holds for $\zetterm \in (0,2/3)$, and $z^* \geq 28$ for this range as well. This means that for all $z \geq z^*$ (and $\zetterm \in (0,2/3)$) we have
    \begin{align} 
        e^{-\frac{3}{28} z} &\leq 2\zetterm z^{\zetterm-1} \,.
    \end{align}
    Then, recalling that $z = i p_1$, we have for all $i \geq z^*/p_1 = \frac{280}{3} \log(1/\zetterm) p_1^{-1}$,
    \begin{align}
        e^{-\frac{3}{28} i p_1} &\leq 2\zetterm (i \, p_1)^{\zetterm-1} \implies e^{-\frac{3}{28} i p_1} + 2\zetterm (i \, p_1)^{\zetterm-1} \leq 3\zetterm (i \, p_1)^{\zetterm-1}
    \end{align}
    Thus, we have the following:
    \begin{align}
        a_i = \min(e^{-\frac{3}{28} i p_1} + 2\zetterm (i \, p_1)^{\zetterm-1}, 1) \leq \begin{cases} 1 & \text{if } i \leq \frac{280}{3} \log(1/\zetterm) p_1^{-1} \\  3\zetterm (i \, p_1)^{\zetterm-1} & \text{if } i \geq \frac{280}{3} \log(1/\zetterm) p_1^{-1}  \end{cases}
    \end{align}
    We now assume that $n \geq \frac{280}{3} \log(1/\zetterm) p_1^{-1}$. In that case, 
    \begin{align}
        \sum_{i=0}^{n-1} a_i^2 &\leq \frac{280}{3} \log(1/\zetterm) p_1^{-1} + \sum_{i=\frac{280}{3} \log(1/\zetterm) p_1^{-1}}^{n-1} 9\zetterm^2 (i \, p_1)^{2\zetterm-2}
        \\ &\leq \frac{280}{3} \log(1/\zetterm) p_1^{-1} + 2 + 9\zetterm^2 p_1^{2\zetterm-2} \int_{\frac{280}{3} \log(1/\zetterm) p_1^{-1}}^{n-1}  t^{2\zetterm-2} \, dt
    \end{align}
    We now have to split this according to $\zetterm > 1/2$, $\zetterm = 1/2$, and $\zetterm < 1/2$. This yields the following: for $\zetterm \in (1/2,2/3)$,
    \begin{align}
        9\zetterm^2 p_1^{2\zetterm-2} \int_{\frac{280}{3} \log(1/\zetterm) p_1^{-1}}^{n-1}  t^{2\zetterm-2} \, dt 
        \\ = 9 \zetterm^2 p_1^{2\zetterm-2} \frac{1}{2\zetterm-1} \bigg((n-1)^{2\zetterm-1} -  \Big(\frac{280}{3} \log(1/\zetterm) p_1^{-1}\Big)^{2\zetterm-1}\bigg)
        \\ \leq 4 p_1^{-1} \frac{1}{2\zetterm-1} \bigg(n^{2\zetterm-1} -  \Big(\frac{280}{3} \log(1/\zetterm)\Big)^{2\zetterm-1}\bigg)
    \end{align}
    and hence we can conclude that (again, for $\zetterm \in (1/2,2/3)$)
    \begin{align}
        \sum_{i=0}^{n-1} a_i^2 &\leq p_1^{-1} \bigg(\frac{280}{3} \log(1/\zetterm) -  \frac{4}{2\zetterm-1}\Big(\frac{280}{3} \log(1/\zetterm)\Big)^{2\zetterm-1}\bigg) + 2 + \frac{4}{2\zetterm-1}p_1^{-1} n^{2\zetterm-1}
        \\ &= O(p_1^{-1} \log(1/\zetterm) + 1) + O(p_1^{-1} (2\zetterm-1)^{-1}) n^{2\zetterm-1}
    \end{align}
    Then, for $\zetterm = 1/2$, we have
    \begin{align}
        \sum_{i=0}^{n-1} a_i^2 &\leq \frac{280}{3} \log(1/\zetterm) p_1^{-1} + 2 + 9\zetterm^2 p_1^{2\zetterm-2} \int_{\frac{280}{3} \log(1/\zetterm) p_1^{-1}}^{n-1}  t^{2\zetterm-2} \, dt
        \\ &= \frac{280}{3} \log(2) p_1^{-1} + 2 + (9/4) p_1^{-1} \bigg( \log(n-1) - \log\Big(\frac{280}{3} \log(2)\Big) + \log(1/p_1) \bigg)
        \\ &\leq \bigg(\frac{280}{3} \log(2) - (9/4) \log\Big(\frac{280}{3} \log(2)\Big) \bigg) p_1^{-1} + 2 + (9/4) p_1^{-1} \big( \log(n) - \log(1/p_1) \big)
        \\ &< (56 - (9/4) \log(1/p_1))p_1^{-1} + 2 + (9/4)p_1^{-1} \log(n)
    \end{align}
    And finally, for $\zetterm \in (0,1/2)$, we have
    \begin{align}
        \sum_{i=0}^{n-1} a_i^2 &\leq \frac{280}{3} \log(1/\zetterm) p_1^{-1} + 2 + 9\zetterm^2 p_1^{2\zetterm-2} \int_{\frac{280}{3} \log(1/\zetterm) p_1^{-1}}^{n-1}  t^{2\zetterm-2} \, dt
        \\ &\leq \frac{280}{3} \log(1/\zetterm) p_1^{-1} + 2 + 9\zetterm^2 p_1^{2\zetterm-2} \int_{\frac{280}{3} \log(1/\zetterm) p_1^{-1}}^\infty  t^{2\zetterm-2} \, dt
        \\ &= \frac{280}{3} \log(1/\zetterm) p_1^{-1} + 2 + 9\zetterm^2 p_1^{2\zetterm-2} \frac{1}{1-2\zetterm} \Big(\frac{280}{3} \log(1/\zetterm) p_1^{-1} \Big)^{2\zetterm-1}
        \\ &= \bigg(\frac{280}{3} \log(1/\zetterm) + 2 + 9\zetterm^2 \frac{1}{1-2\zetterm} \Big(\frac{280}{3} \log(1/\zetterm) \Big)^{2\zetterm-1}\bigg) p_1^{-1}
    \end{align}
    (note that this is constant in $n$).

    We then plug it into Azuma-Hoeffding (keeping in mind that $\sum_{i=1}^n c_i^2 = \sum_{i=0}^{n-1} a_i^2$. We get: for $\zetterm \in (1/2,2/3)$,
    \begin{align}
        \bbE \Bigg[Y(n) \geq 2 n^\zetterm \sum_{j=1}^m p_j^\zetterm \Bigg] &\leq e^{-\frac{n^{2\zetterm} (\sum_{j=1}^m p_j^\zetterm)^2}{2 \sum_{i=1}^n a_i^2 }} 
        \\ &\leq e^{-\frac{n^{2\zetterm} (\sum_{j=1}^m p_j^\zetterm)^2}{2 p_1^{-1} (\frac{280}{3} \log(1/\zetterm) -  \frac{4}{2\zetterm-1}(\frac{280}{3} \log(1/\zetterm))^{2\zetterm-1}) + 4 + \frac{8}{2\zetterm-1}p_1^{-1} n^{2\zetterm-1} }}
        \\ &= e^{-\frac{p_1 n^{2\zetterm} (\sum_{j=1}^m p_j^\zetterm)^2}{2 (\frac{280}{3} \log(1/\zetterm) -  \frac{4}{2\zetterm-1}(\frac{280}{3} \log(1/\zetterm))^{2\zetterm-1}) + 4p_1 + \frac{8}{2\zetterm-1} n^{2\zetterm-1} }}
        \\ &\leq e^{-\frac{p_1 n^{2\zetterm} (\sum_{j=1}^m p_j^\zetterm)^2}{9 + 4 p_1 + \frac{8}{2\zetterm-1} n^{2\zetterm-1} }}
        \\ &\leq e^{-\frac{p_1 n^{2\zetterm} (\sum_{j=1}^m p_j^\zetterm)^2}{13 + \frac{8}{2\zetterm-1} n^{2\zetterm-1} }}
        \\ &= e^{- \Theta(p_1 n (2\zetterm-1) (\sum_{j=1}^m p_j^\zetterm)^2)}
    \end{align}
    where the second-to-last step is simply from numerically taking the maximum of
    \begin{align}
        \frac{280}{3} \log(1/\zetterm) -  \frac{4}{2\zetterm-1}\Big(\frac{280}{3} \log(1/\zetterm)\Big)^{2\zetterm-1}
    \end{align}
    over the domain $\zetterm \in (1/2,2/3)$.
    
    For $\zetterm = 1/2$, we have
    \begin{align}
        \bbE \Bigg[Y(n) \geq 2 n^{1/2} \sum_{j=1}^m p_j^{1/2} \Bigg] &\leq e^{-\frac{n (\sum_{j=1}^m p_j^{1/2})^2}{2 \sum_{i=1}^n a_i^2 }} 
        \\ &\leq e^{-\frac{n (\sum_{j=1}^m p_j^{1/2})^2}{2((56 - (9/4) \log(1/p_1))p_1^{-1} + 2 + (9/4)p_1^{-1} \log(n)) }}
        \\ &= e^{- \frac{p_1 n (\sum_{j=1}^m p_j^{1/2})^2 }{(28 - (9/2) \log(1/p_1)) + 4p_1 + (9/2) \log(n) }}
        \\ &\leq  e^{- \frac{(2/9) p_1 n (\sum_{j=1}^m p_j^{1/2})^2 }{(126 - \log(1/p_1)) + p_1 + \log(n) }}
        \\ &\leq  e^{- \frac{(2/9) p_1 n (\sum_{j=1}^m p_j^{1/2})^2 }{127 - \log(1/p_1) + \log(n) }}
        \\ &=  e^{- \Theta(p_1 (n/\log(n)) (\sum_{j=1}^m p_j^\zetterm)^2)}
    \end{align}
    Finally, for $\zetterm \in (0,1/2)$ we have
    \begin{align}
        \bbE \Bigg[Y(n) \geq 2 n^\zetterm \sum_{j=1}^m p_j^\zetterm \Bigg] &\leq e^{-\frac{n^{2\zetterm} (\sum_{j=1}^m p_j^\zetterm)^2}{2 \sum_{i=1}^n a_i^2 }}
        \\ &\leq e^{-\frac{n^{2\zetterm} (\sum_{j=1}^m p_j^\zetterm)^2}{(\frac{560}{3} \log(1/\zetterm) + 4 + 18 \zetterm^2 \frac{1}{1-2\zetterm} (\frac{280}{3} \log(1/\zetterm) )^{2\zetterm-1}) p_1^{-1} }}
        \\ &= e^{-\frac{p_1 n^{2\zetterm} (\sum_{j=1}^m p_j^\zetterm)^2}{\frac{560}{3} \log(1/\zetterm) + 4 + 18 \zetterm^2 \frac{1}{1-2\zetterm} (\frac{280}{3} \log(1/\zetterm) )^{2\zetterm-1}  }}
        \\ &= e^{- \Theta(p_1 n^{2\zetterm} (\sum_{j=1}^m p_j^\zetterm)^2 (\log(1/\zetterm) + (1-2\zetterm)^{-1})^{-1})}
    \end{align}
\end{proof}

Note that the dependence on $n$ in the above bounds (holding $\zetterm, p_1$ fixed) is the following: $e^{-\Theta(n)}$ for $\zetterm \in (1/2, 2/3)$; $e^{-\Theta(n/\log(n))}$ for $\zetterm = 1/2$; and $e^{-\Theta(n^{2\zetterm})}$ for $\zetterm \in (0,1/2)$. 

Barring fractals (which are of interest for future work) we have integer $\gamma \geq 2$ and hence $1 - \frac{1}{\gamma} = \zetterm \in (0,2/3) \implies \zetterm = 1/2$, i.e. when considering control-affine systems and full-dimensional continuous distributions $f$ of target points, our main concern is $\zetterm = 1/2$. For $\zetterm \geq 2/3$, \Cref{prop::conc-bounds-azhoef} provides a much cleaner very high probability concentration bound (whose dependence on $n$ is $e^{-\Theta(n^{2\zetterm-1})} \geq e^{-\Theta(n^{1/6})}$), without any fuss about how large $n$ has to be or the $p_1$ term.

\subsection{HCS covering algorithm}

We now consider a division of $\wspacef$ into $m$ different $\varepsilon_0$-scale hierarchical cell structures, as given in \Cref{assm:hcs}; we call their roots $\bq_1, \dots, \bq_m$. Let $x_j = [\bq_j]_\wspace$. These will remain fixed while the number of targets $n \to \infty$. 
Let $S_1, S_2, \dots, S_m$ be the top-level cells of the HCS's that cover $\wspacef$ (at scale $\varepsilon_0$), and let $x_j = \proj{\bq_j}$ for all $j$; we define
\begin{align}
    p_j \defeq \bbP_{X \sim f}[X \in S(\bq_j,\varepsilon_0)]
\end{align}
Note that $p_j$ depend on which HCS's we use, which in turn depends on $\varepsilon_0$. For this proof, we will assume that the sets $S(\bq_j,\varepsilon_0)$ are disjoint (see Appendix~\ref{appx:good-target-bad-target}). Thus, $\sum_{j=1}^m p_j = 1$, and distributing $X \sim f$ puts it into a cell $S_j$ according to probability vector $\bp = (p_1,\dots,p_m)$. Let $j(X)$ denote the index of the cell $X$ falls into.

Then we wish to show that each HCS covers a certain amount of probability mass, except a negligible fraction:
\begin{lemma} \label{lem:prob-of-cells}
    For any $\rho_1 > 0$, let a \emph{$\rho_1$-good} cell $S_j$ be defined as one such that
    \begin{align}
        p_j \geq (1-\rho_1) \alpha \check{f}^{(\zeta)}(x_j) \check{g}^{(\zeta)}(x_j)  \varepsilon_0^\gamma \,.
    \end{align}
    Then, for any $\rho_1, \rho_2 \geq 0$, for any sufficiently small $\zeta < 0$ there is some $\varepsilon_0^*$ such that for any $0 < \varepsilon_0 \leq \varepsilon_0^*$ there are $\bq_1, \dots, \bq_m$ satisfying the conditions of \Cref{assm:hcs} such that
    \begin{align}
        \bbP_{X \sim f}[S_{j(X)} \text{ is } \rho_1\text{-good}] \geq 1-\rho_2\,.
    \end{align}
\end{lemma}

\begin{proof}
    Consider some threshold $\rho^* > 0$ and call a cell $S_j$ \emph{$\rho^*$-sufficient} if $\check{f}^{(\zeta)}(x_j) \geq \rho^*$. We claim the following:
    \begin{enumerate}[i.]
        \item For any $\rho_2 > 0$, there is some $\rho^* > 0$ such that for all sufficiently small $\zeta > 0$, $\bbP_{X \sim f}[\check{f}^{(\zeta)}(X) \leq (3/2)\rho^*] \leq \rho_2$.
        \item For any $\rho^* > 0$, there is some $\varepsilon_0^*$ such that if $\check{f}^{(\zeta)}(X) > (3/2)\rho^*$, then $S_{j(X)}$ must be $\rho^*$-sufficient.
        \item For any $\rho^*, \rho_1, \zeta > 0$, there is a sufficiently small $\varepsilon_0^*$ such that for all $0 < \varepsilon_0 \leq \varepsilon_0^*$, any $\rho^*$-sufficient cell is $\rho_1$-good.
    \end{enumerate}
    Claim (i) holds because $f$ is continuous almost everywhere, and therefore $\lim_{\zeta \to 0} \check{f}^{(\zeta)}(x) = f(x)$ almost everywhere; thus we pick $\rho^*$ sufficiently small so that $\bbP_{X \sim f}[f(X) \leq (3/2)\rho^*] \leq p_2/2$, and then letting $\zeta \to 0$ we get
    \begin{align}
        \lim_{\zeta \to 0} \bbP_{X \sim f}[\check{f}^{(\zeta)}(X) \leq (3/2)\rho^*] = \bbP_{X \sim f}[f(X) \leq (3/2)\rho^*] \leq p_2/2
    \end{align}
    which means that for sufficiently small $\zeta > 0$, we have $\bbP_{X \sim f}[\check{f}^{(\zeta)}(X) \leq (3/2)\rho^*] \leq \rho_2$.

    Claim (ii) holds because by definition
    \begin{align}
        X \in S_{j(X)} \subseteq \bar{R}_{\varepsilon_0}(\bq_j) \subseteq \cB_{c_\Pi \varepsilon_0}(x_j)
    \end{align}
    where $\cB$ denotes a ball in the metric on $\wspace$ (with a given radius and center). Since by definition $\check{f}^{(\zeta)}$ is $(1/\zeta)$-Lipschitz continuous, as long as $c_\Pi \varepsilon_0 / \zeta \leq \rho^*/2$ (i.e. $\varepsilon_0 \leq \rho^* \zeta / (2c_\Pi)$) we have
    \begin{align}
        |\check{f}^{(\zeta)}(X) - \check{f}^{(\zeta)}(x_j)| &\leq \rho^*/2
        \\ \implies \check{f}^{(\zeta)}(x_j) &\geq \rho^*
    \end{align}
    for any $X$ such that $\check{f}^{(\zeta)}(X) > (3/2)\rho^*$, which by definition means that $S_{j(X)}$ is $\rho^*$-sufficient.

    Claim (iii) holds because $\check{f}^{(\zeta)}$ is Lipschitz continuous, $S_j \subseteq \cB_{c_\Pi \varepsilon_0}(x_j)$, and $\check{f}^{(\zeta)}(x_j) \geq \rho^*$; thus, by making $\varepsilon_0$ sufficiently small, $\check{f}^{(\zeta)}(X) \geq (1-\rho_1/2) \check{f}^{(\zeta)}(x_j)$ for every $X \in \cB_{c_\Pi \varepsilon_0}(x_j)$, and $\vol_{\wspace}(S_j) \geq (1 - \rho_1/2) \alpha \check{g}^{(\zeta)}(x_j) \varepsilon_0^\gamma$. Thus,
    \begin{align}
        \bbP_{f \sim X}[X \in S_j] &= \int_{S_j} f(x) \, dx
        \\ &\geq \int_{S_j} \check{f}^{(\zeta)}(x) \, dx
        \\ &\geq \int_{S_j} (1-\rho_1/2) \check{f}^{(\zeta)}(x_j) \, dx
        \\ &= (1-\rho_1/2) \check{f}^{(\zeta)}(x_j) \vol_\wspace (S_j)
        \\ &\geq (1-\rho_1/2)^2 \alpha  \check{f}^{(\zeta)}(x_j) \check{g}^{(\zeta)}(x_j) \varepsilon_0^\gamma
        \\ &\geq (1-\rho_1) \alpha  \check{f}^{(\zeta)}(x_j) \check{g}^{(\zeta)}(x_j) \varepsilon_0^\gamma \,.
    \end{align}

    Finally, claims (i)-(iii) show our result because
    \begin{align}
        \bbP_{X \sim f}[S_{j(X)} \text{ is } \rho_1\text{-good}] &\geq \bbP_{X \sim f}[S_{j(X)} \text{ is } \rho^*\text{-sufficient}] 
        \\ &\geq \bbP_{X \sim f}[\check{f}^{(\zeta)}(X) > (3/2) \rho^*] 
        \\ &\geq \rho_2
    \end{align}
    and we are done.
\end{proof}




For the remainder of this proof, we will assume that all cells are $\rho_1$-good. We will show in Appendix~\ref{appx:good-target-bad-target} that this assumption can be made without loss of generality, i.e. that the same very high probability bound can be achieved.

Given covering HCS's with roots $\bq_1, \dots, \bq_m$, we define
\begin{align}
    C_{\varepsilon_0} = \tsp_{\Pi}(\bq_1,\dots,\bq_m) 
\end{align}
i.e. the amount of TSP tour time needed to visit all the roots of the cells, which depends on $\varepsilon_0$ but is constant with regards to $n$. Note that this requires the system to visit particular \emph{configurations} rather than just targets, and note that computing it exactly can be intractable; however, we only need any constant length and, if we have good point-to-point (or configuration-to-configuration) motion planning for our system, we can even get a good constant approximation by computing the distances between all pairs of $\bq_{j_1}, \bq_{j_2}$ and applying a standard TSP approximation algorithm.
%

We now want to apply \Cref{prop::conc-bounds-azhoef} or \Cref{prop::Y-azuma-hoeffding-redux} (depending on whether $\gamma \geq 3$ since $\zetterm = 1-\frac{1}{\gamma}$). To do this, we need to compute $\sum_{j=1}^m p_j^{1-\frac{1}{\gamma}}$. We note that if we define a discrete random variable $Z \sim \bp$ in $[m]$, this is equivalent to
\begin{align}
    \sum_{j=1}^m p_j^{1-\frac{1}{\gamma}} = \bbE_{Z \sim \bp} \Big[p_Z^{-\frac{1}{\gamma}}\Big] 
\end{align}
Let $X \sim f$; then we can interpret $Z = j(X)$. For any $\delta > 0$, we can choose $\rho_1 > 0$ and $\zeta > 0$ and $\varepsilon_0$ small enough that \Cref{lem:prob-of-cells} applies and
\begin{align}
    \int_{\wspacef} f(x) \check{f}^{(\zeta)}(x)^{-\frac{1}{\gamma}} \check{g}^{(\zeta)}(x)^{-\frac{1}{\gamma}} \, dx \approx \int_{\wspacef} f(x)^{1-\frac{1}{\gamma}} g(x)^{-\frac{1}{\gamma}} \, dx
\end{align}
(this approximation can be made arbitrarily close by setting $\zeta$ small) to get
\begin{align}
    \sum_{j=1}^m p_j^{1-\frac{1}{\gamma}} &= \bbE_{Z \sim \bp} \Big[p_Z^{-\frac{1}{\gamma}}\Big] 
    \\ &= \bbE_{X \sim f} \Big[p_{j(X)}^{-\frac{1}{\gamma}}\Big] 
    \\ &\leq \bbE_{X \sim f} \Big[(1-\rho_1)^{-\frac{1}{\gamma}} \alpha^{-\frac{1}{\gamma}} \varepsilon_0^{-1} \check{f}^{(\zeta)}(X)^{-\frac{1}{\gamma}} \check{g}^{(\zeta)}(X)^{-\frac{1}{\gamma}}\Big]
    \\ &= (1-\rho_1)^{-\frac{1}{\gamma}} \alpha^{-\frac{1}{\gamma}} \varepsilon_0^{-1} \int_{\wspacef} f(x) \check{f}^{(\zeta)}(x)^{-\frac{1}{\gamma}} \check{g}^{(\zeta)}(x)^{-\frac{1}{\gamma}} \, dx
    \\ &\leq (1+\delta) \alpha^{-\frac{1}{\gamma}} \varepsilon_0^{-1} \int_{\wspacef} f(x)^{1-\frac{1}{\gamma}} g(x)^{-\frac{1}{\gamma}} \, dx
\end{align}
By \Cref{prop::hcp-upper-bound}, if $n_j$ is the number of targets that fall into $S_j$, the time needed to visit all of them (starting and ending at the root $\bq_j$) is at most $6 s \varepsilon_0 n_j^{1-\frac{1}{\gamma}}$ and hence a TSP trajectory can be constructed by combining a TSP trajectory through the roots $\bq_1, \dots, \bq_m$ (taking a constant $C_{\varepsilon_0}$ time) and these tours within cells; each time you arrive at a $\bq_j$, you execute a tour of the $n_j$ targets within $S_j$. Thus,
\begin{align}
    \tsp_{\Pi}(X_1, &\dots, X_n) \leq C_{\varepsilon_0} + 6 s \varepsilon_0 \sum_{j=1}^m  n_j^{1-\frac{1}{\gamma}}
    \\ \wvhp~~&\leq C_{\varepsilon_0} + 12 s \varepsilon_0 n^{1-\frac{1}{\gamma}} \sum_{j=1}^m  p_j^{1-\frac{1}{\gamma}}
    \\ \wvhp~~&\leq C_{\varepsilon_0} + 12 s \varepsilon_0 n^{1-\frac{1}{\gamma}} (1-\rho_1)^{-\frac{1}{\gamma}} \alpha^{-\frac{1}{\gamma}} \varepsilon_0^{-1} \int_{\wspacef} f(x) \check{f}^{(\zeta)}(x)^{-\frac{1}{\gamma}} \check{g}^{(\zeta)}(x)^{-\frac{1}{\gamma}} \, dx 
    \\ \wvhp~~&\leq C_{\varepsilon_0} + (1+\delta) 12 s \alpha^{-\frac{1}{\gamma}}n^{1-\frac{1}{\gamma}} \int_{\wspacef} f(x)^{1-\frac{1}{\gamma}} g(x)^{-\frac{1}{\gamma}} \, dx 
\end{align}
Exactly what `wvhp' means here is dependent on what regime from \Cref{prop::Y-azuma-hoeffding-redux,prop::conc-bounds-azhoef} the exponent $\zetterm = 1-\frac{1}{\gamma}$ falls in, but in all cases probability of failure is upper bounded (in the limit) by some $c_1 e^{-c_2 x^{c_3}}$ where $c_1, c_2, c_3 > 0$. Note that while $C_{\varepsilon_0}$ increases as $\varepsilon_0$ gets smaller, and that $\varepsilon_0$ might have to be set small to achieve approximation error $\delta > 0$, it is constant with regard to $n$ and hence negligible as $n \to \infty$.

\begin{remark}
A fun consequence of this argument is that, in the Euclidean TSP on $\bbR^d$, if the target points $X_1, \dots, X_n$ are drawn from a bounded set $\wspace^* \subseteq \bbR^d$ with Hausdorff dimension $\gamma \geq 1$, then the worst-case tour length scales according to $\Theta(n^{1 - \frac{1}{\gamma}})$.

The requirement that $\gamma > 1$ comes from the fact that if $\gamma < 1$, the time required to move between hierarchical cell structures, which is $O(1)$, dominates over $n^{1-\frac{1}{\gamma}}$. Thus, if $\gamma < 1$ the Euclidean TSP trajectory length defaults back to $O(1)$, i.e. bounded above by a constant even as $n \to \infty$.
\end{remark}

\section{Adversarial Targets} \label{sec:worst-case}

We now show \Cref{thm:adv-dtsp-bounds}, which deals with the length of the tour when the targets are placed (within a bounded $\wspace^* \subseteq \wspace$) to maximize the tour length. We refer to this as \emph{adversarial} target placement. While our adversarial target lower bound will hold for non-symmetric dynamics, as in \Cref{sec:dstsp-ub} our focus here is on symmetric $\Pi$.

\begin{remark}
    When $\Pi$ is nonsymmetric, there are counterexamples in which adversarial target placement results in a tour length $\Theta(n)$; for instance, if we have a Dubins car with turning radius $1$, we can distribute our targets evenly over the boundary of a circle with radius $1/100$. Then at each `pass' through the circle, the system can visit at most $2$ target points, and it takes $\approx 2\pi$ time to return, thus yielding a tour of length $\approx \pi n$.
\end{remark}

As discussed in \Cref{sec:results}, we consider two related questions which follow naturally from \Cref{thm:dstsp-bounds}:
\begin{enumerate}
    \item Given symmetric dynamics $\Pi$ and bounded (but full dimensional) target region $\wspace^*$, what target point density $f$ over $\wspace^*$ maximizes the length of the optimal TSP trajectory when $X_1, \dots, X_n \simt{iid} f$ as $n \to \infty$?
    \item Given symmetric dynamics $\Pi$, a bounded  set $\wspace^* \subseteq \wspacef$ and a number of target points $n$, what is $\sup_{X_1, \dots, X_n \in \wspace^*} \tsp_\Pi(X_1,\dots,X_n)$?
\end{enumerate}
Although we have lower and upper bounds on the tour length which are tight up to a fixed constant factor for all $f$, we cannot claim to know exactly what $f$ maximizes the DSTSP tour length (either in expectation or with high probability). However, we know that both the lower and upper bounds to our tour length are of the form
\begin{align}
    c n^{1-\frac{1}{\gamma}} \int_{\wspacef} f(x)^{1-\frac{1}{\gamma}} g(x)^{-\frac{1}{\gamma}} \, dx
\end{align}
where $c$ is in both cases a constant which does not depend on $f$. Therefore, instead of question 1, we will answer the related question of: given symmetric dynamics $\Pi$ and bounded (but full dimensional) target region $\wspace^*$, what target point density $f$ over $\wspace^*$ maximizes
\begin{align}
    \int_{\wspace^*} f(x)^{1-\frac{1}{\gamma}} g(x)^{-\frac{1}{\gamma}} \, dx
\end{align}
and what value does this maximum take?

Question 2 then asks for a deterministic upper bound on the tour length, with targets chosen by an adversary to maximize tour length. While adversarially-chosen targets are by definition going to produce a longer tour length than any random distribution, we are interested in knowing whether, as $n \to \infty$, it is possible to do substantially better than the worst-case $f$ from question 1. Note that question 1 requires a single probability distribution to be used for all values of $n$ as $n \to \infty$, while question 2 allows a different set of points to be selected depending on $n$. While this seems like it might allow more flexibility for the adversarial target points, we will show that roughly the same bound applies to the adversarial target points.

\subsection{Adversarial randomness} \label{sec:lb_worst}

In order to show the lower bound from \Cref{thm:adv-dtsp-bounds}, we will find density $f_g$ over $\wspace^*$ which maximizes the lower bound from \Cref{thm:dstsp-bounds}, which is equivalent to maximizing $\int_{\wspace^*} f(x)^{1-\frac{1}{\gamma}} g(x)^{-\frac{1}{\gamma}} \, dx$; this clearly also serves as a lower bound to the longest tour from non-random target points. It also maximizes the upper bound from \Cref{thm:dstsp-bounds} thanks to the constant factor gap.

We thus have to solve the following problem:
\begin{align}\label{eq:worst_distribution}
    \text{maximize} \, &\int_{\wspace^*} f(x)^{{1 - \frac{1}{\gamma}}} g(x)^{-\frac{1}{\gamma}} dx ~~~\text{(where } f : \wspace^* \to \bbR \text{ is integrable)}
    \\ \text{subject to} \, &\int_ {\wspace^*} f(x) dx = 1 \text{, and } f(x) \geq 0 \text{ for all } x \in \wspace^* \label{eq:worst_distribution_constraints}
\end{align}
(i.e. the constraint is that $f$ is a probability density function over $\wspace^*$).

\begin{lemma}\label{lem:worst_distribution}
Objective~\eqref{eq:worst_distribution} under constraints~\eqref{eq:worst_distribution_constraints} is maximized by $f^*(x) \, \propto \, g(x)^{-1}$, i.e.
\begin{align}
    f^*(x) := \frac{g(x)^{-1}}{\int_{\wspace^*}g(y)^{-1} dy} \mathrm.
\end{align}
\end{lemma}

\begin{proof}
First, we note that $f^*$ trivially satisfies the constraints in \eqref{eq:worst_distribution_constraints}: it is normalized so it integrates to $1$ over $\wspace^*$, and it is nonnegative since $g(x)$ (and hence $g(x)^{-1}$) is nonnegative. We also note that it exists since $g(x) \geq g_{\min}$ over $\wspace^*$ and $\wspace^*$ has finite volume (since it is bounded), so $\int_{\wspace^*} g(y)^{-1} \, dy \leq g_{\min}^{-1} \vol_{\wspace}(\wspace^*) < \infty$.

We then note that scaling $g(x)$ by a constant $c$ yields
\begin{align}
    \int_{\wspace^*} f(x)^{1-\frac{1}{\gamma}} (c \, g(x))^{-\frac{1}{\gamma}} \, dx = c^{-\frac{1}{\gamma}} \int_{\wspace^*} f(x)^{1 - \frac{1}{\gamma}} g(x)^{-\frac{1}{\gamma}} \, dx
\end{align}
Thus it does not affect what $f$ maximizes the objective function; hence we may assume without loss of generality that $\int_{\wspace^*} g(x)^{-1} \, dx = 1$ (of course we will have to put in the correct scale when evaluating the optimal value).

We will now show \Cref{lem:worst_distribution} using H{\"o}lder's Inequality. We first define:
\begin{align}\tilde{f}(x) \defeq f(x)^{{1 - \frac{1}{\gamma}}} \text{ and } \tilde{g}(x) = g(x)^{-\frac{1}{\gamma}} \end{align}
and note that since by definition $f(x), g(x) \geq 0$ for all $x$, we know that $\tilde{f}(x), \tilde{g}(x) \geq 0$ for all $x$, so we can ignore the absolute value function in the statement of H{\"o}lder's Inequality. We then define the constant $\eta \defeq \frac{\gamma}{\gamma - 1}$. Note that $\frac{1}{\eta} + \frac{1}{\gamma} = 1$, as required by H{\"o}lder's Inequality. Thus:
\begin{align}\int_{\wspace^*} \tilde{f}(x) \, \tilde{g}(x) dx \leq \Big( \int_{\wspace^*} \tilde{f}(x)^\eta dx \Big)^{\frac{1}{\eta}} \, \Big( \int_{\wspace^*} \tilde{g}(x)^{\gamma} dx \Big)^{\frac{1}{\gamma}} \mathrm. \end{align}
But, using the definitions from above,
\begin{align}\tilde{f}(x)^\eta = \big(f(x)^{{1 - \frac{1}{\gamma}}}\big)^{\frac{\gamma}{\gamma - 1}} = f(x) \text{ and } \tilde{g}(x)^{\gamma} = \big(g(x)^{-\frac{1}{\gamma}}\big)^{\gamma} = g(x)^{-1} \end{align}
so we can rewrite the inequality as
\begin{equation}\label{eq:holder}
\int_{\wspace^*} \tilde{f}(x) \, \tilde{g}(x) dx \leq \Big( \int_{\wspace^*} f(x) dx \Big)^{\frac{1}{\eta}} \, \Big( \int_{\wspace^*} g(x)^{-1} dx \Big)^{\frac{1}{\gamma}}\mathrm;
\end{equation}
however, by the condition that $f$ is a probability density function and our assumption (without loss of generality) about $g(x)^{-1}$, we know that
\begin{align}\int_{\wspace^*} f(x) dx = \int_{\wspace^*} g(x)^{-1} dx = 1\end{align}
implying that the right hand side of expression~\eqref{eq:holder} is just $1$. Thus,
\begin{align}\int_{\wspace^*} f(x)^{{1 - \frac{1}{\gamma}}} \, g(x)^{-\frac{1}{\gamma}} dx = \int_{\wspace^*} \tilde{f}(x) \, \tilde{g}(x) dx \leq 1\end{align}
for any probability density function $f$. But, using $f^*$ as defined in \Cref{lem:worst_distribution}, it is trivial to see that
\begin{align}\int_{\wspace^*} f^*(x)^{{1 - \frac{1}{\gamma}}} \, g(x)^{-\frac{1}{\gamma}} dx = 1\end{align}
thus showing that $f^*$ is the maximizing density function.
\end{proof}

\begin{remark}
Density $f^*$ makes the lucrativity function constant over $\wspace^*$, as for all $x \in \wspace^*$,
\begin{align}
    \cost^*(x) = (f^*(x) g(x))^{\frac{1}{\gamma}} = \Big(\int_{\wspace^*} g(y)^{-1} \, dy \Big)^{-\frac{1}{\gamma}} \,.
\end{align}
\end{remark}

\begin{lemma}
The optimal value of the problem given in \eqref{eq:worst_distribution} and \eqref{eq:worst_distribution_constraints} is
\begin{align}
    \max_{f} \int_{\wspace^*} f(x)^{1-\frac{1}{\gamma}} g(x)^{-\frac{1}{\gamma}} \, dx = \Big(\int_{\wspace^*} g(x)^{-1} \, dx \Big)^{-\frac{1}{\gamma}}
\end{align}
\end{lemma}

\begin{proof}
Plugging $f^*$ into expression~\eqref{eq:worst_distribution} gives
\begin{align}
    \int_{\wspace^*} f^*(x)^{{1 - \frac{1}{\gamma}}} g(x)^{-\frac{1}{\gamma}} dx &= \int_{\wspace^*} \Big(\frac{g(x)^{-1}}{\int_{\wspace^*}g(y)^{-1} dy} \Big)^{{1 - \frac{1}{\gamma}}} \, g(x)^{-\frac{1}{\gamma}} dx 
    \\ &=  \int_{\wspace^*} \frac{g(x)^{-1}}{\big(\int_{\wspace^*}g(y)^{-1} dy\big)^{{1 - \frac{1}{\gamma}}}} dx  
    \\ &= \frac{\int_{\wspace^*}g(x)^{-1} dx}{\big(\int_{\wspace^*}g(y)^{-1} dy\big)^{{1 - \frac{1}{\gamma}}}} 
    \\ &= \Big( \int_{\wspace^*} g(x)^{-1} dx \Big)^{\frac{1}{\gamma}} \mathrm.
\end{align}
\end{proof}

Thus, if we distribute the points $X_1, X_2, \dots, X_n$ according to $f^*$ (iid), by \Cref{thm:dstsp-bounds}, for any $\delta > 0$
\begin{align}
    \tsp_\Pi(X_1, \dots, X_n) \leq (1+\delta) \beta^{-1} n^{1-\frac{1}{\gamma}} \Big( \int_{\wspace^*} g(x)^{-1} dx \Big)^{\frac{1}{\gamma}} ~~~~ \wvhp
\end{align}
holds for sufficiently large $n$. Thus, the deterministic $X_1, \dots X_n$ in $\wspace^*$ which maximize the tour length achieves at least this length, giving us the lower bound in \Cref{thm:adv-dtsp-bounds}. 


\subsection{Adversarial targets upper bound}

We now prove the upper bound in \Cref{thm:adv-dtsp-bounds}. Consider a Hierarchical Cell Structure over $\wspace^*$ with scaling parameter $s$ and overlap parameter $\alpha$ at scale $\varepsilon_0$, which we know exists. The HCS induces a specific algorithm, which by \Cref{prop::hcp-upper-bound} generates a tour of length bounded above by
\begin{align} \label{eq:hcs-ub-adv}
    6 s \varepsilon_0 \sum_{j=1}^m n_j^{1-\frac{1}{\gamma}} + C_{\varepsilon_0}
\end{align}
where $n_j$ is the number of targets that falls into cell $j$ and $C_{\varepsilon_0}$ is a constant denoting the time needed to travel between cells. Let us denote $p_j = n_j/n$ and $\bp = (p_1, \dots, p_m)$; since $\sum_j n_j = n$, we know that $\bp$ is a probability vector. Ignoring the constant $C_{\varepsilon_0}$, we can re-write the above as
\begin{align} \label{eq:pjs}
    6 s \varepsilon_0 \sum_{j=1}^m n_j^{1-\frac{1}{\gamma}} = 6 s \varepsilon_0 n^{1-\frac{1}{\gamma}} \sum_{j=1}^m p_j^{1-\frac{1}{\gamma}}
\end{align}
Note that in contrast with \Cref{sec:dstsp-ub}, here we can simply distribute the target points according to $n_1, \dots, n_m$ so there is no need for probabilistic analysis. Nevertheless, it will be useful to view $\bp$ as a probability vector and, as before, set $Z \sim \bp$ and to consider 
\begin{align}
    \sum_{j=1}^m p_j^{1-\frac{1}{\gamma}} = \bbE_{Z \sim \bp}[p_Z^{-\frac{1}{\gamma}}]
\end{align}
We then consider the probability distribution induced over the cells by $X \sim f^*$ where $f^*$ is the worst-case probability distribution over $\wspace^*$, i.e. $f^*(x) \, \propto \, g(x)^{-1}$. Let $j(x)$ be the index of the cell $x \in \wspace^*$ falls into; if $\bp^*$ is induced this way, then $Z \sim \bp^*$ is equivalent to $Z = j(X)$ where $X \sim \bp^*$. This is
\begin{align}
    p_j^* = \bbP_{X \sim f^*}[X \in S_j] = \int_{S_j} f^*(x) \, dx = \frac{\int_{S_j} g(x)^{-1} \, dx}{\int_{\wspace^*} g(y)^{-1} \, dy} 
\end{align}
Therefore, we can re-write
\begin{align}
    \bbE_{Z \sim \bp^*}[(p_Z^*)^{-\frac{1}{\gamma}}] &= \bbE_{Z \sim \bp^*} \Bigg[ \bigg(\frac{\int_{S_Z} g(x)^{-1} \, dx}{\int_{\wspace^*} g(y)^{-1} \, dy} \bigg)^{-\frac{1}{\gamma}}  \Bigg] 
    \\ &= \Big(\int_{\wspace^*} g(y)^{-1} \, dy\Big)^{\frac{1}{\gamma}} \bbE_{Z \sim \bp^*}\bigg[\Big( \int_{S_Z} g(x)^{-1} \, dx \Big)^{-\frac{1}{\gamma}} \bigg]
\end{align}
Finally, we will analyze $\int_{S_j} g(x)^{-1} \, dx$ for any $j \in [m]$. Note that we can make $\varepsilon_0$ arbitrarily small because a HCS at scale $\varepsilon_0$ with scale factor $s$ and overlap parameter $\alpha$ can be changed to a HCS at scale $\varepsilon_0/s$ with the same parameters (and then to an HCS at scale $\varepsilon_0/s^2$ and so forth) by taking its sub-cells as the new cells. Therefore, letting $\check{g}^{(\eta)}$ be the $\eta$-regularized lower approximation of $g$, for any $\delta_1, \delta_2 > 0$ we can set $\varepsilon_0$ sufficiently small so that for any cell center $\bq_j$ and configuration $\bq'$ such that $d_{\Pi}(\bq,\bq') \leq \varepsilon_0$,
\begin{align}
    &\vol_{\wspace}(S_j) \geq (1-\delta_1) \alpha \vol_{\wspace}(\bar{R}_{\varepsilon_0}(\bq_j)) \geq (1-\delta_2)(1-\delta_1) \alpha \check{g}^{(\eta)}(\proj{\bq}) \varepsilon_0^\gamma
    \\ \implies &\check{g}^{(\eta)}(x) \leq (1-\delta_1)^{-1} (1-\delta_2)^{-1} \alpha^{-1} \varepsilon_0^{-\gamma} \vol_{\wspace}(S_j)
    \\ \implies &\check{g}^{(\eta)}(x)^{-1} \geq (1-\delta_1) (1-\delta_2) \alpha \varepsilon_0^\gamma \vol_{\wspace}(S_j)^{-1}
\end{align}
for all $x \in S_j$ (since $S_j \subseteq \bar{R}_{\varepsilon_0}(\bq)$). Furthermore, for any $\delta_3 > 0$, we can set $\eta$ sufficiently small so that for all $j$,
\begin{align}
    \int_{S_j} g(x)^{-1} \, dx \geq (1-\delta_3) \int_{S_j} \check{g}^{(\eta)}(x)^{-1} \, dx
\end{align}
But then we can conclude that for any $j$,
\begin{align}
    \int_{S_j} g(x)^{-1} \, dx &\geq (1-\delta_3) \int_{S_j} \check{g}^{(\eta)}(x)^{-1} \, dx
    \\ &\geq (1-\delta_1) (1-\delta_2) (1-\delta_3)  \alpha \varepsilon_0^\gamma \int_{S_j} \vol_{\wspace}(S_j)^{-1} \, dx
    \\ &= (1-\delta_1) (1-\delta_2) (1-\delta_3)  \alpha \varepsilon_0^\gamma 
\end{align}
and hence we can flip it around and get
\begin{align}
    \bigg( \int_{S_j} g(x)^{-1} \, dx \bigg)^{-\frac{1}{\gamma}} &\leq (1-\delta_1)^{-\frac{1}{\gamma}} (1-\delta_2)^{-\frac{1}{\gamma}} (1-\delta_3)^{-\frac{1}{\gamma}}  \alpha^{-\frac{1}{\gamma}} \varepsilon_0^{-1} \,.
\end{align}
Since this holds for all $j$, we know that
\begin{align}
    \bbE_{Z \sim \bp^*}\bigg[\Big( \int_{S_Z} g(x)^{-1} \, dx \Big)^{-\frac{1}{\gamma}} \bigg] \leq (1-\delta_1)^{-\frac{1}{\gamma}} (1-\delta_2)^{-\frac{1}{\gamma}} (1-\delta_3)^{-\frac{1}{\gamma}}  \alpha^{-\frac{1}{\gamma}} \varepsilon_0^{-1} \,.
\end{align}
We can then plug this in to get a bound on $\bbE_{Z \sim \bp}[(p_Z^*)^{-\frac{1}{\gamma}}] = \sum_{j=1}^m (p_j^*)^{1-\frac{1}{\gamma}}$. 
Then we plug the bound into expression \eqref{eq:pjs} to get
\begin{align}
    6 s \varepsilon_0 n^{1-\frac{1}{\gamma}} \sum_{j=1}^m (p_j^*)^{1-\frac{1}{\gamma}} &\leq 6 s  \alpha^{-\frac{1}{\gamma}} n^{1-\frac{1}{\gamma}} \Big(\int_{\wspace^*} g(y)^{-1} \, dy\Big)^{\frac{1}{\gamma}} \big((1-\delta_1)(1-\delta_2) (1-\delta_3)\big)^{-\frac{1}{\gamma}} 
\end{align}
One wrinkle is that while this is true of $\bp^*$ induced by $f^*$, our $n_j$'s must be integers which means they cannot in general exactly represent $\bp^*$. However, as $n \to \infty$, they $p_j = n_j/n$ can come arbitrarily close, and for any $\delta_4$ there is some $n^{(\delta_4)}$ such that for all $n \geq n^{(\delta_4)}$, there are some $n_1, \dots, n_m$ such that if $p_j = n_j/n$ then
\begin{align}
    \sum_{j=1}^m p_j^{1-\frac{1}{\gamma}} \leq (1+\delta_4) \sum_{j=1}^m (p_j^*)^{1-\frac{1}{\gamma}}
\end{align}
Then, for any $\delta_5 > 0$, we can set $\delta_1, \delta_2, \delta_3, \delta_4 > 0$ such that
\begin{align}
    1 + \delta_5 = (1-\delta_1)^{-\frac{1}{\gamma}} (1-\delta_2)^{-\frac{1}{\gamma}} (1-\delta_3)^{-\frac{1}{\gamma}} (1+\delta_4)
\end{align}
which then gives that for $n \geq n^{(\delta_4)}$, distributing the $n_j$ according to this approximation of $p^*_j n$, we get that
\begin{align}
    6 s \varepsilon_0 \sum_{j=1}^m n_j^{1-\frac{1}{\gamma}} &\leq (1+\delta_5) 6 s  \alpha^{-\frac{1}{\gamma}} n^{1-\frac{1}{\gamma}} \Big(\int_{\wspace^*} g(y)^{-1} \, dy\Big)^{\frac{1}{\gamma}}
\end{align}
Thus, for sufficiently large $n$, the HCS induces an algorithm which collects all the target points in at most $(1+\delta_5) 6 s \alpha^{-\frac{1}{\gamma}} n^{1-\frac{1}{\gamma}} \big(\int_{\wspace^*} g(y)^{-1} \, dy\big)^{\frac{1}{\gamma}} + C_{\varepsilon_0}$ length; and for any $\delta > 0$ we can choose $0 < \delta_5 < \delta$ and choose $n$ sufficiently large that
\begin{align}
    (1+\delta_5) 6 s \alpha^{-\frac{1}{\gamma}} n^{1-\frac{1}{\gamma}} \Big(\int_{\wspace^*} g(y)^{-1} \, dy\Big)^{\frac{1}{\gamma}} + C_{\varepsilon_0} \leq (1+\delta) 6 s \alpha^{-\frac{1}{\gamma}} n^{1-\frac{1}{\gamma}} \Big(\int_{\wspace^*} g(y)^{-1} \, dy\Big)^{\frac{1}{\gamma}}
\end{align}
Since our specific algorithm can then always collect the target points in at most this amount of time, the optimal algorithm will do at least as well, and hence
\begin{align}
    \sup_{X_1, \dots, X_n \in \wspace^*} \tsp_\Pi(X_1, \dots, X_n) \leq (1+\delta) 6 s \alpha^{-\frac{1}{\gamma}} n^{1-\frac{1}{\gamma}} \Big(\int_{\wspace^*} g(y)^{-1} \, dy\Big)^{\frac{1}{\gamma}}
\end{align}
and we are done with the upper bound to \Cref{thm:adv-dtsp-bounds}.

\begin{remark}
    Since $\sum_j p_j = 1$ by definition and $1-\frac{1}{\gamma} < 1$, the optimal values of $p_1, \dots, p_m$ to maximize $\sum_{j=1}^m p_j^{1-\frac{1}{\gamma}}$ are actually $p_1 = \dots = p_m = \frac{1}{m}$, which yields
    \begin{align}
        \sum_{j=1}^m p_j^{1-\frac{1}{\gamma}} = m^{\frac{1}{\gamma}} \,.
    \end{align}
    This aligns with the intuition that to make the target points hard to visit, they should be spread out over the cells (at a given scale) evenly. While this was not used in the analysis as $m$ is hard to compute directly, it is useful to know in order to find $n$ sufficiently large so that every $p_j = \frac{n_j}{n}$ can approximate $\frac{1}{m}$. This `sufficiently large' value increases with $m$, which increases as the scale $\varepsilon_0$ decreases.
\end{remark}

\section{Conclusion} \label{sec:conclusion}

In this work we gave a very general formulation of the Dynamic Stochastic Traveling Salesman Problem (DSTSP), capturing a wide range of dynamic systems (with a specific focus on control-affine systems, though not limited to such), which extends previous formulations to systems on manifold (rather than Euclidean) workspaces. To do this, we develop a novel technique of discretizing the workspace, adapted from Arias-Castro et al. (\cite{arias-ctd-05}, 2005) and Adler, Karaman (\cite{adler-icra-16}, 2016), as well as a novel Dynamic TSP algorithm for symmetric dynamically-constrained systems based on Hierarchical Cell Structures. We also showed that our techniques work even in cases where the natural properties needed for the system to be well-behaved are only approximate.

Under this expanded formulation, we showed novel results describing not only how the length of the tour scales with the number $n$ of iid targets, but also the effect of the density function $f$ of the target point distribution on the length of the tour. In particular, we defined the \emph{agility function} $g$ induced by the dynamics over the workspace and showed for all dynamics in our formulation a very-high-probability lower bound on the tour length that both scale according to
\begin{align} \label{eq:f-g-interaction}
    n^{1-\frac{1}{\gamma}}\int_{\wspacef} f(x)^{1-\frac{1}{\gamma}} g(x)^{-\frac{1}{\gamma}} \, dx
\end{align}
where $\wspacef$ is the support of $f$, and for \emph{symmetric} dynamics a matching upper bound which scales according to \eqref{eq:f-g-interaction}. For symmetric dynamics, this yields very-high-probability lower and upper bounds that differ by only a constant in terms of $f$ and $n$.

We then use these results to analyze the case of \emph{adversarial} target points, in which the targets are distributed by an adversary on some bounded region $\wspace^*$ with the objective of maximizing the tour length. For symmetric systems, we show similar (but deterministic) matching lower and upper bounds which scale according to
\begin{align}
    n^{1-\frac{1}{\gamma}} \bigg(\int_{\wspace^*} g(x)^{-1} \, dx\bigg)^{\frac{1}{\gamma}}
\end{align}
with the lower bound also applying to non-symmetric systems.

\subsection{Future work}

While we establish lower and upper bounds to the tour lengths of the DSTSP and the Adversarial DTSP which are tight up to a constant factor in both $f$ and $n$, many interesting lines of future work remain.

Our work establishes a $12 s \alpha^{-\frac{1}{\gamma}} \beta$ multiplicative gap between the lower and upper bounds, where $s$ (typically $s = 2$ for control-affine systems) and $\alpha$ are parameters measuring the effectiveness of our cell-tiling algorithm, and $\beta$ is a parameter describing the difficulty of covering larger reachable sets with smaller ones at small scales. In particular, depending on the dynamics, $\alpha \leq 1$ is the degree to which the tiles fill the reachable sets in which they are inscribed, and generally measures how efficiently the reachable sets can be used to tile the workspace. However, it is not settled whether $\alpha$ really represents how efficiently the system can perform the TSP, or whether it is a product of our specific Hierarchical Cell Structure algorithm.

\Cref{thm:dstsp-bounds} almost completely extends the result of Beardwood et al. (\cite{beardwood-59}, 1959) on the Euclidean Stochastic TSP to general symmetric dynamic constraints. In particular, the Beardwood-Halton-Hammersley Theorem states that for the Euclidean Stochastic TSP on $\bbR^d$ with targets $X_1, X_2, \dots \simt{iid} f$, there is some constant $\beta_d$ (depending only on $d$) such that
\begin{align}
    \tsp(X_1, \dots, X_n) \to \beta_d n^{1-\frac{1}{d}} \int_{\wspacef} f(x)^{1-\frac{1}{d}} \, dx
\end{align}
almost surely as $n \to \infty$. Our result extends the $n^{1-\frac{1}{d}}$ to $n^{1-\frac{1}{\gamma}}$ (Euclidean paths in $\bbR^d$ can be formulated as a dynamic system with $\gamma = d$) and extends $\int f(x)^{1-\frac{1}{d}} \, dx$ to $\int f(x)^{1-\frac{1}{\gamma}} g(x)^{-\frac{1}{\gamma}} \, dx$ (when considering only the length of Euclidean paths, $g$ is constant and can therefore be moved to the constant outside the integral); however, it leaves open the possibility that the constant outside fluctuates between our derived lower and upper bounds and doesn't go to a particular value in the range $[\beta^{-1}, 12 s \alpha^{-\frac{1}{\gamma}}]$. Showing that there is some constant $\beta_\Pi$ for any symmetric dynamic constraints $\Pi$ and $X_1, X_2, \dots \simt{iid} f$
\begin{align}
    \lim_{n \to \infty} \frac{\tsp_\Pi(X_1, \dots, X_n)}{n^{1-\frac{1}{\gamma}} \int_{\wspacef} f(x)^{1-\frac{1}{\gamma}} g(x)^{-\frac{1}{\gamma}} \, dx} = \beta_\Pi
\end{align}
would complete the extension of the Beardwood-Halton-Hammersley Theorem to our setting, though it is also possible that such an extension might only be possible for $\Pi$ satisfying stronger regularity conditions.

Furthermore, the techniques developed here (hierarchical cell tiling strategies and Cost-Balanced Orienteering) may yield results on the following extensions of or problems related to the DSTSP or Stochastic Orienteering:

\begin{itemize}
    \item As discussed in Appendix~\ref{appx:fractal-targets}, our techniques can yield strong upper bounds for the Euclidean TSP where the targets are distributed over a subset of fractal dimension; in particular, the Hierarchical Cell Structure is suited to such cases. However, this leaves open the question of fractal distribution of targets with dynamically-constrained tours, as well as matching lower bounds.

    \item Another interesting extension may be to have targets be sets within the space rather than points. While having the targets be balls of fixed size is generally trivial in the limit (a fixed-length trajectory can `sweep' the space and visit all possible $\varepsilon$-balls), possibilities such as having the targets be random $k$-dimensional affine subspaces embedded in a $d$-dimensional workspace may be solvable with our techniques.

    \item The Orienteering problem can also be thought of as trying to pass a curve of limited length obeying constraints through as many targets as possible, which has applications in image processing~\cite{arias-ctd-05}, and an important extension of this is to pass a higher-dimensional region of limited size obeying dynamic constraints through targets.
    
    \item Another problem related to Orienteering is the problem of trying to fit a function satisfying local constraints (for instance, Lipschitz continuity) to target points with a loss function (as opposed to trying to exactly pass through as many targets as possible).

    \item Finally, the TSP is related to a number of other problems in which mobile agents must visit unordered targets, such as the Traveling Repairperson Problem~\cite{itani-thesis-12} in which targets pop up stochastically (for instance, according to a Poisson process) over time and must be visited with a minimum delay per target.
\end{itemize}

Finally, it remains open whether a guaranteed approximation algorithm exists for the Dynamic TSP in general. \Cref{thm:dstsp-bounds} shows that our Hierarchical Collection Problem algorithm achieves a constant factor approximation of the shortest tour with very high probability; this in turn shows that it also achieve a constant factor approximation of the expected length of the shortest tour (see \Cref{cor:dstsp-expectation} and Appendix~\ref{appx:wvhp}). Finally, \Cref{thm:adv-dtsp-bounds} shows that our HCP algorithm achieves a constant factor approximation of the shortest tour through adversarially-distributed target points. However, our analysis leaves open the possibility that the targets may be cleverly placed so as to permit a very short tour while the algorithms only find longer ones (though when the targets are random this becomes vanishingly unlikely). A guaranteed approximation algorithm (as opposed to an algorithm like the HCP algorithm which finds an approximate solution with very high probability or in expectation) would find a tour whose length is within a constant multiplicative factor of the shortest tour on \emph{any} set of $n$ target points.

\bibliographystyle{resources/IEEEbib}
\bibliography{refs}

\newpage

\appendices

\section{With Very High Probability} \label{appx:wvhp}

Recall that a sequence of events $\{A_n\}_{n \in \bbZ_{\geq 0}}$ happens \emph{with very high probability} if there are $c_1, c_2, c_3 > 0$ such that for all sufficiently large $n$, $\bbP[A_n] \geq 1 - c_1 e^{-c_2 n^{c_3}}$. We will explore important implications for this, as well as ways in which it differs from the more common definition of `with high probability' (in which $\{A_n\}_{n \in \bbZ_{\geq 0}}$ happens with high probability if $\lim_{n \to 0} \bbP[A_n] = 1$, without any additional condition on the speed of convergence).

Essentially, \wvhp~converges to $1$ faster than any inverse polynomial, which yields additional information for which `with high probability' is not sufficient:
\begin{lemma} \label{lem:wvhp-as-01}
If $\{A_n\}$ happens with very high probability, then
\begin{align}
	\bbP[ (\max n : \neg A_n) < \infty ] = 1
\end{align}
i.e. almost surely there is some finite $n^*$ s.t. $A_n$ happens for all $n > n^*$.
\end{lemma}

\begin{proof} 
We show this by using the Borel-Cantelli Lemma, which states that if a sequence of events $\{E_n\}$ has $\sum_{n = 1}^\infty \bbP[E_n] < \infty$, then
\begin{align} 
	\bbP[(\max n: E_n ) < \infty] = 1
\end{align}
But letting $E_n = \neg A_n$, we have
\begin{align}
	\bbP[E_n] \leq c_1 e^{-c_2 n^{c_3}}
\end{align}
for sufficiently large $n$ (say, $n > n'$). Of course, for $n \leq n'$, we have the trivial bound $\bbP[E_n] \leq 1$. Thus,
\begin{align} 
	\sum_{n = 1}^\infty \bbP[E_n] \leq n' + \sum_{n > n'} c_1 e^{-c_2 n^{c_3}} < \infty
\end{align}
since $c_1 e^{-c_2 n^{c_3}}$ (which decreases faster than any $n^{-k}$) is a convergent series.

Therefore, $\bbP[(\max n: \neg A_n ) < \infty] = 1$, as we wanted.
\end{proof}

We also define a related notion of a random variable growing ``with very high probability'':
\begin{definition} \label{def:wvhp-def-2}
If $\{Z_n\}$ is a sequence of random variables and $\alpha(n) : \bbZ_{> 0} \to \bbR_{\geq 0}$ is any function then we say ``$Z_n \eqt{{wvhp}} O(\alpha(n))$'' if there exists a constant $c > 0$ such that the sequence of events $ A_n = \{Z_n \leq c \alpha(n)\} $ happens with very high probability.

We similarly say $Z_n \eqt{{wvhp}} \Omega(\alpha(n))$ if there exists a constant $c > 0$ such that $A_n = \{Z_n \leq c \alpha(n)\} $ happens with very high probability, and $Z_n \eqt{{wvhp}} \Theta(\alpha(n))$ if there exist constants $c_2 > c_1 > 0$ such that
$A_n = \{c_1 \alpha(n) \leq Z_n \leq c_2 \alpha(n)\}$ happens with very high probability.

Finally, we say that ``$Z_n \tot{{wvhp}} \alpha(n)$'' (multiplicatively) if there exists a nonnegative sequence $\delta_n$ satisfying $\lim_{n \to \infty} \delta_n = 0$ such that
\begin{align} 
	B_n = \{1 - \delta_n  \leq Z_n/\alpha(n) \leq 1 + \delta_n  \}
\end{align}
happens with very high probability. 
\end{definition}
The following then hold:

\begin{lemma}
If $Z_1, Z_2, \dots$ and $Z'_1, Z'_2, \dots$ are sequences of random variables parameterized by $n$ such that $Z_n \succeq Z'_n$ for all $n$, then 
\begin{align}
	Z_n \eqt{wvhp} O(\alpha(n)) &\implies Z'_n \eqt{wvhp} O(\alpha(n)) \label{eq::domination-wvhp-01} \\ \text{and }~~ Z'_n \eqt{wvhp} \Omega(\alpha(n)) &\implies Z_n \eqt{wvhp} \Omega(\alpha(n)) \label{eq::domination-wvhp-02}
\end{align}
\end{lemma}

\begin{proof}
This follows from the definitions: if $Z_n \eqt{wvhp} O(\alpha(n))$ then we have some $c > 0$ and $c_1, c_2, c_3 > 0$ such that
\begin{align}
	\bbP[Z_n > c \alpha(n)] \leq c_1 e^{-c_2 n^{c_3}}
\end{align}
Since $Z_n \succeq Z'_n$ implies $\bbP[Z'_n > c \alpha(n)] \leq \bbP[Z_n > c \alpha(n)]$, we are done with Eq.~\eqref{eq::domination-wvhp-01}. The proof of Eq.~\eqref{eq::domination-wvhp-02} is analogous.
\end{proof}

\begin{lemma} \label{lem:wvhp-as-02}
If $Z_n \tot{wvhp} \alpha(n)$, then $Z_n/\alpha(n) \to 1$ almost surely.
\end{lemma}

\begin{proof}
By Lemma~\ref{lem:wvhp-as-01}, with probability $1$ there is some finite $n^*$ such that $B_n$ happens for all $n > n^*$. Thus, for $n > n^*$, we have 
\begin{align}
	1 - \delta_n  \leq Z_n/\alpha(n) \leq 1 + \delta_n
\end{align}
and since $\delta_n \to 0$ as $n \to \infty$, the result follows.
\end{proof}

The next lemma (along with \Cref{thm:dstsp-bounds}) shows that our main results on the very-high-probability behavior of the DSTSP translate directly to expected value.
\begin{lemma} \label{lem:wvhp-E}
If $\{Z_n\}$ is a sequence of nonnegative random variables such that  $Z_n = \Theta(n^k)$ with very high probability and $Z_n = O(n^\ell)$ (non-probabilistic upper bound), where $0 < k \leq \ell$, then $ \bbE[Z_n]  = \Theta(n^k)$.
\end{lemma}

\begin{proof}
We give upper and lower bounds for $\bbE[Z_n]$ by splitting it into three parts based on the following events: let $A_n^{(1)} = \{Z_n < c_1 n^k\}$; $A_n^{(2)} = \{c_1 n^k \leq Z_n \leq c_2 n^k\}$ (where $0 < c_1 < c_2$ as in Definition~\ref{def:wvhp-def-2}); and $A_n^{(3)} = \{c_2 n^k < Z_n\}$. Then:
\begin{align}
\bbE[Z_n] &= \sum_{i=1}^3 \bbP[A_n^{(i)}] \, \bbE[Z_n \, | \, A_n^{(i)}]
\end{align}
We can then upper bound and lower bound these since: $\bbP[A_n^{(2)}] \geq 1 - b_1 e^{-b_2 n^{b_3}}$ (for $b_1, b_2, b_3 > 0$ as in Definition~\ref{def:wvhp-def-1}) we know that $\bbP[A_n^{(1)}], \bbP[A_n^{(3)}] \leq b_1 e^{-b_2 n^{b_3}}$; and $\bbE[Z_n \, | \, A_n^{(i)}] = O(n^\ell)$ (and are nonnegative), and $\bbE[Z_n \, | \, A_n^{(2)}] = \Theta(n^k)$ (by definition).

Then since $b_1 e^{-b_2 n^{b_3}} \to 0$ faster than $n^{-\ell}$, terms $i = 1$ and $3$ vanish as $n \to \infty$ while term $i = 2$ goes to $\Theta(n^k)$ and we are done.
\end{proof}

This shows that \Cref{cor:dstsp-expectation} follows directly from \Cref{thm:dstsp-bounds} and \Cref{prop:dtsp-trivial-bound}.


\section{Equiregular Control-Affine Dynamics} \label{appx:meta-assumption}

In this appendix we show how, if $\Pi$ is symmetric and \Cref{meta:regularity} holds (that $\Pi$ is control-affine and equiregular and $\projinv{\wspacef}$ is contained in a compact subset of $\cspace$), many of the assumptions in \Cref{sec:assumptions} follow, either directly or from some other simpler and easier-to-verify condition. We will rely heavily on the notation and terminology given in \emph{Control of Nonholonomic Systems} by Frederic Jean~\cite{jean-14}. We make use of the following notions:
\begin{itemize}
    \item We have control vector fields $h_1^{\mathrm{cont}},\dots,h_m^{\mathrm{cont}}$ (see \Cref{sec:control-affine}) over $d$-dimensional manifold $\cspace$ which define our control-affine system. For a vector field $h$ (not necessarily one of the control vector fields) we denote by $\exp(z h)$ the function which proceeds along $h$ for $z$ time, i.e. it takes $\bq \in \cspace$ and returns a new point $\exp(z h)(\bq)$ which is the point reached from $\bq$ by following $h$ for $z$ time units.
    \item The \emph{Lie algebra} of $h_1^{\mathrm{cont}},\dots,h_m^{\mathrm{cont}}$, which denote the set of directions which can be generated via linear combination and Lie brackets, and which we assume span the tangent space at each $\bq \in \cspace$ (Chow's Condition, \cite{jean-14} Definition 1.6). A Lie bracket $I$ from the \emph{free Lie algebra} on $[m]$ denotes a sequence of Lie bracket operations over $m$ unspecified elements; it can then be applied to our set of control vector fields. We denote by $h_I$ the vector in the Lie algebra over $h_1^{\mathrm{cont}}, \dots, h_m^{\mathrm{cont}}$ which corresponds to $I$, and we denote the length of Lie bracket $I$ (i.e. how many bracket operations needed to generate it) by $|I|$.
    \item \emph{Adapted frames} (\cite{jean-14}, page 36), which denote a set of vector fields which are used to generate a coordinate map around some $\bq \in \cspace$ (and which span $T_\bq(\cspace)$), in a way compatible with the Lie algebra; the boxes of the Uniform Ball-Box Theorem are defined via this coordinate map.
    \item \emph{Regular points} (\cite{jean-14}, Definition 2.4), which are configurations $\bq \in \cspace$ in which the Lie algebra behaves smoothly in a neighborhood around $\bq$.
\end{itemize}
These allow us to use our main tool:
\begin{unifbbt} [\cite{jean-14}, Theorem 2.4] 
    If $\cspace^* \subseteq \cspace$ is compact, there exist constants $0 < c < C$ and $\varepsilon_0 > 0$ such that, for $\bq \in \cspace^*$ and $0 < \varepsilon \leq \varepsilon_0$ and any adapted frame $\bH = (h_{I_1},\dots,h_{I_d})$ (a set of elements of the Lie algebra of the control vector fields) at $(\bq,\varepsilon)$,
    \begin{align}
        \boxx_{\bH}(\bq, c\varepsilon) \subset R_\varepsilon(\bq) \subset \boxx_{\bH}(\bq, C\varepsilon)
    \end{align}
    where $\boxx_{\bH}(\bq,\varepsilon) = \{\exp(z_1 h_{I_1}) \circ \dots \circ \exp(z_d h_{I_d})(\bq) : |z_i| \leq \varepsilon^{|I_i|} ~ \, \forall i\}$.
\end{unifbbt}
This states that at small scales $\varepsilon$ the configuration $\varepsilon$-reachable set $R_\varepsilon(\bq)$ contains a box of particular dimensions and orientation, and is in turn contained by a box of the same dimensions and orientation scaled up by a constant. The coordinate system in which these boxes are defined is given by $\bH$: a point at coordinates $(z_1, \dots, z_d)$ is the one reached from $\bq$ by traveling for $z_d$ along $h_{I_d}$, then $z_{d-1}$ along $h_{I_{d-1}}$, and so forth. The limits on $|z_i|$ give the dimensions of $\boxx_{\bH}(\bq,\varepsilon)$. This then yields, intuitively, a way to cover $R_\varepsilon(\bq)$ with a number of configuration $(\varepsilon/2)$-reachable sets, by taking the circumscribed box of $R_\varepsilon(\bq)$ and tiling it with the inscribed boxes of $R_{\varepsilon/2}(\bq')$ for a set of well-chosen $\bq'$. 

We begin with \Cref{lem:b-coverability}, which asserts that if \Cref{meta:regularity} holds for symmetric $\Pi$, then \Cref{assm:b-coverability} follows. 

\begin{proof}[Proof of \Cref{lem:b-coverability}]
We will prove that a slightly stronger version of \Cref{assm:b-coverability} holds, in which there is some $b > 1$ such that there is some $\varepsilon^* > 0$ such that for any $\bq \in \cspace$ and $0 < \varepsilon \leq \varepsilon^*$ there is a set $Q_\varepsilon(\bq)$ of configurations such that
\begin{align}
    R_{2\varepsilon}(\bq) \subseteq \bigcup_{\bq' \in Q_\varepsilon(\bq)} R_\varepsilon(\bq') \text{ and } |Q_\varepsilon(\bq)| \leq b \,.
\end{align}
This is stronger than \Cref{assm:b-coverability} in that it drops the $\rho > 0$ approximation factor, and therefore if this holds then it \Cref{assm:b-coverability} trivially holds as well (we can simply let $Q_{\rho,\varepsilon}(\bq) = Q_\varepsilon(\bq)$ for all $\rho > 0$). As a remark, \Cref{assm:b-coverability} is phrased in the more general way because there are instances where it yields a lower value of $b$, which improves our lower bound in \Cref{thm:dstsp-bounds}.

We choose some regular $\bq \in \projinv{\wspacef}$, which is contained inside some compact subset of $\cspace$. Thus, we can find an adapted frame $\bH$ at $\bq$ which is an adapted frame for all $\bq'$ in a neighborhood of $\bq$; let $\varepsilon$ be sufficiently small so that all $\bq' \in R_\varepsilon(\bq)$ are in this neighborhood, and that $\varepsilon < \varepsilon_0$ from the Uniform Ball-Box Theorem. Then, if $0 < c < C$ are as defined in the Uniform Ball-Box Theorem, 
\begin{align}
    R_\varepsilon(\bq) \subset \boxx_{\bH}(\bq,C\varepsilon) \,.
\end{align}
This is a box whose size along dimension $i \in [d]$ is $2 (C\varepsilon)^{|I_i|}$  Furthermore, for any $\bq' \in R_\varepsilon(\bq)$, we know
\begin{align}
    \boxx_{\bH}(\bq',c\varepsilon/2) \subset R_{\varepsilon/2}(\bq') \,.
\end{align}
But these are boxes in the same coordinate system as $\boxx_{\bH}(\bq,C\varepsilon)$, whose size along dimension $i \in [d]$ is $2 (c\varepsilon/2)^{|I_i|}$; this means it is a $(c/(2C))^{|I_i|}$ fraction of the size of $\boxx_{\bH}(\bq,\varepsilon)$ along dimension $i$, which is notably constant with regard to $\varepsilon$. Thus, we can use
\begin{align}
    b = \prod_{i=1}^d \lceil (2C/c)^{|I_i|} \rceil
\end{align}
such boxes to cover $\boxx_{\bH}(\bq,C\varepsilon)$. Call their centers $\bq_1,\dots,\bq_b$. Then we have
\begin{align}
    R_\varepsilon(\bq) \subset \boxx_{\bH}(\bq,C\varepsilon) \subseteq \bigcup_{j=1}^b \boxx_{\bH}(\bq_j,c\varepsilon/2) \subseteq \bigcup_{j=1}^b R_{\varepsilon/2}(\bq_j)
\end{align}
and we are done.
\end{proof}


We then consider \Cref{lem:starting-coverability}, which asserts that if \Cref{meta:regularity} holds for symmetric $\Pi$, then \Cref{assm:starting-coverability} follows. 


\begin{proof} [Proof of \Cref{lem:starting-coverability}]
    For simplicity we assume that $\projinv{\wspacef}$ is itself compact (otherwise we take a compact superset, and if $P(1/\varepsilon)$ configuration $\varepsilon$-reachable sets can cover it then $\projinv{\wspacef}$ is trivially covered by them too).
    %
    
    By the Uniform Ball-Box Theorem, there exists constants $0 < c < C$ and $0 < \varepsilon_0$ such that any $\bq$ has an adapted frame $\bH = (h_{I_1}, \dots, h_{I_d})$ such that for any $0 < \varepsilon \leq \varepsilon_0$ and $\bq \in \projinv{\wspacef}$,
    \begin{align}
        \boxx_{\bH}(\bq,c\varepsilon) \subset R_\varepsilon(\bq) \subset \boxx_{\bH}(\bq,C\varepsilon) \,.
    \end{align}
    
    We let $\varepsilon^*(\bq)$ denote any sufficiently small radius of the reachable set so that $R_{\varepsilon^*(\bq)}(\bq)$ is contained in the neighborhood of $\bq$ on which $\bH$ is also an adapted frame at all other $\bq'$. Without loss of generality we can assume $\varepsilon^*(\bq) \leq \varepsilon_0$. We then consider the open version of $R_{\varepsilon^*(\bq)}(\bq)$, which is an open set containing $\bq$ (since control-affine dynamics satisfying Chow's Condition induce reachable sets around $\bq$ containing a neighborhood of $\bq$). Thus (with the open reachable sets)
    \begin{align}
        \projinv{\wspacef} \subseteq \bigcup_{\bq \in \projinv{\wspacef}} R_{\varepsilon^*(\bq)}(\bq)
    \end{align}
    and since $\projinv{\wspacef}$ is compact, there is some finite subcover. Let $S$ be the number of sets in this finite subcover, and let $\bq_1, \dots, \bq_S$ be the configurations at their centers, so that
    \begin{align}
        \projinv{\wspacef} \subseteq \bigcup_{i=1}^S R_{\varepsilon^*(\bq_i)}(\bq_i)
    \end{align}
     By the proof of \Cref{lem:b-coverability} above, we know that for any $\bq' \in R_{\varepsilon^*(\bq_i)}(\bq_i)$ and any $\varepsilon < \varepsilon^*(\bq_i)$, the set $R_\varepsilon(\bq')$ can be covered by $b$ configuration $(\varepsilon/2)$-reachable sets, which can each in turn be covered by $b$ configuration $(\varepsilon/4)$-reachable sets, and so forth. Thus, for any integer $k \geq 1$, $R_\varepsilon(\bq')$ can be covered by $b^k$ configuration $(2^{-k}\varepsilon)$-reachable sets. In particular, $R_{\varepsilon^*(\bq_i)}(\bq_i)$ can be covered by $b^k$ configuration $(2^{-k}\varepsilon^*(\bq_i))$-reachable sets.

    Let $\varepsilon^*_{\min} \defeq \min_i \varepsilon^*(\bq_i)$ and $\varepsilon^*_{\max} \defeq \max_i \varepsilon^*(\bq_i)$. Then given $0 < \varepsilon \leq \varepsilon^*_{\min}$, how many configuration $\varepsilon$-reachable sets are needed to cover $\projinv{\wspacef}$? We let
    \begin{align}
        k_\varepsilon \defeq \min(k : \varepsilon \leq 2^{-k} \varepsilon^*_{\max}) = \lceil \log_2(\varepsilon^*_{\max}/\varepsilon) \rceil \leq \log_2(\varepsilon^*_{\max}/\varepsilon) + 1
    \end{align}
    This also implies that
    \begin{align}
        \varepsilon \geq 2^{-k_\varepsilon} \varepsilon^*_{\max}
    \end{align}
    Then each $R_{\varepsilon^*(\bq_i)}(\bq_i)$ can be covered by $b^{k_\varepsilon}$ configuration $2^{-k_\varepsilon} \varepsilon^*_{\max}$-reachable sets, which (using the same anchors) means each $R_{\varepsilon^*(\bq_i)}(\bq_i)$ can be covered by $b^{k_\varepsilon}$ configuration $\varepsilon$-reachable sets. But we can bound
    \begin{align}
        b^{k_\varepsilon} \leq b^{\log_2(\varepsilon^*_{\max}/\varepsilon) + 1}
        = b (\varepsilon^*_{\max}/\varepsilon)^{\log_2 b} = c^* \varepsilon^{-\log_2 b}
    \end{align}
    where $c^* = b (\varepsilon^*_{\max})^{\log_2 b}$, which is constant with regards to $\varepsilon$. Noting that we have to do this with all $S$ boxes $R_{\varepsilon^*(\bq_i)}(\bq_i)$ (where $S$ also does not depend on $\varepsilon$), we then get that $\projinv{\wspacef}$ can be covered with $\leq S c^* \varepsilon^{-\log_2 b}$ configuration $\varepsilon$-reachable sets, which is a polynomial number in $1/\varepsilon$, so we are done.
\end{proof}

\subsection{Hierarchical Cell Structures under equiregularity}

In this appendix we show how, under \Cref{meta:regularity} (and a few additional assumptions concerning the properties of the function $\proj{\cdot}$), we can guarantee the existence of a Hierarchical Cell Structure (HCS) about any given $x \in \wspacef$ with $s = 2$, i.e. we prove \Cref{lem:existence-of-hcs}. We use the notation from the first part of this appendix, including letting $d = \dim(\cspace)$; and let $d' = \dim(\wspace) \leq d$.

We want to define equivalent notions to the Lie algebra but for the projection of the dynamics onto $\wspace$; we say that $\bq \in \cspace$ is \emph{projection-regular} if the projections of the Lie bracket vector fields to $\wspace$ at $\bq$ (or rather the projections $\proj{h_I(\bq)}$ of $h_I(\bq) \in T_\bq \cspace$ to $T_x \wspace$ where $x = \proj{\bq}$) behave smoothly. In this case, $\gamma$ is consistent within a neighborhood around $\bq$. We assume:
\begin{assumption}
    For any $x \in \wspacef$ and any $\delta > 0$, there is a projection-regular $\bq$ such that $\proj{\bq} = x$ and $g(\bq) \geq (1-\delta) g(x)$.
\end{assumption}

For convenience, we define the \emph{Local Hierarchical Cell Structure} (LHCS), which is a modified version of a HCS:

\begin{definition} \label{def:hcs-config}
    A $\zeta$-regular \emph{Local Hierarchical Cell Structure} $\lhcs(\bq_0,\varepsilon_0,\alpha,s,\zeta)$ rooted at $\bq_0 \in \cspace$ with radius $\varepsilon_0 > 0$, efficiency parameter $\alpha \leq 1$, and scaling parameter $s$ is a structure consisting of the following elements:
    \begin{itemize}
        \item A Jordan-measurable \emph{cell} $S(\bq_0,\varepsilon_0) \subseteq \wspace$ such that
        \begin{align}
            S(\bq_0,\varepsilon_0) \subseteq \bar{R}_{ \varepsilon_0}(\bq_0) \text{ and } \vol_{\wspace} (S(\bq_0,\varepsilon_0)) \geq \alpha \check{g}^{(\zeta)}(\bq_0) \varepsilon_0^\gamma
        \end{align}
        where $\check{g}^{(\zeta)}$ is the lower $\zeta$-regularized approximation of $g$. 
        \item A set of $s^\gamma$ disjoint $\zeta$-regular hierarchical cell structures with radius $\varepsilon_0/s$, efficiency parameter $\alpha$, scaling parameter $s$, and rooted at $\bq_1, \dots, \bq_{s^\gamma}$ such that
        \begin{align}
            S(\bq_0,\varepsilon_0) \subseteq \bigcup_{j=1}^{s^\gamma} S(\bq_j,\varepsilon_0/s) ~~~\text{ and }~~~ \bq_j \in \bar{R}_{ \varepsilon_0}(\bq_0) \text{ for all } j
        \end{align}
    \end{itemize}
\end{definition}

The difference between \Cref{def:hcs-config} and \Cref{def:hcs} is that the volume of the cell is only expected to be a certain size relative to $\check{g}^{(\zeta)}(\bq_0) \varepsilon_0^\gamma$ (which bounds the volume of the workspace reachable set at $\bq_0$ for small $\varepsilon_0$) as opposed to $\check{g}^{(\zeta)}(\proj{\bq_0}) \varepsilon_0^\gamma$; that is, the cell is only expected to be big relative to the workspace reachable set from the anchor $\bq_0$ rather than relative to the largest such cell around $x_0 = \proj{\bq_0}$.

Note that the condition of having a LHCS around every $\bq_0 \in \cspace$ is actually \emph{stronger} than having a HCS around every $x_0 \in \wspacef$; this is because by the definition $g(x) = \sup_{\bq : \proj{\bq} = x} g(\bq)$ we know that we can find $\bq$ such that $\proj{\bq} = x_0$ such that $g(\bq)$ is arbitrarily close to $g(x_0)$, and then just use that as our $\bq_0$ (taking the approximation $\check{\cdot}^{(\zeta)}$ doesn't affect this as $\zeta \to 0$ except at a closed set of measure $0$, see \Cref{appx:appx-everywhere}).

We want to use \Cref{meta:regularity} with the Uniform Ball-Box Theorem as in \Cref{appx:meta-assumption}, in order to create a coordinate map in a neighborhood around $\bq_0$ in which reachable sets can be inscribed with a rectangle of given dimensions (and circumscribed by a rectangle whose dimensions are fixed multiples of the inscribed rectangle). While in \Cref{lem:b-coverability} the focus was on covering the \emph{circumscribed} rectangle of a configuration $\varepsilon$-reachable set $R_\varepsilon(\bq)$ with a set number of \emph{inscribed} rectangles of configuration $(\varepsilon/2)$-reachable sets $R_{\varepsilon/2}(\bq_1), \dots, R_{\varepsilon/2}(\bq_b)$ (thus proving that $R_\varepsilon(\bq) \subseteq \bigcup_{j=1}^b R_{\varepsilon/2}(\bq_j)$), in our case we will use inscribed rectangles as our cells; thus, we will want to cover inscribed rectangles of reachable sets (cells) with the inscribed rectangles of smaller reachable sets (sub-cells).

One additional wrinkle is that in \Cref{lem:b-coverability} we needed to perform this covering for configuration-reachable sets; however, HCS's and LHCS's involve cells in the workspace. Hence, we will need to carefully consider the projection function $\proj{\cdot}$. Additionally, it was sufficient in \Cref{lem:b-coverability} to achieve the covering with any fixed number $b$ of smaller rectangles, for HCS's and LHCS's it is required to achieve the covering with exactly $2^\gamma$ smaller cells (since $s = 2$ for what follows).

We consider first a stylized case with $\cspace = \wspace$ (for instance, where the Dubins car needs to visit target points where each target point has a random direction in which it needs to be visited), in which case the projection function is the identity. Let us fix some $\bq_0 \in \cspace$ and $\bH = (h_{I_1}, \dots, h_{I_d})$ be an adapted frame at $\bq$ and scale $\varepsilon$ where $I_1$ are elements of the Lie algebra; we let $r_i = |I_i|$ and $r = \sum_{i=1}^d r_i$; since $\cspace = \wspace$, in this case $\gamma = r$ (since the volume of this box is proportional to $\varepsilon^r$ as $\varepsilon \to 0$). Then $\boxx_{\bH}(\bq_0,\varepsilon)$ is a box in the coordinate map produced by the adapted frame with sides parallel to the axes and length $2\varepsilon^{r_i}$ along dimension $i$ in the coordinate map. Then there is a neighborhood $\cR$ of $\bq_0$ and some $\varepsilon_0^* > 0$ (we add the $\cdot^*$ to distinguish it from the $\varepsilon_0$ in the HCS and LHCS definitions) and constants $0 < c < C$ such that for any $\bq$ in a neighborhood of $\bq_0$ and any $\varepsilon < \varepsilon_0^*$
\begin{align}
    \boxx_{\bH}(\bq, c\varepsilon) \subset R_\varepsilon(\bq) \subset \boxx_{\bH}(\bq, C\varepsilon)
\end{align}
(see the Uniform Ball-Box Theorem).

Given $\boxx_{\bH}(\bq,c\varepsilon)$, we note that we can cover it with $2^r$ instances of $\boxx_{\bH}(\bq', c(\varepsilon/2))$, since these are smaller boxes with length $2(c\varepsilon/2)^{r_i} = 2^{-r_i} \cdot 2 (c\varepsilon)^{r_i}$, hence it has length exactly $2^{-r_i}$ of the length of $\boxx_{\bH}(\bq,c\varepsilon)$ on dimension $i$. Since $r_i = |I_i|$ is an integer (the length of the Lie bracket $I_i$), this means $2^{r_i}$ is also an integer and hence we can divide the length of $\boxx_{\bH}(\bq,c\varepsilon)$ into $2^{r_i}$ pieces of exactly length $2^{-r_i}$. Repeating this for every $i = 1, \dots, d$ yields a division of $\boxx_{\bH}(\bq,c\varepsilon)$ into exactly $\prod_{i=1}^d 2^{r_i} = 2^r = 2^\gamma$ boxes of the form $\boxx_{\bH}(\bq',c(\varepsilon/2))$. Thus, using $S(\bq,\varepsilon) = \boxx_{\bH}(\bq,c\varepsilon)$ as our cells, we have our Local Hierarchical Cell Structure in this case; because of the circumscribing box $\boxx_{\bH}(\bq,C\varepsilon)$, we know that there is a sufficiently small $\varepsilon^* > 0$ such that if the HCS has scale $\leq \varepsilon^*$, the efficiency parameter $\alpha$ (the ratio of the cell volume to the volume of the reachable set it's inscribed in) satisfies $\alpha \geq \frac{1}{2}(c/C)^r$.

\begin{remark}
    The volume of the `boxes' on the manifold is not the same as the volumes of their representations on the coordinate map; for one thing, the vectors in $\bH$ may have different magnitudes, and the curvature of the manifold $\cspace$ will also have an effect. However, the \emph{ratio} of the volume of $\boxx_{\bH}(\bq',c\varepsilon)$ to the volume of $\boxx_{\bH}(\bq',C\varepsilon)$, as $\varepsilon \to 0$, approaches $(c/C)^r$. Thus, there is a $\varepsilon^* > 0$ such that for all $\varepsilon \leq \varepsilon^*$ and $\bq'$ in the neighborhood of $\bq$
    \begin{align}
        \frac{\vol_\cspace(\boxx_{\bH}(\bq,c\varepsilon))}{\vol_\cspace(\boxx_{\bH}(\bq,C\varepsilon))} \geq \frac{1}{2} (c/C)^r
    \end{align}
\end{remark}

However, we also want to extend this to cases where $\wspace \neq \cspace$; although the same cell structure will achieve covering (just project all cells down to the workspace), the issue is that when $\wspace \neq \cspace$ we get $\gamma < r$ and hence covering with $2^r$ sub-cells is too many. Instead we will choose a size-$2^\gamma$ subset of the configuration space sub-cells whose projections onto $\wspace$ will cover the projection of $\boxx_{\bH}(\bq,c\varepsilon)$ (up to an approximation factor which decreases as $\varepsilon \to 0$).

As discussed in \Cref{sec:prob-formulation}, we assume that $\proj{\cdot}$ locally behaves as a projection from a ($d$-dimensional) Euclidean space to a ($d'$-dimensional) subspace. Let us fix some $\bq \in \cspace$, and let $x = \proj{\bq}$. We can now treat a sufficiently small neighborhood $\cV$ of $\bq$ and its projection $\proj{\cV}$ onto $\wspace$ as both linear (using $\bH$ as our basis in $\cspace$). We note for any $\bq' \in \cV$ and any sufficiently small $\varepsilon$, by the Uniform Ball-Box Theorem we can inscribe $\boxx_{\bH}(\bq', c\varepsilon)$ in $R_\varepsilon(\bq')$; as before, we can divide this into $2^r$ sub-boxes of the form $\boxx_{\bH}(\bq'', c(\varepsilon/2))$. We now consider the projection of these boxes onto $\wspace$: we will show that an appropriately-chosen set of $2^\gamma$ of the $2^r$ sub-boxes (with a tiny bit of scaling up, which will diminish as $\varepsilon \to 0$) will cover the projection $\proj{\boxx_{\bH}(\bq',c\varepsilon)}$.

A vector field $h$ on $\cspace$ produces for any $\bq \in \cspace$ a tangent vector $h(\bq) \in T_\bq \cspace$. Taking the projection to $\wspace$ (and letting $x = \proj{\bq}$), we get a tangent vector on $\proj{h(\bq)} \in T_x \wspace$ (moving along $h(\bq)$ from $\bq \in \cspace$ projects to a movement along $\proj{h(\bq)}$ from $x = \proj{\bq} \in \wspace$). We note that $\bH(\bq) = \{h_{I_1}(\bq), \dots, h_{I_d}(\bq)\}$ is a basis of $T_\bq \cspace$; therefore 
\begin{align}
    \proj{\bH(\bq)} = \{\proj{h_{I_1}(\bq)}, \dots, \proj{h_{I_d}(\bq)}\}
\end{align}
spans $T_x \wspace$. Let us denote $\bar{h}_i(\bq) \defeq \proj{h_{I_i}(\bq)}$ and denote
\begin{align}
    \bar{\bH}(\bq) \defeq \proj{\bH(\bq)} = \{\bar{h}_1(\bq), \dots, \bar{h}_d(\bq)\}\,.
\end{align}
For any $A \subseteq [d]$, we let $\bH_A \defeq \{h_{I_i} : i \in A\}$ (and $\bH_A(\bq) \defeq \{h_{I_i}(\bq) : i \in A\}$) and likewise $\bar{\bH}_A \defeq \{\bar{h}_i(\bq) : i \in A\}$.

\begin{remark}
    It's important to remember that, despite the notation, $\bar{h}_i(\bq)$ is a tangent vector of $T_x \wspace$ at $x = \proj{\bq}$; the $\bq$ is there to tell us which tangent vector. Likewise, $\bar{\bH}(\bq)$ is a collection of $d$ tangent vectors in $T_x \wspace$.
\end{remark}

We then define for any $A \subset [d]$ such that $|A| = d'$ the following for any $\bq' \in \cV$ and $x' = \proj{\bq'}$: 
\begin{align}
    \boxx_{\bH_A}(\bq',\varepsilon) = \{\exp(z_1 h_{I_1}) \circ \dots \circ \exp(z_d h_{I_d})(\bq') : |z_i| \leq \varepsilon^{|I_i|} ~ \, \forall i \in A \text{ and } z_i = 0 ~\, \forall i \not \in A \}
\end{align} 
This is like $\boxx_{\bH}(\bq',\varepsilon)$ but we may only use vector fields in $\bH_A$; note that $\boxx_{\bH_A}(\bq',\varepsilon) \subseteq \boxx_{\bH}(\bq',\varepsilon)$.
Let us also denote $r_A \defeq \sum_{i \in A} r_i$. Then, when $\varepsilon$ is sufficiently small, if $\bar{\bH}_A$ is linearly independent,
\begin{align}
    \vol_\wspace(\proj{\boxx_{\bH_A}(\bq',\varepsilon)}) =  \Theta(\varepsilon^{r_A}) \,.
\end{align}

We first consider a stylized case where for any size-$d'$ subset $A \subseteq [d]$, $\bar{\bH}_A(\bq)$ is linearly independent. Since our vector fields and projection function are smooth, this property will hold in some neighborhood around $\bq$; we will assume that it holds in the neighborhood $\cV$ discussed above (otherwise, just take the intersection of $\cV$ with the neighborhood for which this holds).

\begin{remark} \label{rmk:degenerate-subset}
    Note that this stylized case doesn't always apply: for example, the Reeds-Shepp car on $\bbR^2$ does not satisfy it, since the vector field $h_2 = [0;0;1]$ (note that $h_1 = [\cos(\theta),\sin(\theta),0]$) which is controlled to steer the vehicle projects to nothing on $\wspace = \bbR^2$ (i.e. it is not linearly independent of anything else). Traveling along this vector field rotates the vehicle, which does not affect its location in $\wspace = \bbR^2$. 
\end{remark}

In this case, without loss of generality, let $r_1 \leq \dots \leq r_d$. Then we let $A^* = [d']$, which clearly satisfies $A^* = \argmin_A r_A$. We first claim that $r_{A^*} = \gamma$.

For any $A \subseteq [d]$ such that $|A| = d'$, consider a sequence $\ba = (a_i : i \in [d]\backslash A)$ where $a_i \in \{-1, 1\}$ for all $i \in [d]\backslash A$; that is, we assign $-1$ or $1$ for every index \emph{not} in $A$. Then we define:
\begin{align}
    \boxx_{\bH_A^{(\ba)}}(\bq',\varepsilon) &= 
    \\ \{\exp(z_1 h_{I_1}) &\circ \dots \circ \exp(z_d h_{I_d})(\bq') : |z_i| \leq \varepsilon^{|I_i|} ~ \, \forall i \in A, z_i = a_i\varepsilon^{|I_i|} ~\, \forall i \not \in A \}
\end{align}
This denotes a $d'$-dimensional facet of $\boxx_{\bH}(\bq',\varepsilon)$ whose dimensions run along $h_{I_i}$ for $i \in A$; the sequence $\ba$ denotes which facet (i.e. for the remaining dimensions, which side of the box is it on). Note that there are a fixed number $2^{d-d'}$ of these facets for each $A$, and a fixed number ${d \choose d'}$ of size-$d'$ sets $A$, hence a total of $2^{d-d'} {d \choose d'}$ such facets. We now note that
\begin{align} \label{eq:facets-cover-box}
    \proj{\boxx_{\bH}(\bq',\varepsilon)} = \bigcup_{A \subseteq [d] : |A| = d'} \bigcup_{\ba \in \{-1,1\}^{[d]\backslash A}} \proj{\boxx_{\bH_A^{(\ba)}}(\bq',\varepsilon)}
\end{align}
which is just saying that the projection of $\boxx_{\bH}(\bq',\varepsilon)$ onto $\wspace$ is the same as the union of the projections of all its dimension-$d'$ facets onto $\wspace$. This means
\begin{align} \label{eq:facets-cover-box-volume}
    \vol_{\wspace}(\proj{\boxx_{\bH}(\bq',\varepsilon)}) &\leq \sum_{A \subseteq [d] : |A| = d'} \sum_{\ba \in \{-1,1\}^{[d]\backslash A}} \vol_{\wspace}(\proj{\boxx_{\bH_A^{(\ba)}}(\bq',\varepsilon)})
    \\ &= \Theta(\varepsilon^{r_{A^*}})
\end{align}
since it is a sum of (a bounded number of) terms of order $\varepsilon^{r_A}$ for various $A$, and $r_{A^*} = \min_A r_A$. Furthermore, since $\boxx_{\bH_{A^*}}(\bq',\varepsilon) \subseteq \boxx_{\bH}(\bq',\varepsilon) $, we have
\begin{align}
    \Theta(\varepsilon^{r_{A^*}}) = \vol_\wspace(\proj{\boxx_{\bH_{A^*}}(\bq',\varepsilon)}) \leq \vol_\wspace(\proj{\boxx_{\bH}(\bq',\varepsilon)})
\end{align}
and hence we have
\begin{align}
    \vol_\wspace(\proj{\boxx_{\bH}(\bq',\varepsilon)}) = \Theta(\varepsilon^{r_{A^*}})
\end{align}
But since the small-time constraint factor was defined as the value $\gamma$ satisfying
\begin{align} \label{eq:facets-cover-box-finish}
    \vol_\wspace(\proj{\boxx_{\bH}(\bq',\varepsilon)}) = \Theta(\varepsilon^\gamma)
\end{align}
so $\gamma = r_{A^*}$. This is not surprising as both reflect the volume of $\boxx_\bH(\bq',\varepsilon)$ projected on $\wspace$.

We now take a cell in our Local Hierarchical Cell Structure about $\bq'$ at scale $\varepsilon$ to be $S(\bq', \varepsilon) = \proj{\boxx_{\bH_{A^*}}(\bq',\varepsilon)}$. We check that it is indeed a LHCS (with scaling factor $2$). We can subdivide $\boxx_{\bH_{A^*}}(\bq',\varepsilon)$ into $2^{r_{A^*}} = 2^\gamma$ sub-boxes $\boxx_{\bH_{A^*}}(\bq'',\varepsilon/2)$ in the same way as the Uniform Ball-Box Theorem does with $\boxx_{\bH}(\bq',\varepsilon)$; the projections of these sub-boxes then become the sub-cells. One thing is that it may not have a good efficiency parameter $\alpha$ and it may be possible to do much better; but we are only proving the existence of a LHCS here.

What happens when $\bar{\bH}_A(\bq)$ is not linearly independent for all $A$? It is still possible that an $A'$ such that $\bar{\bH}_{A'}(\bq)$ is not independent can still have
\begin{align}
    \vol_\wspace(\proj{\boxx_{\bH_{A'}}(\bq',\varepsilon)}) > 0
\end{align}
though the volume must be $o(\varepsilon^{r_{A'}})$. However, if it has volume, it is only through a Lie bracket interaction; thus, it has volume $\Theta(\varepsilon^{r'_{A'}})$ where $r'_{A'} > r_{A'}$ is an integer, and
\begin{align}
    \min(r_A : A \subseteq [d]  \text{ s.t. } |A| = d' \text{ and } \bar{\bH}_{A}(\bq') \text{ is linearly independent}) \leq r'_{A'} \,.
\end{align}
Therefore, we may choose 
\begin{align}
    A^* \defeq \argmin(r_A : A \subseteq [d]  \text{ s.t. } |A| = d' \text{ and } \bar{\bH}_{A}(\bq') \text{ is linearly independent})
\end{align}
Then the arguments from above (particularly equations \eqref{eq:facets-cover-box} through \eqref{eq:facets-cover-box-finish}) still hold, showing that $r_{A^*} = \gamma$. We can then use $S(\bq', \varepsilon) = \proj{\boxx_{\bH_{A^*}}(\bq',\varepsilon)}$ as our cell, as before.

Then, almost everywhere in $\wspacef$, for any $\delta > 0$ we can choose $\bq$ such that $\proj{\bq} = x$ and $\check{g}^{(\zeta)}(\bq) \geq (1-\delta) \check{g}^{(\zeta)}(x) \geq (1-\delta) g_{\min}$; the Local HCS around $\bq$ is then a HCS around $x$. There exists a efficiency parameter $\alpha > 0$ which, for sufficiently small $\varepsilon$, holds for all the sub-HCS's because $\check{g}^{(\zeta)}(\bq)$ is bounded away from $0$ and is Lipschitz continuous, i.e. on small scales it can be treated as constant.



\section{Existence of accurate HCS covers} \label{appx:hcs-subdivision}

In this appendix we show formally the existence of $\rho$-accurate HCS covers, i.e. we prove \Cref{lem:hcs-cover-01} and \Cref{lem:hcs-cover-02} from \Cref{assm:hcs}. Recall that \Cref{assm:hcs} states that every point in $\wspacef$ is contained inside some HCS; \Cref{lem:hcs-cover-01} states that we can therefore cover $\wspacef$ with a finite number of HCS's of some scale $\varepsilon_0$; and \Cref{lem:hcs-cover-02} states that we can do so at arbitrarily small scales $\varepsilon_0$ with arbitrarily small accuracy parameter $\rho > 0$.

\begin{proof}[Proof of \Cref{lem:hcs-cover-01}]
    By \Cref{assm:hcs} each $x \in \wspacef$ is contained in the interior of some HCS $S_x$. Thus, letting $S_x^{\mathrm{int}}$ be the interior of $S_x$, we know that $\wspacef \subseteq \bigcup_x S_x^{\mathrm{int}}$. But by \Cref{assm:density-fn}, $\wspacef$ is compact, and hence there is a finite subset of these $S_x$'s whose interiors cover $\wspacef$.
\end{proof}

\begin{proof}[Proof of \Cref{lem:hcs-cover-02}]
    We use the fact that by definition a $\varepsilon_0$-scale Hierarchical Cell Structure is composed of $s^\gamma$ smaller $\varepsilon_0/s$-scale (sub)-Hierarchical Cell Structures, which are then each divisible again and so forth. We also use the fact that within each HCS, the sub-HCS's are all disjoint, and that any scale-$\varepsilon$ HCS is contained in some ball in $\wspace$ with radius $c_\Pi \varepsilon$.

    By \Cref{lem:hcs-cover-01} we have a HCS cover $S^{(\mathrm{start})}_1, \dots, S^{(\mathrm{start})}_{m^{(\mathrm{start})}}$ at some scale $\varepsilon_0^{(\mathrm{start})}$ which are $\zeta$-regular with efficiency and scaling parameters $\alpha,s$; it may have a very large accuracy parameter $\rho > 0$ but since it is a cover we know that $\wspacef \subseteq \bigcup_{j=1}^{m^{(\mathrm{start})}} S^{(\mathrm{start})}_j$. We then will use this to create an alternative covering $S_1, \dots, S_m$ (at a smaller scale $\varepsilon_0$) which in addition to being $\zeta$-regular is also $\rho$-accurate. 
    For any Jordan-measurable subset $A \subseteq \wspacef$, let $\mathrm{Int}(A)$ denote its interior and
    \begin{align}
        \mathrm{Int}_\eta(A) \defeq \{x \in A : d_\wspace(x,x') \geq \eta \text{ for all } x' \not \in A\}
    \end{align}
    i.e. the $\eta$-interior of $A$ (points at least $\eta$ distance away from anything outside of $A$). We then note the following (all sets involved are Jordan-measurable):
    \begin{enumerate}[a.]
        \item $\vol_\wspace(A) = \vol_\wspace(\mathrm{Int}(A)) =  \lim_{\eta \to 0} \vol_\wspace(\mathrm{Int}_\eta(A))$.
        \item For any $A_1, \dots, A_k \subseteq \wspace$, letting $A = \bigcap_{j=1}^k A_j$, we have $\mathrm{Int}_\eta(A) = \bigcap_{j=1}^k \mathrm{Int}_\eta(A_j)$ and $\mathrm{Int}(A) = \bigcap_{j=1}^k \mathrm{Int}(A_j)$. Since intersections of finitely many Jordan-measurable sets are Jordan-measurable, we also have $\vol_\wspace(A) = \vol_\wspace(\mathrm{Int}(A))$.
    \end{enumerate}


    We now subdivide each $S^{(\mathrm{start})}_j$ them $k$ times, for any integer $k$ to get a cover of $m \defeq m^{\mathrm{start}} s^{\gamma k}$ HCS's at scale $\varepsilon_0 \defeq \varepsilon_0^{(\mathrm{start})} s^{-k}$. Let these be denoted $S_1, \dots, S_m$; let $S^{(j)}_1, \dots, S^{(j)}_{s^{\gamma k}}$ denote the HCS's which were created by subdividing $S^{(\mathrm{start})}_j$. For each $S^{(j)}_i$, we then remove it if there is some $j' < j$ such that $S^{(j)}_i \subseteq S^{(\mathrm{start})}_{j'}$, i.e. if it is completely contained within a different HCS from our starting cover with a lower index. This is a simple way to remove redundant HCS's, and preserves the covering property; to see this, we can perform this action starting with the descendants of $S^{(\mathrm{start})}_{m^{(\mathrm{start})}}$, then the descendants of $S^{(\mathrm{start})}_{m^{(\mathrm{start})}-1}$ and so forth. We now claim that as long as we set $k$ sufficiently large, we can reduce the redundancy (that is, the region contained within multiple HCS's) to an arbitrarily small set. Let
    \begin{align}
        S_k(j,j') \defeq \{x : x \in S^{(j)}_i \cap S^{(j')}_{i'} \text{ for some } i, i'\} 
    \end{align}
    i.e. the region in which descendants of $S^{(\mathrm{start})}_j$ and $S^{(\mathrm{start})}_{j'}$ overlap, and its volume, given $k$ subdivisions (after removing redundant descendants as per the above steps). Without loss of generality let $j' < j$. Note that $S_0(j,j') = S^{(\mathrm{start})}_j \cap S^{(\mathrm{start})}_{j'}$, and that
    \begin{align}
        S_{k}(j,j') \subseteq S_{k'}(j,j') \text{ for all } k > k' \,.
    \end{align}
    We now let 
    \begin{align}
        \eta \defeq 2 c_\Pi \varepsilon_0 = 2 c_\Pi \varepsilon_0^{(\mathrm{start})} s^{-k}
    \end{align}
    and claim that 
    \begin{align} \label{eq:interior-exclusion}
        S_k(j,j') \cap \mathrm{Int}_\eta(S_0(j,j')) = \emptyset
    \end{align}
    that is, that deleting redundant cells will progressively `hollow out' the interior of $S_0(j,j')$. This happens because for the root $\bq$ of cell $S^{(j)}_i$ at $x$ (i.e. $x = \proj{\bq}$)
    \begin{align}
        S^{(j)}_i \subseteq \bar{R}_{\varepsilon_0}(\bq) \subseteq \cB_{c_\Pi \varepsilon_0}(x)
    \end{align}
    This means that any cell $S^{(j)}_i$ rooted at $x$ is necessarily deleted if $x \in \mathrm{Int}_{c_\Pi \varepsilon_0}(S_0(j,j'))$; and this additionally means that if $S^{(j)}_i$ \emph{contains} $x$ where $x \in \mathrm{Int}_{2 c_\Pi \varepsilon_0}(S_0(j,j')) = \mathrm{Int}_{\eta}(S_0(j,j'))$, then $S^{(j)}_i$ is removed as well. Thus we get \eqref{eq:interior-exclusion}, which further implies
    \begin{align}
        S_k(j,j') \subseteq S_0(j,j') \backslash \mathrm{Int}_\eta(S_0(j,j')) \,.
    \end{align}
    But since $S_0(j,j') = S^{(\mathrm{start})}_j \cap S^{(\mathrm{start})}_{j'}$, it is Jordan-measurable and hence the volume of the above goes to $0$ as $\eta \to 0$, which happens as $k \to \infty$. But this means if we let the total region of overlaps be
    \begin{align}
        S'_k \defeq \{x \in \wspacef : x \in S^{(j)}_i \cap S^{(j')}_{i'} \text{ for some } i,j,i',j' \text{ where } j \neq j'\}
    \end{align}
    we get that
    \begin{align}
        S'_k &= \bigcup_{j \neq j'} S_k(j,j')
        \\ \implies \vol_\wspace(S'_k) &\leq \sum_{j \neq j'} \vol_\wspace(S_k(j,j'))
        \\ \implies \lim_{k \to \infty} \vol_\wspace(S'_k) &\leq \lim_{k \to \infty} \sum_{j \neq j'} \vol_\wspace(S_k(j,j')) = 0
    \end{align}
    Thus for any $\rho > 0$, we can set $k$ sufficiently large so that 
    \begin{align}
        \bbP_{X \sim f} [X \in S'_k] \leq \rho
    \end{align}
    which means that we have a $\rho$-accurate cover.
    

    Finally, once we have it at scale $\varepsilon_0$, we can scale it down by as many factors of $s$ as we like, and in particular until the radius is smaller than the given $\varepsilon^*_0$.
\end{proof}

\section{Approximately Everywhere} \label{appx:appx-everywhere}

We now prove \Cref{prop:appx-everywhere}, which states that our main results (\Cref{prop:dstsp-lower-bound} and \Cref{prop:dstsp-upper-bound}) hold even when our assumptions are true only approximately everywhere (\Cref{def:appx-everywhere}) rather than everywhere. 

\begin{proof}[Proof of \Cref{prop:appx-everywhere}]
    First, we note that we have a finite number $k$ of assumptions, which we can label as $1, \dots, k$. Then for $\eta > 0$, let $\wspace^{(i)}_\eta$ denote the $\eta$-approximator of $\wspacef$; we then denote $\wspace_\eta \defeq \bigcap_{i=1}^k \wspace^{(i)}_\eta$ and $\wspace'_\eta$ be its $\eta$-interior. We claim that:
    \begin{enumerate} [i.]
        \item All the assumptions hold over $\wspace_\eta$.
        \item $\vol_\wspace(\wspacef \backslash \wspace'_\eta) \leq k \eta$.
    \end{enumerate}
    To show (i), we note that all our assumptions take the form discussed in \Cref{rem:appx-everywhere}, i.e. `for all $\zeta > 0$, there is a $\varepsilon^*_\zeta > 0$ such that for all $0 < \varepsilon \leq \varepsilon^*_\zeta$, a certain condition holds across the set'. We then fix $\zeta > 0$ and we let $\varepsilon^{(i)}_{\zeta,\eta} > 0$ be the value which corresponds to assumption $i$ over $\wspace^{(i)}_\eta$; then we let $\varepsilon^*_{\zeta,\eta} = \min_i \varepsilon^{(i)}_{\zeta,\eta}$. In this case, $\varepsilon < \varepsilon^*_{\zeta,\eta} \implies \varepsilon < \varepsilon^{(i)}_{\zeta,\eta}$ and hence for such $\varepsilon$ the conditions all hold over their respective $\wspace^{(i)}_\eta$, and hence hold over $\wspace_\eta$.

    To show (ii), we first show that $\wspace'_\eta = \bigcap_{i=1}^k \wspace^{(i)'}_\eta$ where $\wspace^{(i)'}_\eta$ is the $\eta$-interior of $\wspace^{(i)}_\eta$. This is because
    \begin{align}
        x \in \wspace'_\eta &\iff \inf_{x' \not \in \wspace_\eta} d_\wspace(x,x') \geq \eta
        \\ &\iff \inf_{i \in [k]} \inf_{x' \not \in \wspace^{(i)}_\eta} d_\wspace(x,x') \geq \eta
        \\ &\iff x \in \wspace^{(i)}_\eta \text{ for all } i \in [k]
        \\ &\iff x \in \bigcap_{i=1}^k \wspace^{(i)'}_\eta
    \end{align}
    Then we can compute:
    \begin{align}
        \vol_\wspace (\wspacef \backslash \wspace'_\eta) &= \vol_\wspace \Big(\wspacef \backslash \bigcap_{i=1}^k \wspace^{(i)'}_\eta \Big)
        \\ &= \vol_\wspace \Big(\bigcup_{i=1}^k (\wspacef \backslash \wspace^{(i)'}_\eta) \Big)
        \\ &\leq \sum_{i=1}^k \vol_\wspace (\wspacef \backslash \wspace^{(i)'}_\eta)
        \\ &\leq k \eta
    \end{align}
    Thus, in particular we know that any point $x \in \wspace'_\eta$ has a $\eta$-radius ball around it in which all the assumptions hold and that  as $\eta \to 0$, the volume of $\wspacef$ that's outside the set $\wspace'_\eta$ goes to $0$, which in turn shows that
    \begin{align}
        \lim_{\eta \to 0} \bbP_{X \sim f}[X \in \wspace'_\eta] = 1 \text{ and }  \lim_{\eta \to 0} \int_{\wspace'_\eta} f(x)^{1-\frac{1}{\gamma}} g(x)^{-\frac{1}{\gamma}} \, dx = \int_{\wspacef} f(x)^{1-\frac{1}{\gamma}} g(x)^{-\frac{1}{\gamma}} \, dx
    \end{align}
    since $f$ represents a continuous probability distribution and $g$ is bounded below.

    We now consider our main theorems, and specifically how they are proven. 
    
    First, we consider \Cref{prop:dstsp-lower-bound} (the very high probability lower bound to the DSTSP). We then fix $\eta > 0$ and note that for any $X_1, \dots, X_n$,
    \begin{align}
        \tsp_\Pi(\{X_1, \dots, X_n\} \cap \wspace'_\eta) \leq \tsp_\Pi(X_1, \dots, X_n)
    \end{align}
    Thus, any very high probability lower bound for the DSTSP where the trajectory only needs to visit target points in $\wspace'_\eta$ also applies to the original problem. Furthermore, as $\lim_{\eta \to 0} \bbP_{X \sim f}[X \in \wspace'_\eta] = 1$, for any $\delta_1 > 0$ we can set $\eta$ sufficiently small so that
    \begin{align}
        \bbP_{X \sim f}[X \not \in \wspace'_\eta] \leq \delta_1/2
    \end{align}
    Then $\{X_1, \dots, X_n\} \cap \wspace'_\eta$ will with very high probability have $\geq (1-\delta_1)n$ targets by the Chernoff bound. 
    
    Thus, we can modify the problem to distribute $(1-\delta_1)n$ targets over $\wspace'_\eta$ with density proportional to $f$; this yields a TSP tour length at most as long (with very high probability) as distributing $n$ targets according to $f$ and keeping only whose which fall in $\wspace'_\eta$, which in turn is at most as long as having to visit all $n$ targets.

    We note that the scaled version of $f$ on $\wspace'_\eta$ is
    \begin{align}
        f|_{\wspace'_\eta}(x) = \begin{cases} f(x)/\bbP_{X \sim f}[X \in \wspace'_\eta] &\text{if } x \in \wspace'_\eta \\ 0 &\text{if } x \not \in \wspace'_\eta \end{cases}
    \end{align}
    and that this value for all $x \in \wspace'_\eta$ can be bounded as
    \begin{align}
        f(x) \leq f|_{\wspace'_\eta}(x) \leq (1-\delta_1/2)^{-1} f(x)
    \end{align}
    which in turn means that
    \begin{align}
        \int_{\wspace'_\eta} f|_{\wspace'_\eta}(x)^{1-\frac{1}{\gamma}} g(x)^{-\frac{1}{\gamma}} \, dx &\leq (1-\delta_1/2)^{-(1-\frac{1}{\gamma})} \int_{\wspace'_\eta} f(x)^{1-\frac{1}{\gamma}} g(x)^{-\frac{1}{\gamma}} \, dx
        \\ &\leq (1-\delta_1/2)^{-(1-\frac{1}{\gamma})} \int_{\wspacef} f(x)^{1-\frac{1}{\gamma}} g(x)^{-\frac{1}{\gamma}} \, dx
    \end{align}

    We can then turn to the Orienteering problem, in particular the Cost-Balanced Orienteering problem \Cref{def:cbo} and the very high probability upper bound \Cref{prop:cbo-more-precise} we derive for it. We can consider this problem on $(1-\delta_1)n$ random target points for approximation factor $\delta_2$. We fix cost lower bound $\zeta$ as in the proof of \Cref{prop:cbo-more-precise} and set
    \begin{align}
        \lambda < \frac{\eta \zeta}{c_\Pi} 
    \end{align}
    where $c_\Pi$ is the speed limit of $\Pi$. Then any cost-$\lambda$ trajectory must have length $\leq \lambda/\zeta < \eta/c_\Pi$, which can cover a distance of at most $\eta$ in the metric on $\wspace$. But since all targets $X_i$ in this problem are in $\wspace'_\eta$, which is the $\eta$-interior of $\wspace_\eta$, any trajectory visiting any target point (which are the only ones we care about) must remain entirely in $\wspace_\eta$, which is exactly where we know our assumptions hold. Therefore, the proof of \Cref{prop:cbo-more-precise} holds and we obtain the bound
    \begin{align}
        \bbP[\cbo_{\Pi}(X_1, X_2, \dots, X_n; \lambda, f) \leq (1 + \delta_2) \beta \lambda \, (1-\delta_1)^\frac{1}{\gamma} n^{\frac{1}{\gamma}}] \geq 1 - e^{-\frac{4}{5} (1+\delta_2) \log(b) \lambda (1\delta_1)^{\frac{1}{\gamma}}n^{\frac{1}{\gamma}}}
    \end{align}
    Finally, we apply this to the TSP lower bound in the same way. In this case, instead of visiting $n$ target points, we need to visit $(1-\delta_1)n$, but the argument is the same. Then, by setting $\delta_1,\delta_2$ small enough we can achieve an approximation factor $\delta$ for the TSP for any $\delta > 0$ with very high probability.
\end{proof}

\section{Good Target Bad Target} \label{appx:good-target-bad-target}

We now claim that even though our HCS cover might not be entirely overlap-free, because it can be arbitrarily overlap-free (see \Cref{lem:hcs-cover-02}) we can derive the main results in the same way as in \Cref{sec:dstsp-ub}. Recall that we showed that the probability that $X \sim f$ falls into multiple HCS's can be made arbitrarily small, i.e. less than any $\rho > 0$. Additionally, in \Cref{lem:prob-of-cells} we made a distinction of cells in a HCS cover between \emph{$\rho_1$-good} and \emph{not $\rho_1$-good} cells, and showed that for any $\rho_1, \rho_2 > 0$ we could find a sufficiently fine scale so that the probability of a target point not falling into a $\rho_1$-good cell is at most $\rho_2$.

This allows us to define two kinds of target points $X_i$: those that fall into a single $\rho_1$-good cell, and those that don't (either by falling into a not-$\rho_1$-good cell, or by falling into multiple cells) which we respectively call \emph{good} and \emph{bad} target points. Each target point has at most a $\rho^* \leq \rho + \rho_2$ chance of being bad, where $\rho^*$ can be made arbitrarily small, and they are independent. We now show that we can simplify to ignore the bad ones. We first fix some HCS cover of $\wspacef$ at scale $\varepsilon^{(\bad)}$ with $m^{(\bad)}$ different HCS's, as in \Cref{lem:hcs-cover-01}; this is allowed to overlap to any amount, but note that it is fixed independent of $\rho^*$.

Let $n^{(\bad)}$ be the number of bad target points, and $n^{(\bad)}_j$ be the number of bad target points in the $j$th cell of the fixed HCS cover with $m^{(\bad)}$ cells at scale $\varepsilon^{(\bad)}$. Note by \Cref{prop::hcp-upper-bound} and the concavity of $x^{1-\frac{1}{\gamma}}$ that this means that
\begin{align}
    \sum_{j=1}^{m^{(\bad)}} \hcp^*(n^{(\bad)}; \hat{b},s) \leq \sum_{j=1}^{m^{(\bad)}} 6 s (n^{(\bad)}/m^{(\bad)})^{1-\frac{1}{\gamma}} = 6 s (m^{(\bad)})^{\frac{1}{\gamma}} (n^{(\bad)})^{1-\frac{1}{\gamma}} 
\end{align}
Thus, letting $\bq_1^{(\bad)}, \dots, \bq_{m^{(\bad)}}^{(\bad)}$ be the roots of this HCS cover and letting
\begin{align}
    C^{(\bad)} \defeq \tsp_\Pi(\bq_1^{(\bad)}, \dots, \bq_{m^{(\bad)}}^{(\bad)}) \,.
\end{align}
As before we don't necessarily need to know the exact value of this; what's important is that it is fixed with respect to $n$.

Thus, we know that
\begin{align}
    \tsp_\Pi(\{X_i : X_i \text{ is bad}\}) \leq C^{(\bad)} + 6 s (m^{(\bad)})^{\frac{1}{\gamma}} \varepsilon^{(\bad)} (n^{(\bad)})^{1-\frac{1}{\gamma}} \,.
\end{align}
This is a guaranteed bound, not a probabilistic one. We can then note that $n^{(\bad)}$ is itself a binomial random variable of $n$ draws of $\leq \rho^*$ probability of success each. Therefore, with very high probability (for any fixed $\rho^* > 0$) we have $n^{(\bad)} \leq 2\rho^* n$, which implies 
\begin{align}
    \tsp_\Pi(\{X_i : X_i \text{ is bad}\}) \leq C^{(\bad)} + 6 s (m^{(\bad)})^{\frac{1}{\gamma}} \varepsilon^{(\bad)} 2^{1-\frac{1}{\gamma}} (\rho^*)^{1-\frac{1}{\gamma}} n^{1-\frac{1}{\gamma}}
\end{align}
By \Cref{lem:prob-of-cells}, for any $\delta^* > 0$, we can then fix $\rho^*$ sufficiently small so that for sufficiently large $n$,
\begin{align}
    C^{(\bad)}/n^{1-\frac{1}{\gamma}} + 6 s (m^{(\bad)})^{\frac{1}{\gamma}} \varepsilon^{(\bad)} 2^{1-\frac{1}{\gamma}} (\rho^*)^{1-\frac{1}{\gamma}} \leq \delta^*
\end{align}
i.e. while the above scales according to $n^{1-\frac{1}{\gamma}}$ we can make the constant term as small as we need to so that it is negligible compared to the overall constant as derived in the main text.

Finally, we can combine this with the result from the main text (in which the remaining $\leq n$ target points are good) to achieve the same bound; this is because we can collect the good target points and then the bad target points with only an additional constant (added for the arc between the last good target visited and the first bad target).

\section{Euclidean TSP for Fractally-Distributed Targets} \label{appx:fractal-targets}


While $\gamma$ is an integer for control-affine systems and continuous target point distributions, it is interesting theoretically to consider the case of noninteger $\gamma$. This can occur when the targets are distributed over a set of fractal dimension. Since this appendix is meant to illustrate how our techniques (particularly the Hierarchical Cell Structures) can be applied in different settings, we will make some simplifying assumptions.


First, we limit ourselves to targets distributed in strictly self-similar fractal sets in $\bbR^d$ and a single scaling factor (such as the Sierpinski triangle, Menger sponge, or Koch curve) and the Euclidean TSP with varying top speed. Note that in this case $\wspace = \cspace = \bbR^d$, so there is no distinction between $\bq \in \cspace$ and $x \in \wspace$.\footnote{This means that standard TSP approximation algorithms such as Christofides can be applied, but our analysis is still needed to determine the tour length as $n \to \infty$.} We will therefore refer to configurations as $x \in \bbR^d$, and the reachable sets $R_\varepsilon(x)$ are the same for both workspace and configuration space. 

Formally, we assume that $\wspacef$ (the set on which the targets are distributed) is composed of $\hat{b}$ copies of itself at $s^{-1}$ scale (translated and/or rotated and/or reflected), which makes imposing a Hierarchical Cell Structure with Euclidean dynamics trivial. Using $\gamma$ as revealed through the HCS, we find that $\gamma = \log_s \hat{b}$ is the Hausdorff dimension of $\wspacef$; we assume that $\gamma \geq 1$ (otherwise the $O(1)$ time to travel between HCS's in the HCS cover will overwhelm the $O(n^{1-\frac{1}{\gamma}})$ tour time from the hierarchical collection problem). Additionally, choosing a random $X \in \wspacef$ can be done by iteratively choosing sub-copies.


We also limit ourselves to considering the Euclidean TSP over such distributions, with a top speed function $h$, i.e. our control law is
\begin{align}
    \dot{x} = h(x) \bu 
\end{align}
where the control set is $\cU = \{\bu \in \bbR^d : \norm{\bu}_2 \leq 1 \}$. We assume $h$ is Lipschitz-continuous and $h_{\min} \defeq \inf_x h(x) > 0$ and $h_{\max} \defeq \sup_x h(x) < \infty$. Formally, the Lipschitz continuity of $h$ allows us to get a $\zeta$-approximate HCS cover for any $\zeta > 0$ (locally $h$ is approximately constant so we can just use nesting cells at a small scale $\varepsilon_0$).

Instead of `density $f$' and `agility $g$', we let
\begin{align}
    \phi(x) = \lim_{\varepsilon \to 0} \frac{\bbP_{X \sim f}[X \in \bar{R}_\varepsilon(x)]}{\varepsilon^\gamma}
\end{align}
(where $X \sim f$ means the given distribution over the fractal, even though $f$ is no longer a probability density function). Then we replace the integral $\int f(x)^{1-\frac{1}{\gamma}} g(x)^{-\frac{1}{\gamma}} \, dx$ with its equivalent $\bbE_{X \sim f} [\phi(x)^{-\frac{1}{\gamma}}]$.

Our upper bound can still tell us about this case:
\begin{proposition}
    For noninteger $\gamma$, we get the following bounds: if $\gamma \geq 3$ is noninteger, then bounds \eqref{eq:dstsp-bound-upper-01}, \eqref{eq:dstsp-bound-upper-02}, and \eqref{eq:dstsp-bound-upper-03} from \Cref{prop:dstsp-upper-bound} still hold. If $\gamma \in (2,3)$, then for all sufficiently large $n$,
    \begin{align}
        \bbP\big[\tsp_{\Pi}(\{X_i\}) \leq (1+\delta) \big(12 s \alpha^{-\frac{1}{\gamma}}\big)n^{1-\frac{1}{\gamma}} \bbE_{X \sim f} [\phi(x)^{-\frac{1}{\gamma}}] \big]
        \\ \geq 1 - e^{-\frac{(\gamma-2) p_1 n (\sum_{j=1}^m p_j^{1-1/\gamma})^2}{13 \gamma}}
    \end{align}
    And, finally, if $\gamma \in (1,2)$, then for all sufficiently large $n$,
    \begin{align}
        \bbP\big[\tsp_{\Pi}(\{X_i\}) \leq (1+\delta) \big(12 s \alpha^{-\frac{1}{\gamma}}\big)n^{1-\frac{1}{\gamma}} \bbE_{X \sim f} [\phi(x)^{-\frac{1}{\gamma}}] \big]
        \\ \geq 1 - e^{-\frac{p_1 n^{2(1-1/\gamma)} (\sum_{j=1}^m p_j^{1-1/\gamma})^2} {\frac{560}{3} \log(\gamma/(\gamma-1)) + 4 + 18 (1-1/\gamma)^2 \frac{\gamma}{\gamma-2} (\frac{280}{3} \log(\gamma/(\gamma-1)) )^{1-2/\gamma}  }}
    \end{align}
\end{proposition}

\begin{proof}
    This is proved in the same manner as the upper bounds in \Cref{sec:dstsp-ub} using the HCS cover and the collection problem.
\end{proof}



\end{document}